%% file: acl_latex.tex
\pdfoutput=1

\documentclass[11pt]{article}

\usepackage[preprint]{acl}

\usepackage{times}
\usepackage{latexsym}
\usepackage{dsfont}

\usepackage{tikz}
\usepackage{xcolor}
\usepackage{enumitem}
\usetikzlibrary{fit,positioning,calc,arrows.meta,backgrounds,decorations.pathmorphing}
\usepackage{graphicx}
\usepackage{subcaption}
\usepackage{colortbl}

\colorlet{arithmetic_circuit}{cyan!50!gray}
\colorlet{validation_circuit}{red!50!gray}
\colorlet{template_problem}{teal!50!gray}
\colorlet{template_solution}{orange!60!gray}
\colorlet{template_reasoning}{blue!40!gray}
\colorlet{template_error_position}{cyan!50!black}

\input{math_commands.tex}

\usepackage[T1]{fontenc}

\usepackage[utf8]{inputenc}

\usepackage{microtype}

\usepackage{inconsolata}

\usepackage{graphicx}
\usepackage{booktabs}
\usepackage{multirow}
\usepackage{subcaption}

\usepackage{cleveref}
\crefname{figure}{Figure}{Figures}
\Crefname{figure}{Figure}{Figures}

\newcommand{\customsize}{\fontsize{6pt}{8pt}\selectfont}

\title{The Validation Gap: A Mechanistic Analysis of How Language Models Compute Arithmetic but Fail to Validate It}

\author{Leonardo Bertolazzi\hspace{0.1mm}\thanks{Equal contribution. Listing order is random.}\textsuperscript{\normalfont 1} \hspace{3mm} Philipp Mondorf\hspace{0.3mm}\footnotemark[1]\textsuperscript{\normalfont 2, 3} \hspace{3mm} Barbara Plank\textsuperscript{\normalfont 2, 3} \hspace{3mm} Raffaella Bernardi\textsuperscript{\normalfont 4}\\
  \textrm{\textsuperscript{1}}DISI, University of Trento, Italy \\
  \textrm{\textsuperscript{2}}MaiNLP, Center for Information and Language Processing, LMU Munich, Germany \\
  \textrm{\textsuperscript{3}}Munich Center for Machine Learning (MCML), Munich, Germany \\
  \textrm{\textsuperscript{4}}Free University of Bozen-Bolzano, Italy \\
  {\tt \footnotesize leonardo.bertolazzi@unitn.it} \hfill {\tt \footnotesize \{p.mondorf, b.plank\}@lmu.de} \hfill {\tt \footnotesize raffaella.bernardi@unibz.it
}}

\begin{document}
\maketitle
\begin{abstract}
The ability of large language models (LLMs) to validate their output and identify potential errors is crucial for ensuring robustness and reliability. However, current research indicates that LLMs struggle with self-correction, encountering significant challenges in detecting errors. While studies have explored methods to enhance self-correction in LLMs, relatively little attention has been given to understanding the models' internal mechanisms underlying error detection. In this paper, we present a mechanistic analysis of error detection in LLMs, focusing on simple arithmetic problems. Through circuit analysis, we identify the computational subgraphs responsible for detecting arithmetic errors across four smaller-sized LLMs. Our findings reveal that all models heavily rely on \emph{consistency heads}\textemdash{}attention heads that assess surface-level alignment of numerical values in arithmetic solutions. Moreover, we observe that the models' internal arithmetic computation primarily occurs in higher layers, whereas validation takes place in middle layers, before the final arithmetic results are fully encoded. This structural dissociation between arithmetic computation and validation seems to explain why smaller-sized LLMs struggle to detect even simple arithmetic errors.
\end{abstract}

\section{Introduction}\label{sec:Introduction}
In recent years, large language models have demonstrated notable performance across a variety of reasoning tasks, including arithmetic problem-solving~\citep{sawada2023arb, phan2025humanity, liu2024deepseek, achiam2023gpt}. However, a gap appears to exist between the models' ability to generate solutions and their capacity to validate them effectively~\citep{huang2024cannotselfcorrect, hong-etal-2024-closer, DBLP:journals/corr/abs-2404-04298}. Specifically, while LLMs are often able to correct mistakes once they have been identified, they struggle to detect errors in the first place~\citep{tyen2024errorlocation, kamoi2024evaluatingdetecting, kamoi2024correctionsurvey}.

\begin{figure}[t!]
  \centering 
  \input{figures/tikz/intro_figure}
  \caption{
  A schematic overview of the structurally dissociated circuits responsible for \textcolor{arithmetic_circuit}{arithmetic computation} and \textcolor{validation_circuit}{validation}. While the models' internal arithmetic computation primarily occurs in higher layers, validation takes place in mid-to-lower layers, before the final arithmetic results are fully encoded (see Section~\ref{subsec:Dissociation}).
  }
  \label{fig:schematic_overview}
\end{figure}

Several studies have proposed methods to enhance LLMs' ability to detect errors and correct their own output~\citep{welleck2023learnselfcorrect, selfee2023, gou2024critic}. However, comparatively little attention has been given to understanding why current models inherently struggle with error detection~\citep{hong-etal-2024-closer, kamoi2024evaluatingdetecting, DBLP:journals/corr/abs-2402-12563}. In particular, few studies have examined the internal mechanisms responsible for error detection in LLMs~\citep{liu2024moralselfcorrection}.

In this paper, we seek to bridge this gap by presenting a \emph{mechanistic} analysis of error detection in LLMs, focusing on math word problems involving basic addition. We examine four LLMs\textemdash{}Qwen-2.5-(Math)-1.5B-Instruct~\citep{DBLP:journals/corr/abs-2407-10671, yang2024qwen2}, Llama-3.2-3B-Instruct~\citep{dubey2024llama}, and Phi-3-Mini-4k-Instruct~\citep{abdin2024phi}\textemdash{}to understand how these models detect arithmetic errors and why they struggle with this task. Specifically, we identify and analyze the computational subgraphs (or \emph{circuits}) responsible for error detection, examining the role of identified modules within the broader task context. Additionally, we analyze how these circuits compare to those involved in computing arithmetic results, seeking to understand the structural differences between arithmetic computation and validation in LLMs. To the best of our knowledge, this is the first study to examine arithmetic error detection in LLMs through the lens of mechanistic interpretability. Our findings reveal:

\begin{itemize}[itemsep=1pt]
    \item Circuits for detecting arithmetic errors are structurally similar across different models.
    \item The error detection process is governed by \emph{consistency heads}\textemdash{}attention heads located in lower to middle layers that check for surface-level alignment of numerical values in the arithmetic solution. By patching a small subset of these heads, we can effectively control the models’ error detection behavior.
    \item The mechanisms for arithmetic computation and validation appear to be structurally dissociated. While the models' internal arithmetic computation is predominantly conducted in higher layers, validation is performed in middle layers, before the final arithmetic results are fully encoded (see Figure~\ref{fig:schematic_overview}).
    \item Adding latent activations from higher layers to the residual stream in lower layers significantly enhances the models' ability to detect errors, effectively closing the validation gap.\footnote{By validation gap, we mean both the structural separation and the performance gap between arithmetic computation and validation in LLMs.}
\end{itemize}

Our analysis shows that mechanistic interpretability can offer valuable insights into how models detect\textemdash{}or fail to detect\textemdash{}arithmetic errors. While focused on smaller LLMs and basic math word problems, we hope that this work provides a foundation for future work to study error detection in larger models and more complex tasks.

\section{Background}\label{sec:Circuit_Analysis}
A common goal in interpretability research is to gain a deeper understanding of the internal mechanisms that drive the behavior of language models for a given task~\citep{ferrando2024primer, DBLP:journals/corr/abs-2408-01416, bereska2024mechanistic}. The \emph{circuit} framework seeks to achieve this by identifying model components that causally influence the model's task output~\citep{elhage2021mathematical, wang2023interpretability, NEURIPS2023_efbba771}. In essence, a \emph{circuit} refers to the computational subgraph $\mathcal{C} \subset \mathcal{G} = (\mathcal{V}, \mathcal{E})$ that represents the task-relevant flow of information across the model's layers~\citep{NEURIPS2023_34e1dbe9, bhaskar2024finding}. A node $v \in \mathcal{V}$ in this graph can represent different components, depending on the desired level of granularity\textemdash{}ranging from entire attention or MLP layers, to individual attention heads, to single neurons~\citep{DBLP:journals/corr/abs-2408-01416}. An edge $e_{ij} = (v_i, v_j) \in \mathcal{E}$ denotes a connection between two nodes, where the output of the source node $v_i$ serves as the input to the destination node $v_j$. The total input received by a node $v_j$ can be expressed as $\sum_{e_{ij} \, = \, (v_i, v_j) \, \in \, \mathcal{E}_{v_j}} \vz_i$ where $\vz_i$ represents the activation of node $v_i$ and $\mathcal{E}_{v_j}$ denotes the set of incoming edges to $v_j$.

\paragraph{Circuit Identification.} A method for identifying circuits in language models is \emph{activation patching}~\citep{NEURIPS2020_92650b2e, NEURIPS2021_4f5c422f, NEURIPS2022_6f1d43d5}. The key idea is to intervene on the latent activations of components in the computation graph $\mathcal{G}$ to measure their indirect effect~\citep{pearl2001effects} on the model's output. Adopting the terminology of~\citet{zhang2024towards}, activation patching requires three forward passes to determine a component's indirect effect for a given input:

\begin{figure*}[t!]
  \centering 
  \input{figures/tikz/data_setup}
  \caption{
  Our data generation setup. We use eight templates to generate samples that consist of a simple \textcolor{template_problem}{arithmetic problem}, its corresponding \textcolor{template_solution}{solution}, and a \textcolor{template_reasoning}{final statement} assessing the solution’s validity. Words enclosed in [square brackets] serve as placeholders for components that are substituted with specific content. For each generated sample, a pair of (\emph{clean}, \emph{corrupt}) prompts is derived. Counterintuitively, \emph{clean} prompts \emph{contain} errors, as they represent prompts for which the model exhibits the desired error detection behavior (predicting ``invalid'').
  }
  \label{fig:data_setup}
\end{figure*}

\begin{enumerate}
    \item Clean run: run the model on a \emph{clean} prompt $X_{clean}$, for which the model generates the desired task-specific output $y_{clean}$. Cache the component's latent activations, denoted as $\vz_i$.
    \item Corrupted run: run the model on a \emph{corrupted} prompt $X_{corrupt}$, for which the model generates a related but altered output $y_{corrupt}$.
    \item Patched run: run the model on $X_{corrupt}$, but this time, replace the component’s activations associated with $X_{corrupt}$ with the cached activations $\vz_i$ from the clean run.
\end{enumerate}

Finally, the indirect effect is calculated by comparing the output of the \emph{patched} run to that of the \emph{corrupted} run using a predefined metric $\mathcal{P}$.\footnote{This metric typically evaluates differences in logits or output probabilities relative to the clean output $y_{clean}$.} If the component under consideration causally influences the model's task output, the patched activations should shift the prediction $y_{corrupt}$ toward $y_{clean}$.

As performing these steps for every model component and sample can become computationally expensive, several approximations trade off computational cost against accuracy~\citep{syed-etal-2024-attribution, nanda_attribution_patching, hanna2024have}. In this work, we consider \emph{\textbf{edge attribution patching}} \textbf{(EAP)}~\citep{syed-etal-2024-attribution}, a linear approximation of activation patching requiring only two forward and one backward pass. EAP focuses on the indirect effect of \emph{\textbf{edges}} $e_{ij} \in \mathcal{E}$, which represent inputs to a node $v_j$ from earlier nodes $v_i$. Specifically, the causal influence is approximated using the \emph{absolute attribution score} $|\nabla_{\vz} \mathcal{P}|$, which measures the change in $\mathcal{P}$ under the intervention (for further details, see Appendix \ref{app:circuit}). Once these scores are computed, the top-$k$ edges with the highest absolute attribution values are selected to define the circuit $\mathcal{C}$. Although EAP is only a linear approximation of activation patching, it has been successfully employed in studies to identify circuits within language models for tasks such as indirect object identification, subject-verb agreement, and greater-than attribution~\citep{syed-etal-2024-attribution, hanna2024have, miller2024autocircuit}.

\section{Circuits for Arithmetic Error Detection}\label{sec:Circuits_Arithmetic_Error_Detection}
In this section, we present the dataset used to study arithmetic error detection in LLMs. Additionally, we outline our use of \emph{edge attribution patching} to identify circuits responsible for the task.

\begin{table*}[tbp]
\small
\centering
\begin{tabular}{lcccc}
\toprule
Error Type & \textbf{Qwen-2.5-1.5B} & \textbf{Qwen-2.5-Math-1.5B} & \textbf{Llama-3.2-3B} & \textbf{Phi-3-Mini-3.8B} \\
\midrule
\textit{Arithmetic Result} & 60.53 $\pm$ 10.92 & 98.99 $\pm$ 2.18 & 87.67 $\pm$ 12.51 & 89.51 $\pm$ 26.75 \\
\textit{Numeric Answer} & 59.03 $\pm$ 10.72 & 98.53 $\pm$ 3.17 & 86.44 $\pm$ 13.35 & 89.44 $\pm$ 26.84 \\
\bottomrule
\end{tabular}
\caption{Accuracy of models in correctly classifying the solutions' validity of (\emph{clean}, \emph{corrupt}) prompt pairs. Values represent the mean accuracy across all templates, reported with their corresponding standard deviation.}
\label{tab:behavioral_accuracy}
\end{table*}

\paragraph{Dataset.} In this study, we focus on simple math word problems. As illustrated in Figure~\ref{fig:data_setup}, we employ templates to systematically generate data~\citep{wang2023interpretability, NEURIPS2023_efbba771}. Each sample consists of a basic \textcolor{template_problem}{arithmetic problem}, its corresponding \textcolor{template_solution}{solution}, and a \textcolor{template_reasoning}{final statement} that evaluates the solution’s validity. Samples derived from the same template maintain a consistent sentence structure but incorporate \emph{\textbf{variable}} components such as [names] or numerical [values] (left box in Figure~\ref{fig:data_setup}). To analyze the models' error detection mechanisms, we introduce simple arithmetic errors into the sample's \textcolor{template_solution}{solution} statement. Specifically, we consider \emph{two} types of errors separately: \emph{i)} a miscalculation of the arithmetic \textcolor{template_error_position}{result}, and \emph{ii)} an incorrect final numeric \textcolor{template_error_position}{answer}. Note that the \emph{\textbf{perturbed}} sample forms our \emph{clean prompt}, for which models can \emph{\textbf{successfully detect}} the arithmetic error (see upper-right box in Figure~\ref{fig:data_setup}, showing an \textcolor{red!60!black}{error} for the arithmetic \textcolor{template_error_position}{result}). Additionally, we construct a \emph{corrupt prompt} without errors, for which models predict the solution to be ``valid''  (lower-right box). We use single-digit numerical values that sum to a two-digit arithmetic result across all templates and samples. The introduced errors always correspond to a different, incorrect two-digit number ranging from 10 to 19.

For each template, we generate 6,000 pairs of (\emph{clean}, \emph{corrupt}) prompts. We use eight different templates that vary in syntactic structure and token length while preserving the fundamental task logic. Each type of error (arithmetic \textcolor{template_error_position}{result} vs. final numeric \textcolor{template_error_position}{answer}) is examined separately. In total, we obtain a dataset of 6,000 (\emph{clean}, \emph{corrupt}) prompt pairs for each template $\mathcal{T}_{\{1:8\}}$ per error type. For further details on the data generation process and templates, please refer to Appendix~\ref{app:dataset}.

\paragraph{Method.} As described in Section~\ref{sec:Circuit_Analysis}, we employ \emph{edge attribution patching} (EAP)~\citep{syed-etal-2024-attribution} to identify circuits responsible for arithmetic error detection in LLMs. For each template $\mathcal{T}_i$, we use 5,000 pairs of (\emph{clean}, \emph{corrupt}) prompts to determine the circuit\textemdash{}specifically, the set of edges $\mathcal{C}_i = \mathcal{E}_i \subset \mathcal{E}$\textemdash{}that causally influences the model's error detection behavior (see Section~\ref{sec:Circuit_Analysis} for more details). Since all samples within a template contain the same number of tokens, we apply \emph{\textbf{token-wise}} EAP, which allows us to assess the causal impact of edges at each token position of the prompt. Following~\citet{syed-etal-2024-attribution}, the \emph{absolute attribution score} $|\nabla_{\vz} \mathcal{P}|$ is computed using the average logit difference related to the models' answer tokens (``valid'', ``invalid'') as metric $\mathcal{P}$ (see Appendix~\ref{app:circuit} for further details). Once the attribution scores are obtained, we use the template's remaining 1,000 (\emph{clean}, \emph{corrupt}) prompt pairs to find the minimal set of top-$k$ edges for which the circuit achieves a faithfulness score between 99\%--101\%. For a more detailed explanation of the search procedure and the computation of the faithfulness score, please refer to Appendix~\ref{app:faithfulness} and~\ref{app:circuit_search}.

\paragraph{Soft Intersection Circuit.} After identifying a circuit $\mathcal{C}_i$ for each template $\mathcal{T}_i \in \{\mathcal{T}_1, \ldots, \mathcal{T}_8\}$, we aim to find a \emph{\textbf{final}} subset of edges $\mathcal{E}_{\mathcal{C}} \subset \mathcal{E}$ that \emph{\textbf{generalizes across all templates}} $\mathcal{T}_i$, ensuring high faithfulness. To achieve this, we compute the \emph{soft intersection circuit}, which includes edges present in at least $\frac{1}{8} \leq \tau \leq \frac{8}{8}$ of the identified circuits $\{\mathcal{C}_1, \ldots, \mathcal{C}_8\}$. The soft intersection is defined through a \emph{membership function} that determines the proportion of identified circuits in which a given edge $e \in \mathcal{E}$ appears:

\begin{equation}
    f(e) = \frac{1}{8}\sum_{i=1}^8 \mathds{1}_{\mathcal{C}_i}(e)
\end{equation}

where $\mathds{1}_{\mathcal{C}_i}(e)$ is an indicator function that assigns a value of 1 if $e \in \mathcal{C}_i$ and 0 otherwise. Consequently, $f(e)$ takes values in $\{0, \frac{1}{8}, \ldots, \frac{8}{8}\}$. The soft intersection circuit is then formally defined as $\mathcal{C}^{(\tau)} = \mathcal{E}_{\mathcal{C}^{(\tau)}} = \{e \in \mathcal{E} \mid f(e) \geq \tau\}$. This formulation allows for a flexible trade-off: setting $\tau = \frac{1}{8}$ yields the union of all identified circuits, while $\tau = \frac{8}{8}$ results in their strict intersection.\footnote{Note that since we employ token-wise EAP to identify relevant edges \emph{per token position}, we assign abstract yet meaningful labels to each token position, ensuring transferability across templates. Further details on this labeling process can be found in Appendix~\ref{app:align_token_pos}.} By varying $\tau$ from $\frac{1}{8}$ to $\frac{8}{8}$, we balance faithfulness against the numbers of edges considered, progressively filtering out template-specific redundant edges.

\paragraph{Models.} In this study, we consider four different LLMs\textemdash{}Qwen-2.5-(Math)-1.5B-Instruct~\citep{DBLP:journals/corr/abs-2407-10671, yang2024qwen2}, Llama-3.2-3B-Instruct~\citep{dubey2024llama}, and Phi-3-Mini-4k-Instruct~\citep{abdin2024phi}, 
to assess the influence of varying architectures, model scales, and fine-tuning procedures (particularly for math). Appendix~\ref{app:models} provides details about models and prompts. Our code is publicly available at: 
\href{https://github.com/mainlp/validation-gap}{{https://github.com/mainlp/validation-gap}}.

\section{Experiments}\label{sec:Experiments}
\begin{figure*}[tbp]
    \centering
    \begin{subfigure}{0.45\textwidth}
        \centering
        \includegraphics[width=\textwidth]{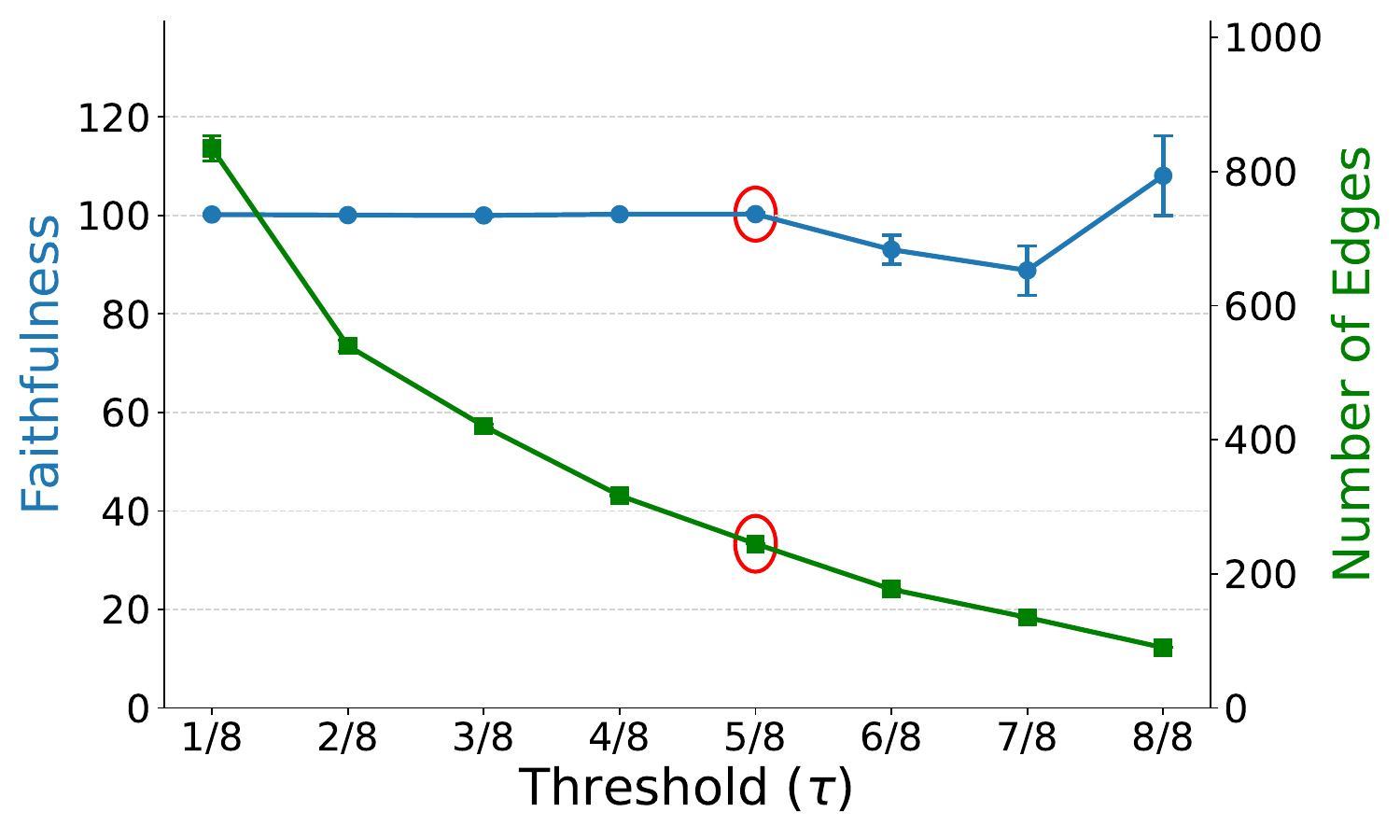}
        \caption{Error at the arithmetic result}
        \label{fig:soft-intersection-qwen-a}
    \end{subfigure}
    \hfill
    \begin{subfigure}{0.45\textwidth}
        \centering
        \includegraphics[width=\textwidth]{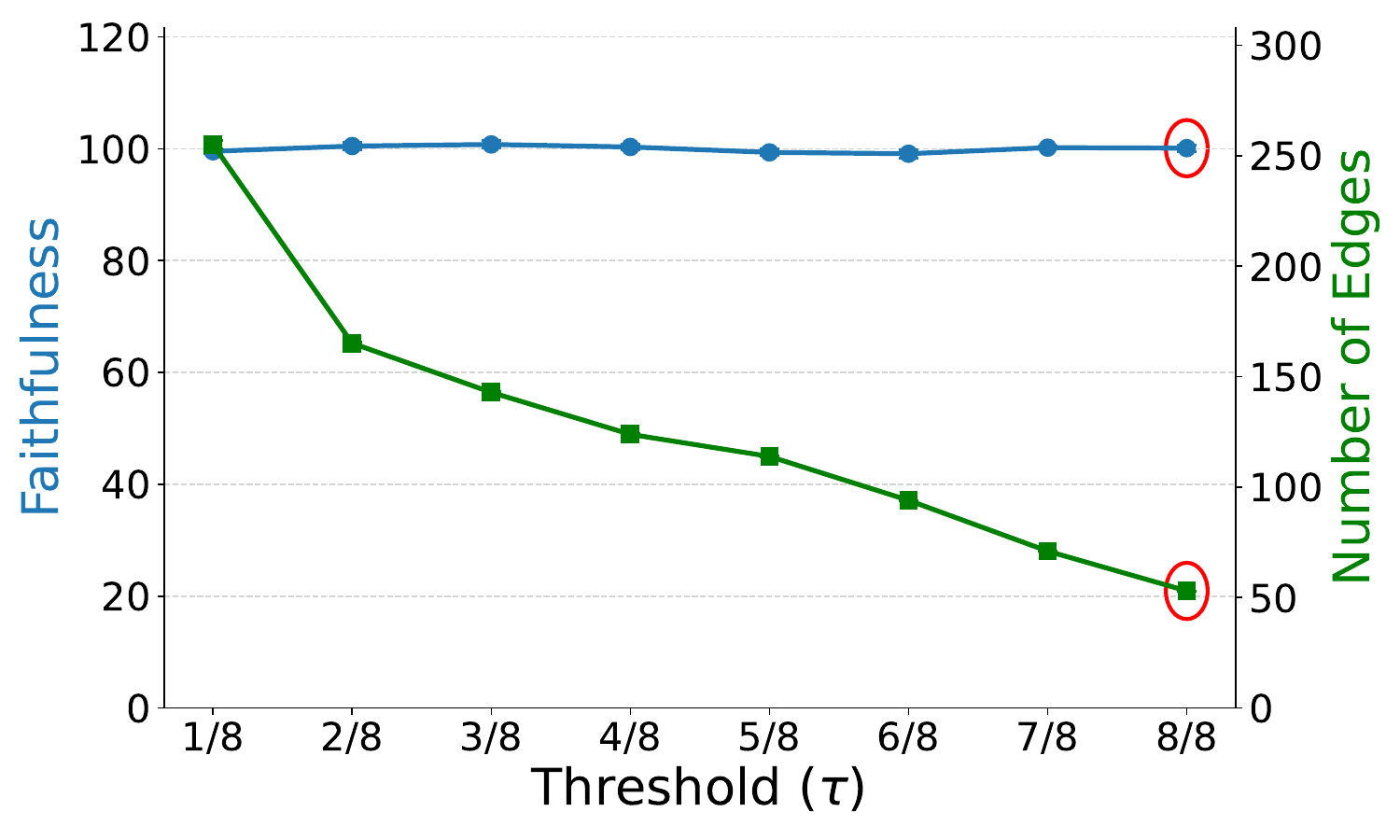}
        \caption{Error at the final numeric answer}
        \label{fig:soft-intersection-qwen-b}
    \end{subfigure}
    \caption{The number of edges and faithfulness scores averaged across all templates (with standard deviation) of the \emph{soft intersection circuit} for different $\tau$ values. Red circles mark the circuit that best balances size and faithfulness.}
    \label{fig:soft-intersection-qwen}
\end{figure*}

Before identifying circuits, we ensure that all models are capable of detecting the types of arithmetic errors described in Section~\ref{sec:Circuits_Arithmetic_Error_Detection}. Specifically, we randomly generate 5,000 (\emph{clean}, \emph{corrupt}) prompt pairs for each template and evaluate whether the models predict tokens that correctly indicate the validity of the presented solutions (we expect predictions such as ``invalid'', ``incorrect'', or ``wrong'' for \emph{clean} prompts, and ``valid'', ``correct'', or ``right'' for \emph{corrupt} prompts). Table~\ref{tab:behavioral_accuracy} summarizes the models' average accuracy along with the standard deviation across templates. A pair is considered correctly classified if the model's highest logit falls within the set of correct labels for the clean and corrupt prompts, respectively. We observe that all models are able to detect the type of errors considered, with Qwen-2.5-Math-1.5B achieving near-perfect accuracy, while Qwen-2.5-1.5B performs worst with approximately 60\% accuracy for both error types.

In the subsequent circuit identification process, we filter the generated prompt pairs to ensure that for all samples, the models predict the desired outputs, i.e., $y_{clean} \in \{\text{``invalid''}, \text{``incorrect''}, \text{``wrong''}\}$ and $y_{corrupt} \in \{\text{``valid''}, \text{``correct''}, \text{``right''}\}$ (see Section~\ref{sec:Circuit_Analysis}).

\subsection{Identified Circuits}\label{subsec:results_error_detection_circuits}
As described in Section~\ref{sec:Circuits_Arithmetic_Error_Detection}, we employ edge attribution patching to identify circuits $\mathcal{C}_i = \mathcal{E}_i \subset \mathcal{E}$ that achieve faithfulness scores between 99\% and 101\% for each template $\mathcal{T}_i \in \{\mathcal{T}_1, \ldots \mathcal{T}_8\}$. We find that for all models, only 100 to 900 edges (less than 0.1\% of total edges) are sufficient to achieve around 100\% faithfulness for the task. Due to space constraints, we present the faithfulness scores and the corresponding number of edges for each circuit in Table~\ref{tab:faith_size} in the Appendix\textemdash{}categorized by model, template, and error type. Once a circuit is identified for each template, we compute the \emph{soft intersection circuit} $\mathcal{C}^{(\tau)}$ as outlined in Section~\ref{sec:Circuits_Arithmetic_Error_Detection}. Figure~\ref{fig:soft-intersection-qwen} illustrates the faithfulness scores and associated edge counts of the \emph{soft intersection circuits} for Qwen-2.5-1.5B across different threshold values $\tau \in \{\frac{1}{8}, \ldots, \frac{8}{8}\}$. For errors at the position of the arithmetic result (e.g., ``$5+8=\textcolor{red!60!black}{16}$''), the soft intersection circuit $\mathcal{C}_{\text{result}}^{(5/8)}$ achieves an average faithfulness of around 100\%, effectively \emph{\textbf{generalizing across all templates}}, while retaining only 245 edges (red circles in Figure~\ref{fig:soft-intersection-qwen-a}). For errors at the position of the final numeric answer (e.g., ``\ldots $5+8=13$. Thus, Jane has $\textcolor{red!60!black}{16}$ apples.''), even the strict intersection ($\tau = \frac{8}{8}$) yields an almost perfect average faithfulness score across all templates, with the number of relevant edges reduced to 53 (see Figure~\ref{fig:soft-intersection-qwen-b}). In subsequent analyses, we focus on $\mathcal{C}_{\text{result}}^{(5/8)}$ and $\mathcal{C}_{\text{answer}}^{(8/8)}$ for Qwen-2.5-1.5B. Notably, similar results can be found for \emph{\textbf{all}} models, as presented in Appendix~\ref{app:error_detection_circuits}, Figures~\ref{fig:faithfulnes_qwen-math_full} to~\ref{fig:faithfulnes_phi_full}.

\begin{figure}[bp!]
  \centering
  \includegraphics[width=\linewidth]{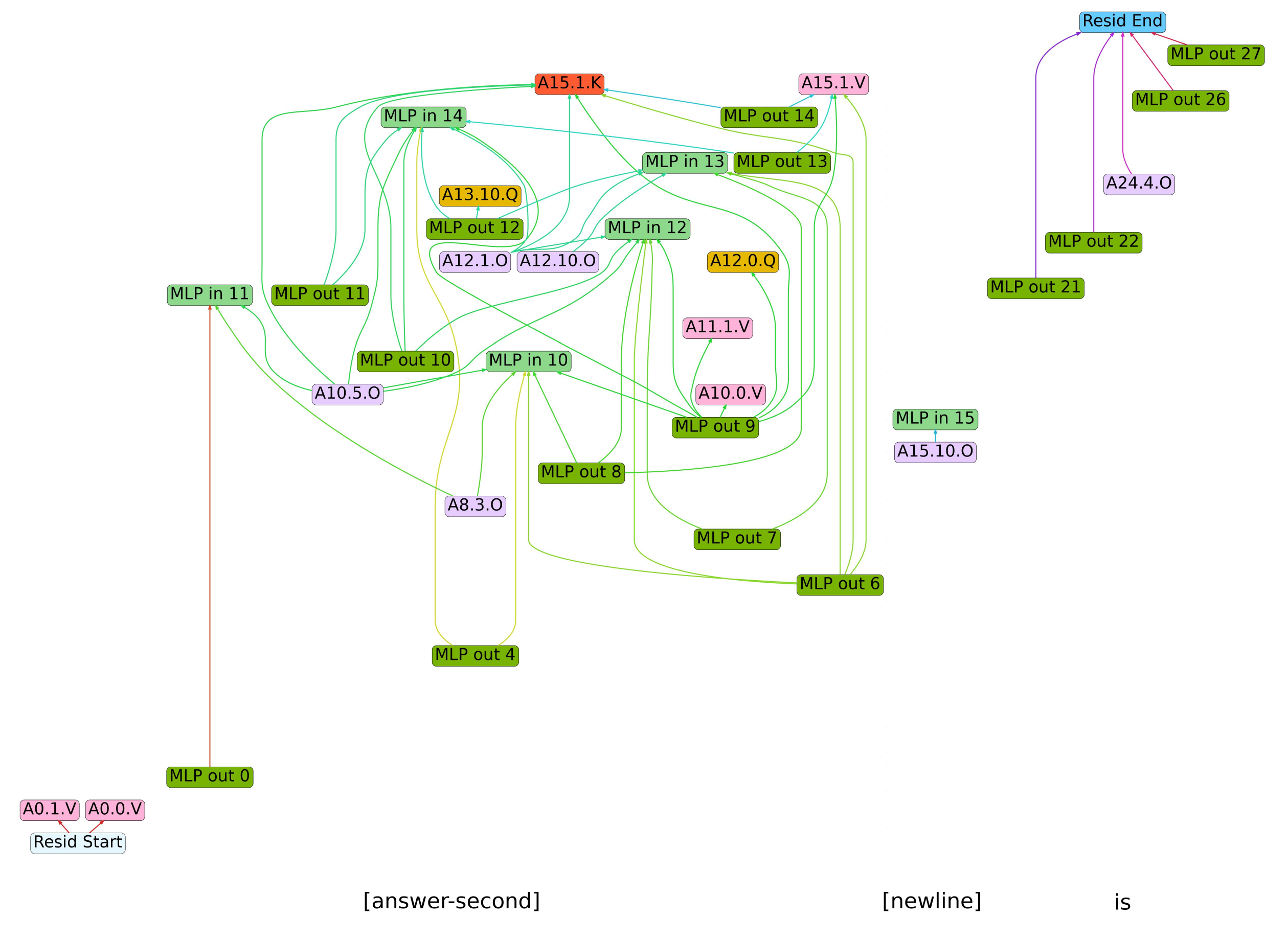}
  \caption{The soft intersection circuit $\mathcal{C}^{(8/8)}_{\text{answer}}$, representing the set of edges that causally influence the output of Qwen-2.5-1.5B when detecting errors at the position of the final numeric answer. Attention heads are abbreviated as \emph{A.layer.head.K\tiny{(ey)}}\emph{/V\tiny{(alue)}}\emph{/Q\tiny{(uery)}}\emph{/O\tiny{(ut)}}, while MLPs are represented as \emph{MLP in/out layer}. Corresponding token positions are indicated by the labels at the bottom.}
  \label{fig:qwen_numeric_answer_circuit}
\end{figure}

\begin{figure*}[tbp]
    \centering
    \includegraphics[width=1\textwidth]{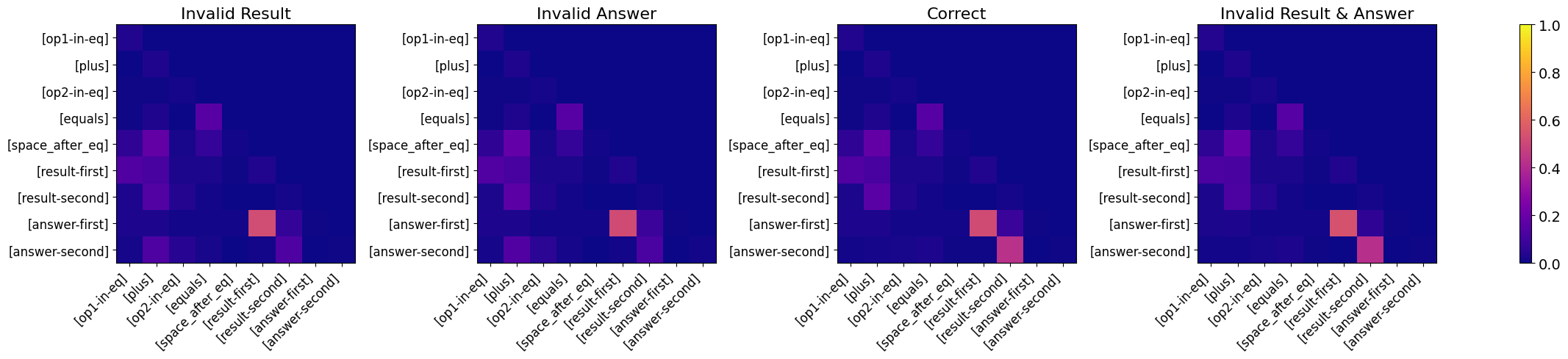}
    \caption{Attention patterns of a \emph{consistency head} in  Qwen-2.5-1.5B (head 2 in layer 12). Reported scores are averaged over 5,000 prompts where (\emph{left}) an error is present at the position of the arithmetic result, (\emph{second to left}) an error is present at the position of the final numeric answer, (\emph{second to right}) no error is present, and (\emph{right}) a consistent error is present at both considered positions.}
    \label{fig:consistency-heads}
\end{figure*}

A visualization of Qwen-2.5-1.5B's circuit $\mathcal{C}_{\text{answer}}^{(8/8)}$ for detecting errors at the position of the final numeric answer is shown in Figure~\ref{fig:qwen_numeric_answer_circuit}. As mentioned in Section~\ref{sec:Circuits_Arithmetic_Error_Detection}, we employ \emph{\textbf{token-wise}} EAP. This means that we identify edges for each token position of the prompt. Figure~\ref{fig:qwen_numeric_answer_circuit} shows that most edges are concentrated at the position of the second digit of the final numeric answer ([answer-second])\footnote{Note that Qwen-2.5 uses a one-digit tokenization scheme; i.e., the number $16$ is encoded into two separate tokens.} in the middle layers 4 to 15 of the model. For example, this corresponds to the position of the $\textcolor{red!60!black}{6}$ in ``\ldots $5+8=13$. Thus, Jane has $\textcolor{red!60!black}{16}$ apples.'' Additionally, a smaller number of edges appear in higher layers 21 to 27 at the final token position of the prompt (``is''), predominantly connecting MLP layers with the final residual output. Interestingly, this structural pattern appears \emph{\textbf{consistent}} across all models (see~\cref{fig:qwen_math_z2_circuit,fig:llama_z2_circuit,fig:phi_z2_circuit} in the Appendix) and even across error types (\cref{fig:qwen_z1_circuit,fig:qwen_math_z1_circuit,fig:llama_z1_circuit,fig:phi_z1_circuit} in the Appendix).

\paragraph{One Circuit for Arithmetic Error Detection.} The structural similarity between circuits identified for the two distinct error types is further supported by their edge overlap. When computing the Intersection over Minimum (IoM)\footnote{Detailed information on the computation of the edge overlap between circuits can be found in Appendix~\ref{app:method_edge_overlap}.} between the circuits $\mathcal{C}_{\text{result}}^{(5/8)}$ and $\mathcal{C}_{\text{answer}}^{(8/8)}$, we obtain a value of 0.92, indicating that a substantial subset of edges of $\mathcal{C}_{\text{answer}}^{(8/8)}$ is also present in $\mathcal{C}_{\text{result}}^{(5/8)}$. Additionally, we compute the intersection of $\mathcal{C}_{\text{result}}^{(5/8)}$ and $\mathcal{C}_{\text{answer}}^{(8/8)}$, and evaluate the faithfulness of the resulting circuit. Despite comprising only 49 edges, the intersected circuit achieves an average faithfulness score of $78.60\% \pm 7.46\%$ on samples involving errors in the arithmetic result and $82.50\% \pm 5.08\%$ on samples with errors at the final numeric answer. Notably, these observations are consistent across all models considered. Corresponding results are provided in Appendix~\ref{app:results_edge_overlap}, specifically in Table~\ref{tab:iou_iom} and~\ref{tab:z1-z2-intersection}.

\begin{table*}[tbp]
\small
\centering
\begin{tabular}{lcccc}
\toprule
Error Type & \textbf{Qwen-2.5-1.5B} & \textbf{Qwen-2.5-Math-1.5B} & \textbf{Llama-3.2-3B} & \textbf{Phi-3-Mini-3.8B} \\
\midrule
\textit{Result \& Answer} & 12.39 $\pm$ 6.00 & 3.37 $\pm$ 2.06 & 12.27 $\pm$ 9.08 & 40.98 $\pm$ 18.41 \\
\bottomrule
\end{tabular}
\caption{Accuracy of models in correctly classifying the solutions' validity of (\emph{clean}, \emph{corrupt}) prompt pairs where \emph{clean} prompts contain a \emph{\textbf{consistent}} error at both the position of the arithmetic result and the position of the final numeric answer. Values represent the mean accuracy across all templates, reported with their standard deviations.}
\label{tab:both_behavioral_accuracy}
\end{table*}

\subsection{Decoding the Error Detection Process}\label{subsec:Decode_Error_Detection}
Once we obtain a soft intersection circuit $\mathcal{C}^{(\tau)}$, we can analyze its components to gain deeper insights into the model’s error detection mechanisms. For instance, Figure~\ref{fig:qwen_numeric_answer_circuit} illustrates that $\mathcal{C}_{\text{answer}}^{(8/8)}$ contains several edges that connect attention heads in the model's middle layers 8 to 15 at the position of the error ([answer-second]). To better understand the function of such attention heads, we compute their average attention scores over a set of input prompts. Figure~\ref{fig:consistency-heads} shows the average scores of the second attention head in layer 12 present in Qwen-2.5-1.5B's $\mathcal{C}_{\text{result}}^{(5/8)}$ circuit. Specifically, we visualize attention scores for four different sets of prompts: \emph{i)} prompts with an error at the position of the arithmetic result, \emph{ii)} prompts with an error at the position of the final numeric answer, \emph{iii)} prompts without errors, and \emph{iv)} prompts with a consistent error at \emph{\textbf{both}} the arithmetic result and the final numeric answer (e.g., ``\ldots $5+8=\textcolor{red!60!black}{16}$. Thus, Jane has $\textcolor{red!60!black}{16}$ apples.''). Two notable attention patterns emerge. For prompts where an error is present either at the position of the arithmetic result (e.g., ``\ldots $5+8=\textcolor{red!60!black}{16}$. Thus, Jane has $13$ apples.'') or at the position of the final numeric answer (e.g., ``\ldots $5+8=13$. Thus, Jane has $\textcolor{red!60!black}{16}$ apples.''), we observe high attention scores between the first digit of the arithmetic result ([result-first]) and the first digit of the final numeric answer ([answer-first]), but \emph{\textbf{not}} for the corresponding second digits ([result-second] and [answer-second], respectively). In contrast, for prompts \emph{without} errors or those with \emph{consistent} errors at both positions (e.g., ``\ldots $5+8=\textcolor{red!60!black}{16}$. Thus, Jane has $\textcolor{red!60!black}{16}$ apples.''), we observe high average attention scores for \textbf{\emph{both}} the first and second digits of the result and the final numeric answer. In essence, we observe that the attention head exhibits high average attention scores when the digits of the arithmetic result and the final numeric answer align. We refer to such attention heads as \emph{consistency heads}\textemdash{}attention heads that assess surface-level alignment of numerical values in the solution prompt. Notably, we find several consistency heads across \emph{\textbf{all}} models (see Figures~\ref{fig:attention-pattern-llama-consistency} and~\ref{fig:attention-pattern-phi-consistency} in the Appendix). Additionally, we observe \emph{(in)}consistency heads, which display high attention scores for numerical values that misalign (Figures~\ref{fig:attention-pattern-qwen-inconsistency} to~\ref{fig:attention-pattern-phi-inconsistency} in the Appendix).

\paragraph{Consistency Heads Govern Arithmetic Error Detection.} We hypothesize that \emph{consistency heads} in the lower to middle layers of the models play an important role in the arithmetic error detection process. This hypothesis has two major implications: \emph{i)} models may struggle to distinguish between samples \emph{\textbf{without}} errors and those containing a \emph{\textbf{consistent}} error at \emph{\textbf{both}} the position of the arithmetic result and the final numeric answer (e.g., ``\ldots $5+8=\textcolor{red!60!black}{16}$. Thus, Jane has $\textcolor{red!60!black}{16}$ apples.''); and \emph{ii)} a small subset of \emph{consistency heads} can significantly influence the models' arithmetic error detection behavior. To test our hypothesis, we first evaluate models on 1,000 (\emph{clean}, \emph{corrupt}) prompt pairs for each template, where \emph{clean} prompts contain a \emph{\textbf{consistent}} error at both error positions. As shown in Table~\ref{tab:both_behavioral_accuracy}, all models exhibit significant difficulties in detecting these type of errors. For instance, Qwen-2.5-1.5B achieves an average accuracy of only $12.39\% \pm 6.00\%$, indicating a strong bias toward labeling prompts with a consistent error as ``valid''. Even for Phi-3-Mini-4k-Instruct, the accuracy drops from about 89\% when detecting errors at either the arithmetic result or the final numeric answer (see Table~\ref{tab:behavioral_accuracy}) to $40.98\% \pm 18.41\%$ when both positions contain a \emph{\textbf{consistent}} error.

Next, we analyze the influence of individual consistency heads on the model’s error detection behavior through two complementary experiments. First, when Qwen-2.5-1.5B is given a prompt with a \emph{\textbf{consistent}} error at both positions (for which it incorrectly predicts ``valid''), we \emph{patch} the latent activations of \emph{\textbf{six}} consistency heads (see Table~\ref{tab:heads} in the Appendix for the exact heads) with the corresponding activations from a prompt containing a \emph{\textbf{single}} error at the arithmetic result (for which the model correctly predicts ``invalid''). If consistency heads indeed govern error detection, the model should change its initial prediction from ``valid'' to ``invalid'' under this intervention. Second, we perform the reverse: running the model on a single-error prompt (correctly predicted as ``invalid''), we patch the activations of the consistency heads with those from a consistent-error prompt. In this case, we would expect the intervention to reduce the error detection accuracy.

\begin{figure}[b!]
    \centering
    \includegraphics[width=0.45\textwidth]{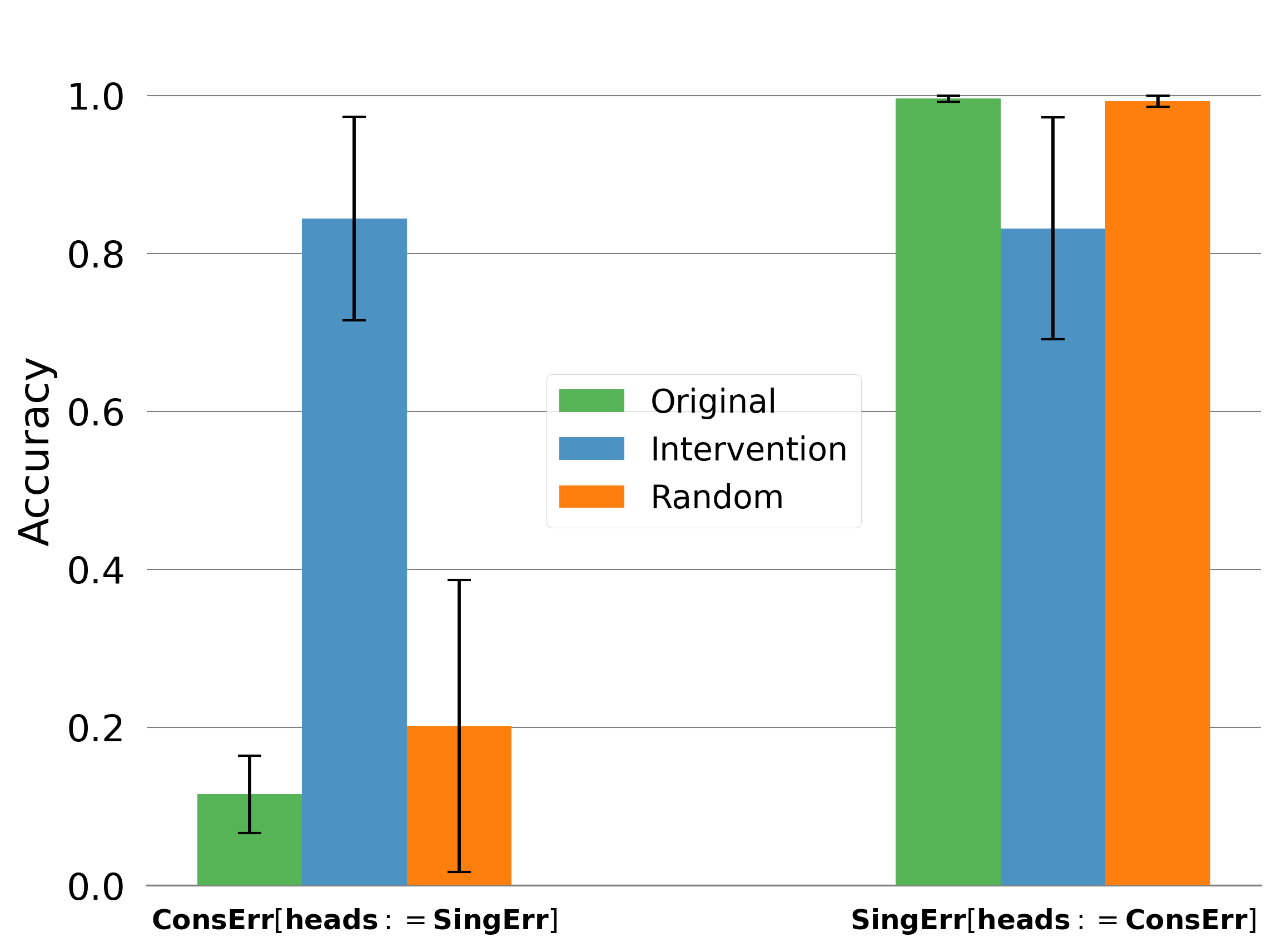}
    \caption{Error detection accuracy of Qwen-2.5-1.5B before and after patching six \emph{consistency heads}. On the left, we evaluate the model on consistent-error prompts and patch activations from a single-error run; on the right, we evaluate on single-error prompts and patch with activations from a consistent-error run. As a control, we patch six randomly chosen attention heads.}
    \label{fig:attention-patching}
\end{figure}

Figure~\ref{fig:attention-patching} shows the resulting changes in detection accuracy (predicting ``invalid'') for Qwen-2.5-1.5B. As expected, the first intervention $\left(\text{ConsErr}[\text{heads} := \text{SingErr}]\right)$ reliably flips the model’s prediction from ``valid'' to ``invalid,'' demonstrating that inconsistency signals injected via these heads can causally alter the model’s output. By contrast, the reverse intervention $\left(\text{SingErr}[\text{heads} := \text{ConsErr}]\right)$ changes the prediction from ``invalid'' to ``valid'' only in some cases. Similar results hold across \textbf{\textit{all}} models (see Figure~\ref{fig:attention-patching-full} in the Appendix). A plausible explanation might be that the model has more consistency heads than the ones we patch, which still signal an inconsistency for the given prompt. This touches upon the problem of completeness in circuit discovery, where circuits that can faithfully reproduce the model's task behavior are not guaranteed to include all components involved in the model’s task behavior~\cite{mondorf-etal-2025-circuit}. For details on the patching procedure, we refer to Appendix~\ref{app:consistency_patching_method}.

\begin{figure}[tbp]
    \centering
    \includegraphics[width=0.45\textwidth]{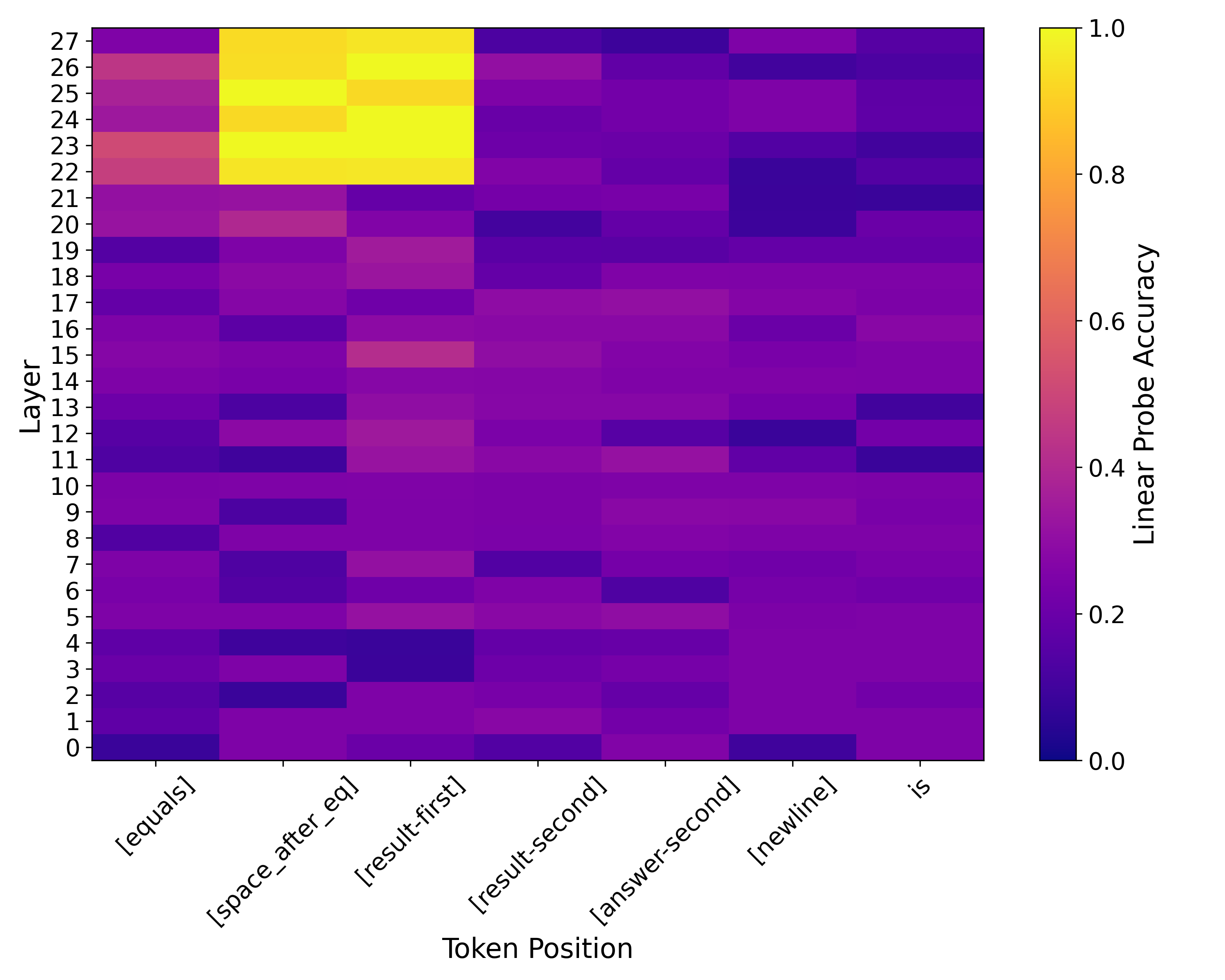}
    \caption{The linear probe's accuracy across all layers of Qwen-2.5-1.5B at selected token positions.}
    \label{fig:probing_qwen}
\end{figure}

\subsection{Dissociation of Arithmetic Validation and Computation}\label{subsec:Dissociation}
Our findings presented in Section~\ref{subsec:Decode_Error_Detection} suggest that the considered models tend to rely on surface-level consistency checks rather than re-evaluating the given arithmetic equation and comparing the result with the final numeric answer. Notably, we find that all models achieve 100\% accuracy in predicting the correct result of the arithmetic equation provided in each prompt (e.g., $5 + 8 = \text{?}$). However, we hypothesize that this correctly predicted result is not used for validation. To better understand the relationship between the models' arithmetic computation and validation procedures, we identify another set of circuits responsible for computing the correct arithmetic result at the position of the arithmetic equation, following a similar procedure as described in Section~\ref{sec:Circuits_Arithmetic_Error_Detection}. For further details on the identification process, please refer to Appendix~\ref{app:computation_circuit}. Interestingly, we find that for \emph{\textbf{all}} models, the identified arithmetic circuits predominantly contain edges in higher layers (due to space constraints, visualizations of these circuits are provided in Figures~\ref{fig:qwen_computation_circuit} to~\ref{fig:phi_computation_circuit} in the Appendix). This structural \emph{\textbf{dissociation}} between the circuits responsible for arithmetic computation and those involved in validation seems to explain the models' difficulties in detecting basic arithmetic errors. Specifically, although the models can successfully re-compute the result of a given arithmetic equation, the final arithmetic outcome is not fully encoded when the model checks for numeric consistency in middle layers. To support this hypothesis, we train a linear probe to predict the correct arithmetic result based on the hidden states of the model's residual stream (for training details, see Appendix~\ref{app:linear_probe}). Figure~\ref{fig:probing_qwen} shows the probe's accuracy across different layers of Qwen-2.5-1.5B and selected token positions. Only in the higher layers (after the consistency check) does the model's residual stream linearly encode information about the correct arithmetic result. Interestingly, similar patterns are observed for other models, too (see Figure~\ref{fig:probing-full} in the Appendix).

\begin{figure}[b!]
    \centering
    \includegraphics[width=0.45\textwidth]{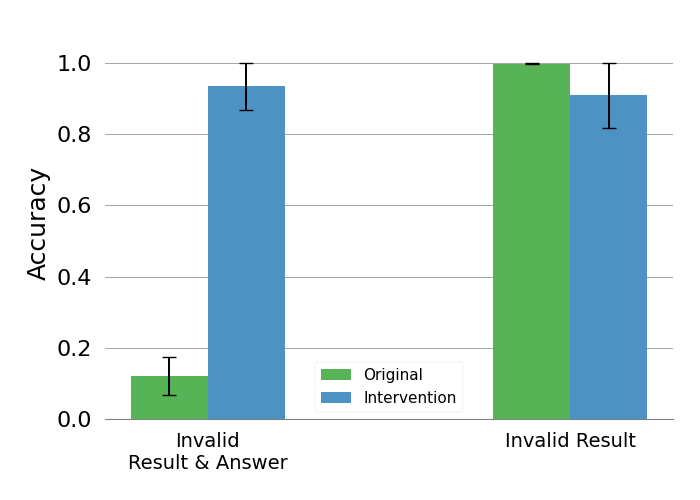}
    \caption{Error detection accuracy of Qwen-2.5-1.5B on (\emph{clean}, \emph{corrupt}) prompt pairs where (\emph{left}) the \emph{clean} prompt contains a consistent error at both error positions, and (\emph{right}) an error is present only at the position of the arithmetic result. The blue intervention bar denotes the result after adding the hidden representation of the residual stream in layer 22 (at [result-first]) to the residual stream of layer 1 (at [result-second]).}
    \label{fig:residual_patching_qwen}
\end{figure}

We demonstrate that by ``bridging'' the gap between arithmetic computation and validation, the model’s error detection capacity for prompts with consistent errors at both error positions can be significantly enhanced. Figure~\ref{fig:residual_patching_qwen} shows the error detection accuracy of Qwen-2.5-1.5B before and after we \emph{\textbf{add}} the hidden representation of the residual stream from layer 22 at token position [result-first] to the residual stream of layer 1 at token position [result-second] (essentially ``moving'' information from the top yellow layers in Figure~\ref{fig:probing_qwen} to lower layers). Notably, this approach improves the model's ability to detect \emph{\textbf{consistent}} errors by 81\%. Furthermore, accuracy on samples containing errors only at the position of the arithmetic result does not decline markedly, suggesting that the added representation enhances rather than overwrites existing information. A similar trend for other models is presented in Figure~\ref{fig:residual-patching-full} in the Appendix.

\begin{figure}[tbp]
    \centering
    \includegraphics[width=0.45\textwidth]{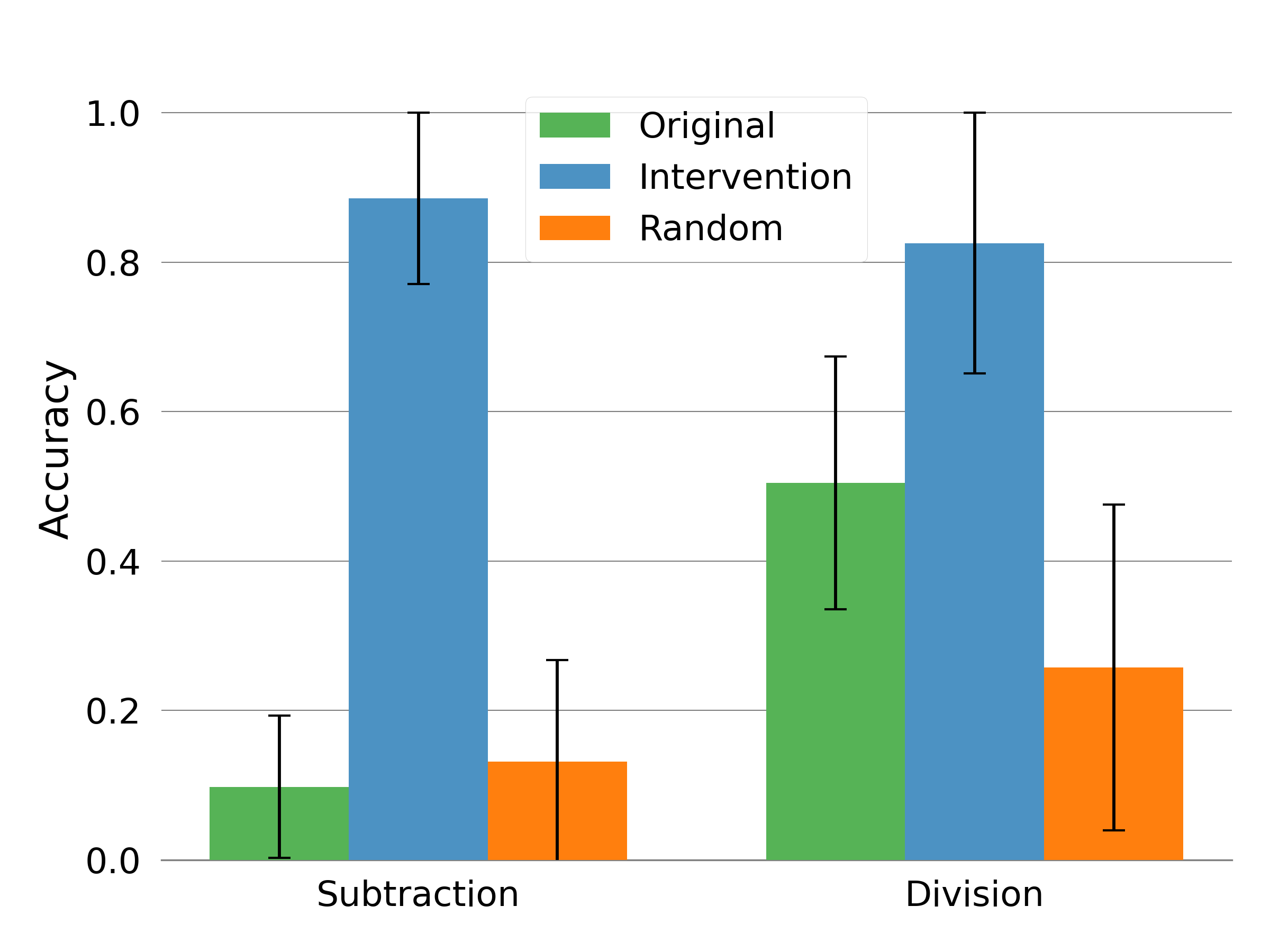}
    \caption{Error detection accuracy of Qwen-2.5-1.5B before and after patching six consistency heads. We evaluate the model on consistent-error prompts related to subtraction (\emph{left}) or division (\emph{right}) and patch activations from a single-error run. As a control, we patch six randomly chosen attention heads.}
    \label{fig:attention-patching-operations}
\end{figure}

\subsection{Consistency Heads in Other Arithmetic Operations}
Given the important role of consistency heads in our experiments on addition, we conduct further analyses to investigate whether their influence extends to other arithmetic operations. To this end, we generate new math word problems involving subtraction, multiplication, and division, using the same template structure with the two key error positions at the arithmetic result and the final numeric answer.\footnote{See Table~\ref{tab:operations_templates} in the Appendix for the full set of templates and variables considered when generating data for other arithmetic operations.} As for addition, we first evaluate the models' error detection accuracy on 1,000 prompt pairs per template. Similarly, we find that all models (except for Phi-3-Mini-4k-Instruct) encounter greater difficulty in detecting \emph{\textbf{consistent}} errors at both error positions compared to single-position errors. Complete results for all models and operations are reported in Table~\ref{tab:arithmetic_operations_performance} in the Appendix.

Next, for models that struggle with consistent errors, we replicate the consistency head patching experiment presented in Section~\ref{subsec:Decode_Error_Detection}, this time for prompts involving other arithmetic operations. Focusing on Qwen-2.5-1.5B, we evaluate the model on subtraction and division prompts with \emph{\textbf{consistent}} errors at both positions, where it incorrectly judged the reasoning as ``valid,'' and patch the latent activations of the consistency heads with those obtained from prompts containing an error only at the arithmetic result. Notably, across all operations, we patch the same consistency heads initially identified with prompts involving additions (Table~\ref{tab:heads} in the Appendix). Figure~\ref{fig:attention-patching-operations} shows the change in error detection accuracy, indicating that the causal role of consistency heads in error detection extends beyond addition. Similar effects are observed for Qwen-2.5-Math-1.5B and Llama-3.2-3B (Figure~\ref{fig:attention-patching-operations-full} in the Appendix).

Finally, we find that some larger models, such as Llama-3.1-70B, similarly struggle with detecting consistent arithmetic errors, as shown in Table~\ref{tab:big_model_behavioral_accuracy} in the Appendix. This may indicate that, to some extent, these models also rely on consistency heads\textemdash{}a direction that future work could explore.

\section{Related Work}\label{sec:Related_Work}
\paragraph{Self-correction in LLMs.} Self-correction in LLMs refers to the ability of models to correct their own generated output \citep{kamoi2024correctionsurvey, huang2024cannotselfcorrect, madaan2023selfrefine}. Recent studies \citep{tyen2024errorlocation, kamoi2024evaluatingdetecting} suggest that LLMs tend to struggle with \textit{intrinsic} self-correction, especially with \emph{detecting} errors in their own output~\citep{huang2024cannotselfcorrect, tyen2024errorlocation, kamoi2024correctionsurvey}. While most studies focus on improving the models' ability to self-correct~\citep{kamoi2024correctionsurvey, madaan2023selfrefine, chen2024selfdebug, zhao2023verifyandedit}, we study error detection from a mechanistic point of view.

\paragraph{Arithmetic and Error Detection in LLMs.} The underlying processes of arithmetic reasoning and error detection have been studied independently in LLMs so far. Several studies \citep{stolfo2023arithmetic, zhang2024arithmetic, nikankin2024bagofheuristics} use causal mediation analysis \citep{pearl2001effects} to identify circuits that account for how LLMs process arithmetic operations. As of now, only few studies have analyzed self-correction in LLMs beyond the models' generated output~\citep{DBLP:journals/corr/abs-2402-12563, liu2024moralselfcorrection}.

\section{Conclusion}\label{sec:Conclusion}
This paper presents a mechanistic analysis of arithmetic error detection in LLMs. Our findings reveal that smaller-sized LLMs heavily rely on \emph{consistency heads}\textemdash{}attention heads that evaluate surface-level alignment of numerical values in an arithmetic solution. Moreover, we highlight a structural dissociation between the models' arithmetic computation and validation processes. Finally, we show that bridging this gap can significantly improve the models' arithmetic error detection capacity.

\section{Limitations}\label{sec:Limitations}
While our study provides new insights into the mechanisms underlying arithmetic error detection in LLMs, several limitations exist that can be addressed by future work.

\paragraph{Task Design.} This study focuses on examining the error detection behavior of LLMs in the context of simple arithmetic tasks. Specifically, we analyze math word problems involving the addition of two single-digit numbers that yield a two-digit result, as described in Section~\ref{sec:Circuits_Arithmetic_Error_Detection}. Future research could extend these findings to other arithmetic operations, such as subtraction, multiplication, and division, or explore their applicability to more complex mathematical problems. It would also be valuable to investigate how these insights generalize to other domains, such as logical or causal reasoning tasks.

\paragraph{Model Selection.} As discussed in Sections~\ref{sec:Introduction} and~\ref{sec:Circuits_Arithmetic_Error_Detection}, our analysis is limited to four smaller-sized LLMs. Although we observe consistent patterns across various model architectures, sizes, and fine-tuning procedures (particularly within the mathematical domain), future research could investigate how these findings extend to larger models with more advanced arithmetic capabilities. Our behavioral experiments with models such as Llama-3.1-70B-Instruct and Qwen-2.5-32B-Instruct (see Table~\ref{tab:big_model_behavioral_accuracy} in the Appendix) show that even larger LLMs tend to struggle more with consistent errors at both error positions, compared to detecting an error present only at the position of the arithmetic result or the final answer. This may indicate that, similarly to smaller models, these models\textemdash{}at least to some extent\textemdash{}rely on \emph{consistency heads} that are susceptible to the validation gap. We believe this is a promising direction for future work to explore.

\paragraph{Circuit Identification Method.} As highlighted in Section~\ref{sec:Circuit_Analysis}, edge attribution patching~\citep{syed-etal-2024-attribution} serves as a linear approximation of activation patching~\citep{NEURIPS2020_92650b2e}. It involves a trade-off between accuracy and computational efficiency. Notably, circuits identified using EAP are not guaranteed to be \emph{complete}~\citep{wang2023interpretability}. Although the circuits identified in this study are highly sparse (comprising less than 0.1\% of the total edges) and achieve near-perfect task faithfulness (see Section~\ref{sec:Experiments}), future research should explore how these circuits compare to those identified through more exact methods.

\section*{Acknowledgments}
We would like to thank ELLIS\textemdash{}the European Laboratory for Learning and Intelligent Systems\textemdash{}as the collaboration that led to this paper started at a workshop organized by Raquel Fernández and Sandro Pezzelle at MFO, the Oberwolfach Research Institute for Mathematics in the German Black Forest, on behalf of the ELLIS NLP program. We are also grateful to the members of the MaiNLP lab for their valuable feedback, with special thanks to Michael Hedderich, Robert Litschko, Diego Frassinelli, Rob van der Goot, Siyao Peng, Yang Janet Liu, Xinpeng Wang, Verena Blaschke, Raoyuan Zhao, Elena Senger, Felicia Körner, Soh-Eun Shim, Andreas Säuberli, Florian Eichin, Shijia Zhou, and Beiduo Chen. We further thank the anonymous reviewers for their insightful comments and suggestions. RB acknowledges the generous support of Amazon Alexa, whose donation helped make this work possible. Finally, we acknowledge the support for BP through the ERC Consolidator Grant DIALECT 101043235.


\bibliography{custom}

\appendix

\section{Reproducibility Statement}
\label{app:reproducibility_statement}
To ensure the reproducibility of our experiments, we make all code publicly available at: \href{https://github.com/mainlp/validation-gap}{{https://github.com/mainlp/validation-gap}}. Details of our circuit identification process are described in Section~\ref{sec:Circuits_Arithmetic_Error_Detection}, Appendix~\ref{app:circuit_search} and~\ref{app:computation_circuit}. Documentation of our computational budget and the software we use can be found in Appendix~\ref{app:implementation_details}. Furthermore, a detailed account of the data used in this work is provided in Section~\ref{sec:Circuits_Arithmetic_Error_Detection} and Appendix~\ref{app:dataset}.

\section{Circuit Discovery Details}
\label{app:circuit}

\subsection{Edge Attribution Patching}
\label{app:eap}
Attribution patching \citep{nanda_attribution_patching}, and specifically \emph{edge attribution patching} (EAP) \citep{syed-etal-2024-attribution}, is a computationally efficient linear approximation of activation patching to estimate the effect of interventions on latent activations. Following the activation patching terminology proposed by~\citet{zhang2024towards}, consider a clean prompt $X_{clean}$ and a corrupted prompt $X_{corr}$. EAP approximates the change in a predefined metric $\mathcal{P}$ on the model's output when a specific activation $\vz$ is patched from its corrupted value $\vz(X_{corr})$ to its clean value $\vz(X_{clean})$. This approximation is formulated using a first-order Taylor expansion around the corrupted input $X_{corr}$. Specifically, EAP approximates $f_{\vz}(X_{corr}; \vz(X_{clean})) - f_{\vz}(X_{corr}; \vz(X_{corr}))$ as:

\begin{equation*}  
\begin{aligned}
& f_{\vz}(X_{corr}; \vz(X_{clean})) - f_{\vz}(X_{corr}; \vz(X_{corr}))  \\
& \quad \approx (\vz(X_{clean}) - \vz(X_{corr})) \cdot \frac{\partial f_{\vz}}{\partial \vz}\Bigr|_{\vz=\vz(X_{corr})}
\end{aligned}
\end{equation*}

where $f_{\vz}(X, \vz) = \mathcal{P}(M_{\vz}(X; \vz))$ represents the metric $\mathcal{P}$ applied to the patched model $M_{\vz}$. Here, $M_{\vz}(X; \vz)$ denotes the model $M$ where the activation $\vz$ is replaced with the value $\vz$ for input $X$.

To compute the gradient $\frac{\partial f_{\vz}}{\partial \vz}\Bigr|_{\vz=\vz(X_{corr})}$, a backward pass is performed on the corrupted input $X_{corr}$ with respect to the activation $\vz$. The absolute value of the resulting score, often referred to as the \emph{absolute attribution score} $|\nabla_{\vz} \mathcal{P}| = |(\vz(X_{clean}) - \vz(X_{corr})) \cdot \frac{\partial f_{\vz}}{\partial \vz}\Bigr|_{\vz=\vz(X_{corr})}|$, quantifies the estimated influence of patching activation $\vz$. This facilitates efficient circuit identification in EAP by ranking edges according to these scores.

\subsection{Faithfulness Metric}
\label{app:faithfulness}
Let \( \left( X_{\text{clean}}, X_{\text{corr}} \right)_i \) represent a pair of \textit{clean} and \textit{corrupt} prompts within a dataset of size $N$. For each input, let \( \mathcal{C}(X_{i,clean}) \) and \( \mathcal{C}(X_{i,corr}) \) denote the logits of the clean and corrupt answer tokens in the circuit's output, and let \( M(X_{i,clean}) \) and \( M(X_{i,corr}) \) be the corresponding logits for the full model. In this study, we define the faithfulness as the \emph{logit difference recovered}:

\begin{equation}\label{eq:faithfulness}
\frac{1}{N} \sum_{i=1}^{N} \left( \frac{\mathcal{C}(X_{i,clean}) - \mathcal{C}(X_{i,corr})}{M(X_{i,clean}) - M(X_{i,corr})} \times 100 \right)
\end{equation}

A faithfulness score of $100\%$ indicates that the circuit preserves the same logit difference as the full model, thus effectively recovering the model's task behavior.

\subsection{Circuit Identification Process}
\label{app:circuit_search}
To identify a minimal set of edges whose circuit achieves a faithfulness score between 99\% and 101\%, we employ an iterative search process. Starting from the sorted \emph{absolute attribution scores}, denoted by $|\nabla_{\vz} \mathcal{P}|$, we initially select the top-\( k \) edges and evaluate the corresponding circuit. In each iteration, we then add the next \( n \) edges from this sorted list and re-evaluate the faithfulness of the resulting circuit. The search stops as soon as a circuit with a faithfulness score (see Equation~\ref{eq:faithfulness}) within the desired interval (99\% to 101\%) is found. In our experiments, we set \( k = 100 \) and \( n = 20 \). 

\begin{table*}[t!]
    \centering
    \begin{tabular}{lccccccc}
        \toprule
        Model & Parameters & Layers & Hidden Dim & Num Heads & License \\
        \midrule
        {\small\href{https://huggingface.co/Qwen/Qwen2.5-1.5B-Instruct}{Qwen2.5-1.5B-Instruct}} & 1.54B & 28 & 1536 & 12 & \href{https://huggingface.co/Qwen/Qwen2.5-1.5B-Instruct/blob/main/LICENSE}{{\small Apache}}\\
        {\small\href{https://huggingface.co/Qwen/Qwen2.5-Math-1.5B-Instruct}{Qwen2.5-Math-1.5B-Instruct}} & 1.54B & 28 & 1536 & 12 & \href{https://huggingface.co/Qwen/Qwen2.5-Math-1.5B-Instruct/blob/main/LICENSE}{{\small Apache}}\\
        {\small\href{https://huggingface.co/meta-llama/Llama-3.2-3B-Instruct}{Llama-3.2-3B-Instruct}} & 3.21B & 28 & 3072 & 24 & \href{https://huggingface.co/meta-llama/Llama-3.2-3B-Instruct/blob/main/LICENSE.txt}{{\small Llama 3.2}}\\
        {\small\href{https://huggingface.co/microsoft/Phi-3-mini-4k-instruct}{Phi-3-Mini-4k-Instruct}} & 3.82B & 32 & 3072 & 32 & \href{https://huggingface.co/microsoft/Phi-3.5-mini-instruct/blob/main/LICENSE}{{\small MIT}}\\
        \bottomrule
    \end{tabular}
    \caption{Properties of the models studied in this work. provide details on the number of parameters, layers, the hidden dimension size of the residual stream, and the number of attention heads for each model. Model weights were obtained from their respective Hugging Face repositories, accessible via the model names listed in the tables. Additionally, we specify the licenses under which the models are distributed.}
    \label{tab:model_comparison}
\end{table*}

\section{Dataset}
\label{app:dataset}

\subsection{Templates}
We generate our dataset of \textit{clean} and \textit{corrupt} prompts based on the templates in Table~\ref{tab:templates}. These templates have a set of variables, namely \texttt{[instruction]}, \texttt{[person]}, \texttt{[object]}, \texttt{[verb]}, \texttt{[pronoun]}, \texttt{[num1]}, \texttt{[num2]}, \texttt{[num3]}, each of which can be assigned different values. Table~\ref{tab:template_variables} lists all possible values. For the numerical variables, we use single-digit numerical values (\texttt{[num1]} and \texttt{[num2]}) that add to a two-digit arithmetic result (\texttt{[num3]}). To ensure that each variable assignment occupies the same position in the token sequence within a template, we retain only variables that are tokenized as a single token for each model. For the \texttt{[instruction]} variable, we include only instructions that share the same number of tokens. Finally, for the \texttt{[correct\_pair]} variable, we select assignments where labels are tokenized as a single token across models.

For other operations (subtraction, multiplication, and division) we follow the same procedure and construct eight templates for each operation (see Table~\ref{tab:operations_templates}). These templates use the same variable structure as addition, with subtraction including also a \texttt{[verb]} variable. For the numerical variables, we restrict number pairs to ensure two-digit results: subtraction uses 2-digit minus 1-digit inputs yielding a 2-digit result, multiplication uses 2-digit times 1-digit inputs yielding a 2-digit result, and division uses 2-digit dividends and 1-digit divisors producing 2-digit quotients.

\subsection{Aligning Token Positions Across Templates}\label{app:align_token_pos}
Since we employ token-specific EAP to identify relevant edges at specific token positions, the same element (e.g., the arithmetic result or the final numeric answer token) might appear at different token positions depending on the specific template (see Table~\ref{tab:templates}). This variation in token positions makes it difficult to determine whether edges from two different template-specific circuits appear at semantically similar tokens (e.g., the arithmetic result in template 1 at token position 13 and the arithmetic result in template 2 at token position 16). To address this challenge, we assign shared abstract labels to corresponding elements across templates. Examples of such labels include \texttt{[op1-in-eq]}, \texttt{[op2-in-eq]}, \texttt{[equals]}, \texttt{[result-first]}, \texttt{[result-second]}, \texttt{[answer-first]}, and \texttt{[answer-second]}, which represent the two operands of the addition, the equal sign, and the digits of the arithmetic result and the numeric answer, respectively. Mapping tokens to a shared set of labels enables us to compute the \emph{soft intersection circuits} between templates~\textemdash{}allowing for the comparison of circuits associated with semantically equivalent elements without being confounded by their varying positions within the sequence.

\section{Experiment Details}
\label{app:exp_details}

\begin{table*}[tbp]
    \centering
    \small
    \begin{tabular}{llp{0.6\textwidth}}
        \toprule
         & \textbf{Model} & \textbf{Attention Heads} \\
        \midrule
        \multirow{4}{*}{\rotatebox{90}{\textbf{Consistency}}} & Qwen-2.5-1.5B-Instruct & L12H0, L12H2, L12H10, L13H0, L13H1, L13H10 \\
         \cmidrule{2-3}
         & Qwen-2.5-Math-1.5B-Instruct & L13H0, L13H1 \\
         \cmidrule{2-3}
         & Llama-3.2-3B-Instruct & L4H14, L5H3, L7H11, L8H1, L10H5, L10H18 \\
         \cmidrule{2-3}
         & Phi-3-Mini-4k-Instruct & L10H1, L10H5, L10H14, L14H19, L16H18 \\         
        \midrule
        \multirow{4}{*}{\rotatebox{90}{\textbf{Random}}} & Qwen-2.5-1.5B-Instruct & L1H0, L4H9, L9H6, L10H5, L11H8, L27H9 \\
         \cmidrule{2-3}
         & Qwen-2.5-Math-1.5B-Instruct & L4H9, L27H5 \\
         \cmidrule{2-3}
         & Llama-3.2-3B-Instruct & L1H1, L4H19, L9H16, L10H15, L11H23, L27H12 \\
         \cmidrule{2-3}
         & Phi-3-Mini-4k-Instruct & L0H25, L3H18, L8H25, L20H19, L23H28 \\ 
        \bottomrule
    \end{tabular}
    \caption{Attention heads used in the patching experiment. \emph{Consistency heads} refer to attention heads that demonstrate a behavior consistent with the pattern shown in Figure~\ref{fig:consistency-heads}, assessing numerical alignment between the digits of the arithmetic result and the final numeric answer. In contrast, \emph{random heads} are arbitrarily selected attention heads not classified as consistency heads, serving as a control group for comparison in the experiment.}
    \label{tab:heads}
\end{table*}

\subsection{Models}\label{app:models}
Details of the models used in this study are presented in Table~\ref{tab:model_comparison}. Specifically, we include information on the number of parameters, the number of layers, the size of the hidden model dimension, the number of attention heads per attention block, and the respective model licenses. All models are instruction-tuned and expect a series of special tokens (e.g., to indicate the beginning of the user prompt or the end of a turn). Therefore, we wrap all prompts in the respective chat templates of the models\footnote{\href{https://huggingface.co/docs/transformers/main/chat_templating}{huggingface.co/docs/transformers/chat\_templating}.}. When applicable, we use the models' default system prompts.

\subsection{Edge Overlap}\label{app:method_edge_overlap}
To quantify the proportion of shared edges between two circuits, $\mathcal{C}_1$ and $\mathcal{C}_2$, we compute both the Intersection over Union (IoU) and the Intersection over Minimum (IoM).

As discussed in Section~\ref{sec:Circuits_Arithmetic_Error_Detection}, we employ token-level EAP to identify relevant edges for each token position $t$ in the prompt. Therefore, the set of edges $e_{ij}^{(t)} \in \mathcal{E = \mathcal{C}}$ is determined by the specific token position $t$. When computing the IoU and IoM between two circuits, the intersection and union of edge sets are computed separately for each token position. Specifically, we define the two metrics as:

\begin{equation}
    \mathrm{IoU}(\mathcal{C}_{\text{result}}, \mathcal{C}_{\text{answer}}) = \frac{|\mathcal{C}_{\text{result}} \cap \mathcal{C}_{\text{answer}}|}{|\mathcal{C}_{\text{result}} \cup \mathcal{C}_{\text{answer}}|},
\end{equation}

 \begin{equation}
     \mathrm{IoM}(\mathcal{C}_{\text{result}}, \mathcal{C}_{\text{answer}}) = \frac{|\mathcal{C}_{\text{result}} \cap \mathcal{C}_{\text{answer}}|}{\min(|\mathcal{C}_{\text{result}}|,\, |\mathcal{C}_{\text{answer}}|)}
 \end{equation}

This provides an efficient way of measuring the degree of edge overlap at each token position in the analyzed circuits.

\subsection{Circuits for Arithmetic Computation}\label{app:computation_circuit}
To gain a deeper understanding of the relationship between models' mechanisms for arithmetic computation and validation, we identify an additional set of circuits responsible for correctly computing the arithmetic result at the position of the equation (e.g., ``5 + 8 = 13''). This process involves generating a new set of (\textit{clean}, \textit{corrupt}) prompt pairs for each template $\mathcal{T}_i \in \{\mathcal{T}_1, \ldots, \mathcal{T}_8\}$.

We construct these datasets based on the data samples used for identifying circuits for arithmetic error detection (as described in Section~\ref{sec:Circuits_Arithmetic_Error_Detection}). Specifically, we first truncate both \textit{clean} and \textit{corrupt} prompts at the position of the equation sign (e.g., ```...5 + 8 =''). Next, we modify the \textit{corrupt} prompt by replacing the numbers with a different set of numbers that produces an alternative result (e.g., ``...3 + 9 =''). This process results in two prompts\textemdash{}neither containing any errors yet\textemdash{}where the next token is expected to be the correct outcome of the arithmetic equation (e.g., ``13'' for the \textit{clean} prompt and ``12'' for the \textit{corrupt} prompt). The corresponding labels represent the correct results for each prompt.

Using the new sets of \textit{clean} and \textit{corrupt} prompt pairs, we aim to identify the model components involved in computing the correct arithmetic result. We follow the same steps described in Section~\ref{sec:Circuits_Arithmetic_Error_Detection} and Appendix~\ref{app:circuit_search} to identify circuits for each template. Finally, we compute the corresponding \textit{soft intersection circuits}, which are responsible for generating the correct arithmetic result across templates.

\subsection{Patching Consistency Heads}\label{app:consistency_patching_method}
To evaluate the influence of individual consistency heads on the models' error detection behavior, we conduct two complementary patching interventions. In the first intervention, we run models on prompts $X_{both}$, which contain a consistent error at \emph{both} the arithmetic result and the final numeric answer (for which models tend to incorrectly predict ``valid''), and patch the latent activations of the consistency heads listed in Table~\ref{tab:heads} with activations from prompts $X_{result}$, where the error appears only at the arithmetic result (for which models typically predict ``invalid''). This intervention is expected to increase the rate of ``invalid'' predictions. In the second, reverse intervention, we run models on prompts $X_{result}$ and patch the same heads with activations from $X_{both}$. This setup is expected to decrease the rate of ``invalid'' predictions. 

Specifically, for an attention head $h$, let $A^h(X)$ denote its attention matrix for a given prompt $X$. The patching operation we perform is defined as $A^h(X_{target}) = \alpha \cdot A^h(X_{source})$, where $\alpha$ is a scaling factor that controls the influence of the patched activation. For the first intervention, we set $\alpha$ to $3.1$ for Qwen-2.5-1.5B-Instruct and Llama-3.2-3B-Instruct, and to $3.3$ and $3.4$ for Qwen-2.5-Math-1.5B-Instruct and Phi-3-Mini-4k-Instruct, respectively. For the second intervention, we set $\alpha = 1.0$ for all models. We perform the patching over $1,000$ prompt pairs. As a control setup for this experiment, we compare the result to patching randomly selected attention heads that are not labeled as consistency heads (see the full list in Table~\ref{tab:heads}).

\subsection{Linear Probes}\label{app:linear_probe}
We train linear probes on the hidden states of the models' residual stream to test whether a specific layer linearly encodes information about the correct result of the arithmetic equation within the prompts described in Section~\ref{sec:Circuits_Arithmetic_Error_Detection}. The probes are trained separately for each layer and a set of selected token positions. For each layer and token position, we use a training set of 500 hidden states per template (i.e., 4,000 samples in total per layer and token position) and a test set of 100 samples per template (i.e., 800 samples in total). The hidden states are collected from prompts where both the arithmetic result and the numeric answer contain consistent errors. All probes are trained for one epoch using the Adam optimizer with a learning rate of $0.001$.

\section{Additional Results}
\label{app:complete_results}
In this section, we present the results of additional experiments we conducted.

\subsection{Error Detection Circuits}\label{app:error_detection_circuits}
As described in Section~\ref{subsec:results_error_detection_circuits}, we identify a circuit with faithfulness score between 99\% and 101\% for each template $\mathcal{T}_i \in \{\mathcal{T}_1, \ldots \mathcal{T}_8\}$. Table~\ref{tab:faith_size} provides a comparison between the size of the identified circuits, their exact faithfulness scores, and the total number of edges in the full computational graph for all models and templates. 

Once a circuit is identified for each template, we compute the \emph{soft intersection circuit} $\mathcal{C}^{(\tau)}$ to derive a final circuit that generalizes across templates, as described in Section~\ref{sec:Circuits_Arithmetic_Error_Detection}. For each model, we analyze the faithfulness scores and edge counts for different threshold values $\tau$ in the soft intersection circuit $\mathcal{C}^{(\tau)}$.

\begin{figure*}[htbp]
    \centering
    \begin{subfigure}[b]{0.32\textwidth}
        \centering
        \includegraphics[width=\linewidth]{figures/faithfulness/z1/qwen_intersection_faithfulness_z1_template_intersection_gradfunc_logit_ansfunc_avg_diff_train_size_5000_marked.pdf}
        \caption{Invalid Result}
        \label{fig:soft_qwen_result}
    \end{subfigure}
    \hfill
    \begin{subfigure}[b]{0.32\textwidth}
        \centering
        \includegraphics[width=\linewidth]{figures/faithfulness/z2/qwen_intersection_faithfulness_z2_template_intersection_gradfunc_logit_ansfunc_avg_diff_train_size_5000_marked.pdf}
        \caption{Invalid Answer}
        \label{fig:soft_qwen_answer}
    \end{subfigure}
    \hfill
    \begin{subfigure}[b]{0.32\textwidth}
        \centering
        \includegraphics[width=\linewidth]{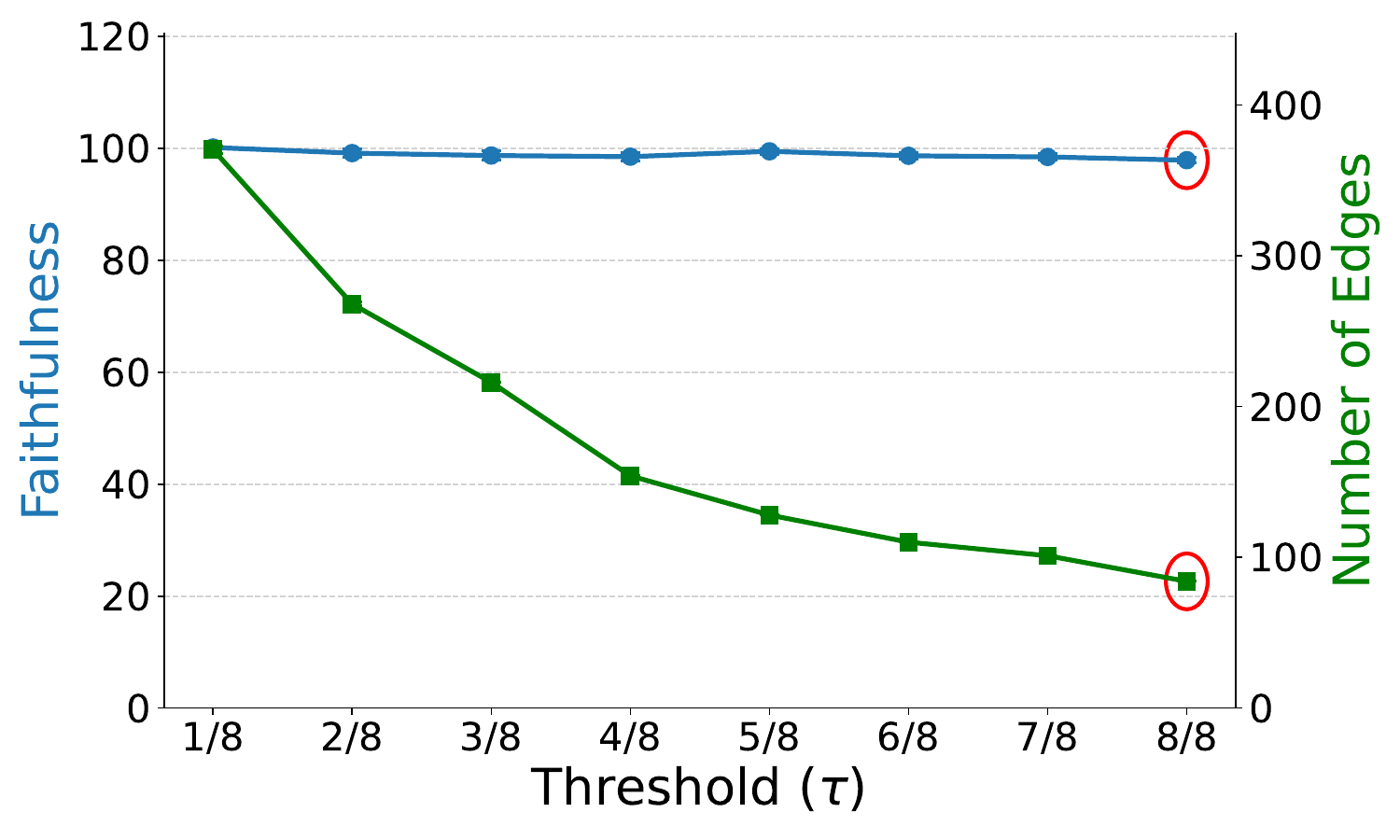}
        \caption{Computation}
        \label{fig:soft_qwen_computation}
    \end{subfigure}
    
    \caption{The number of edges and average faithfulness scores of the \emph{soft intersection circuit} for different threshold values, $\tau$. Red circles indicate the soft intersection circuit that best trade offs size with faithfulness. Results are shown for Qwen-2.5-1.5B-Instruct.}
    \label{fig:faithfulnes_qwen_full}
\end{figure*}

\begin{figure*}[tbp]
  \centering
  \rotatebox{90}{
  \includegraphics[width=0.9\textheight]{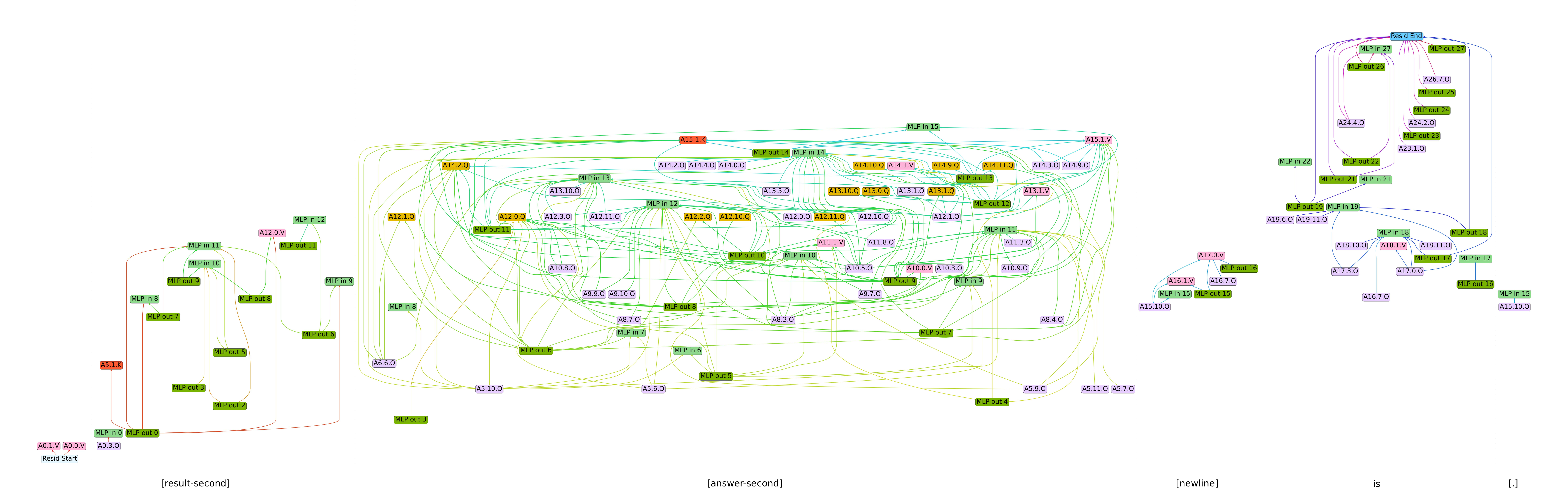}
  }
  \caption{The arithmetic result error identification circuit $\mathcal{C}_{\text{result}}^{(5/8)}$ of Qwen-2.5-1.5B-Instruct obtained after taking the soft intersection between all template circuits with a threshold value of $\tau = \frac{5}{8}$.}
  \label{fig:qwen_z1_circuit}
\end{figure*}

\begin{figure*}[tbp]
  \centering
  \includegraphics[width=\textwidth]{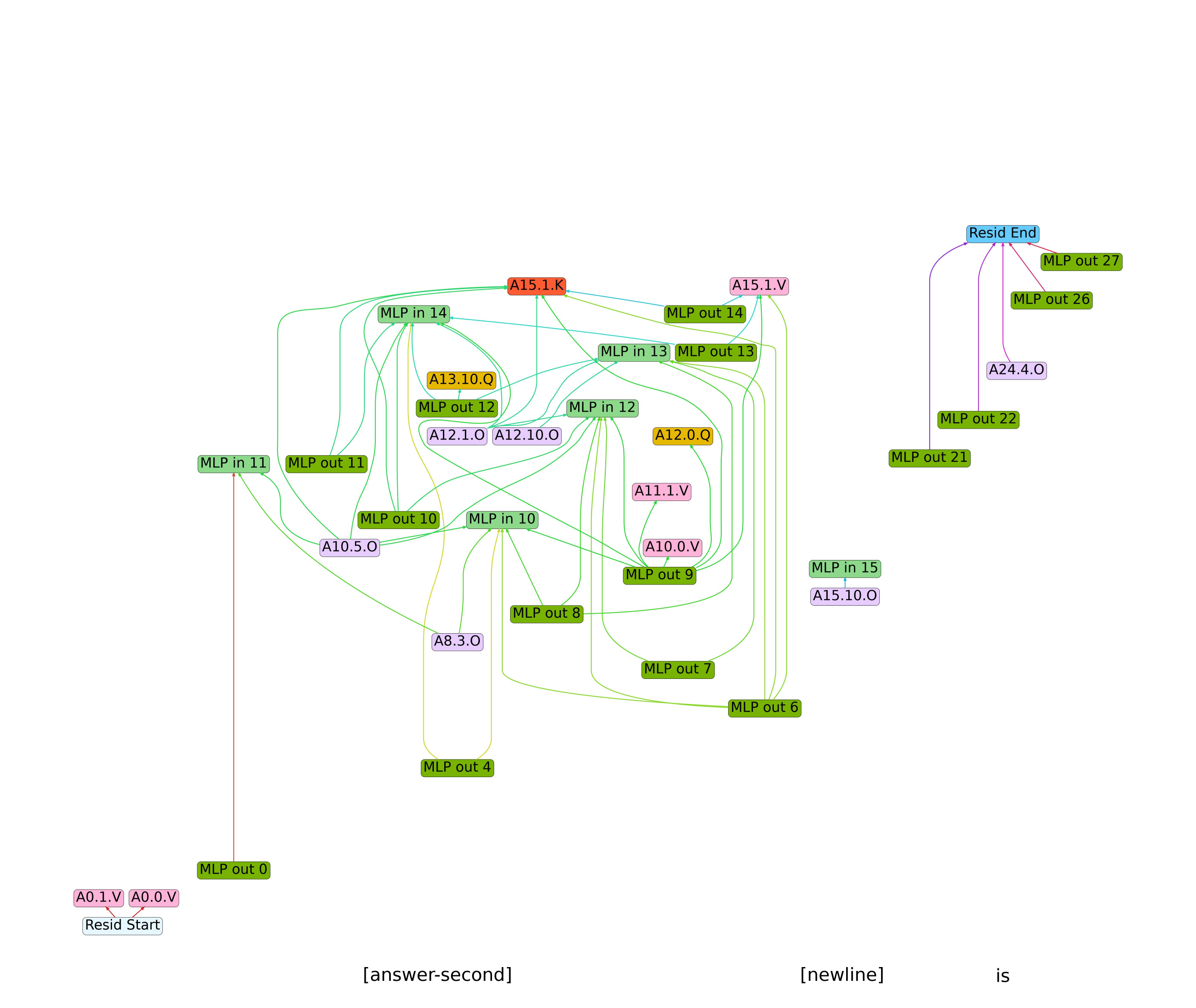}
  \caption{The numeric answer error identification circuit $\mathcal{C}_{\text{answer}}^{(8/8)}$ of Qwen-2.5-1.5B-Instruct obtained after taking the soft intersection between all template circuits with with a threshold value of $\tau = \frac{8}{8}$.}
  \label{fig:qwen_z2_circuit}
\end{figure*}

Figure~\ref{fig:faithfulnes_qwen_full} illustrates the faithfulness scores and number of edges for the \emph{soft intersection circuit} $\mathcal{C}^{(\tau)}$ of Qwen-2.5-1.5B-Instruct across various threshold values $\tau$. Specifically, Figure~\ref{fig:soft_qwen_result} displays values for the circuit responsible for detecting errors at the position of the arithmetic results, while Figure~\ref{fig:soft_qwen_answer} shows values for an error at the final numeric answer. The red circles indicate the circuits that offer the best balance between faithfulness and size. For detecting arithmetic errors at the position of the arithmetic result, the optimal threshold is $\tau = \frac{5}{8}$, while for detecting errors at the final numeric answer, the strict intersection at $\tau = \frac{8}{8}$ provides the best trade-off. The corresponding circuits are visualized in Figures~\ref{fig:qwen_z1_circuit} and~\ref{fig:qwen_z2_circuit}, respectively.

\begin{figure*}[tbp]
  \centering
  \includegraphics[width=\textwidth]{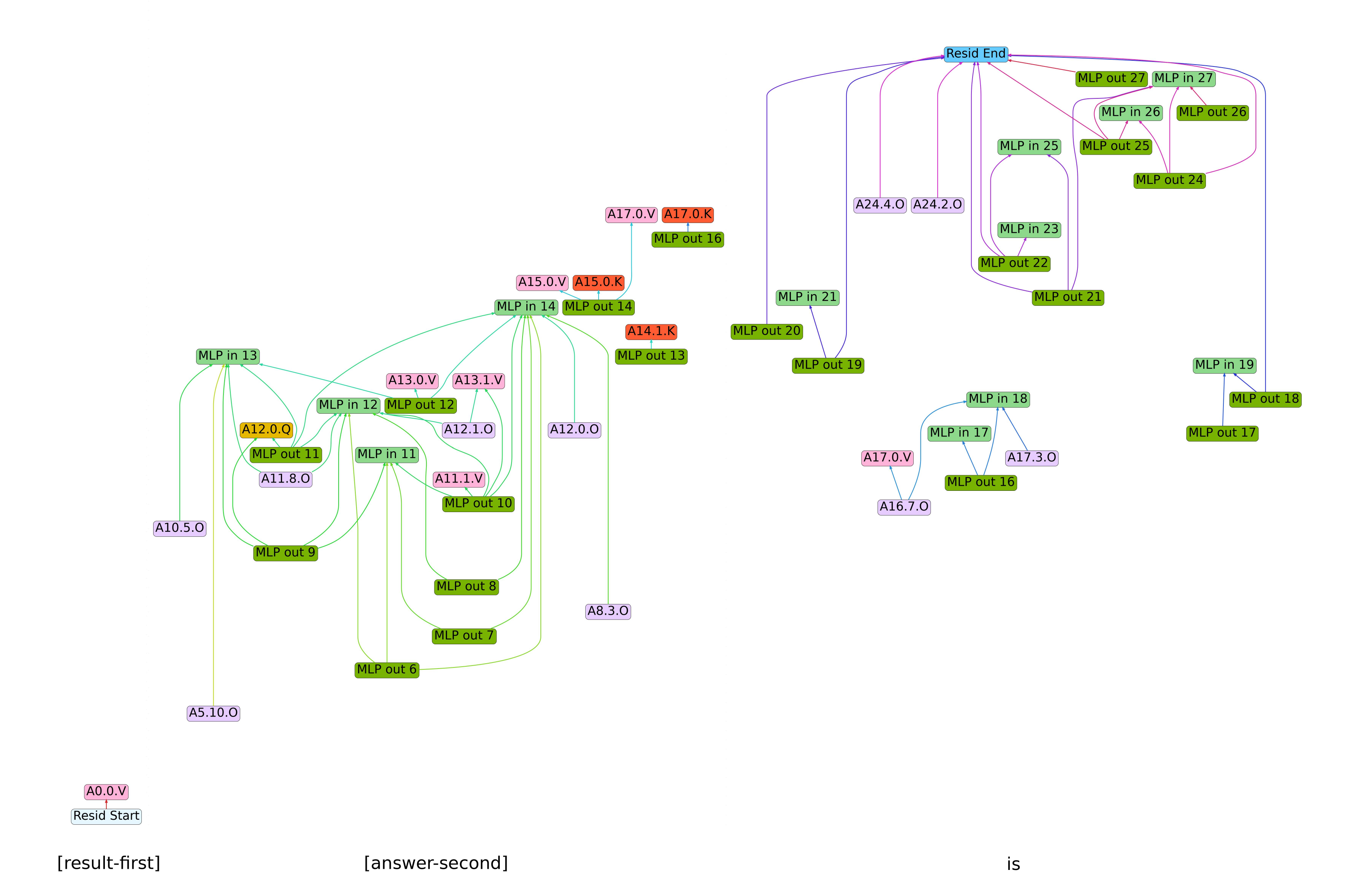}
  \caption{The arithmetic result error identification circuit $\mathcal{C}_{\text{result}}^{(7/8)}$ of Qwen-2.5-Math-1.5B-Instruct obtained after taking the soft intersection between all template circuits with a threshold value of $\tau = \frac{7}{8}$.}
  \label{fig:qwen_math_z1_circuit}
\end{figure*}

\begin{figure*}[tbp]
  \centering
  \rotatebox{90}{
  \includegraphics[width=0.925\textheight]{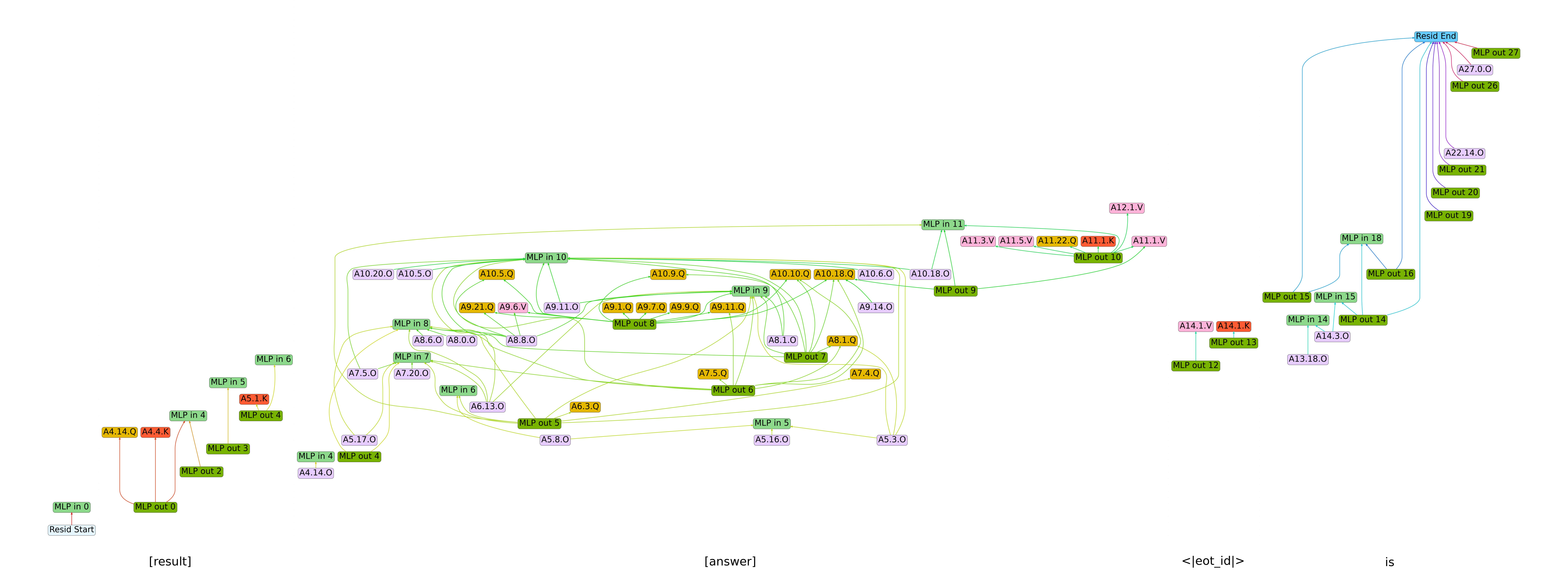}
  }
  \caption{The arithmetic result error identification circuit $\mathcal{C}_{\text{result}}^{(7/8)}$ of Llama-3.2-3B-Instruct obtained after taking the soft intersection between all template circuits with a threshold value of $\tau = \frac{7}{8}$.}
  \label{fig:llama_z1_circuit}
\end{figure*}

\begin{figure*}[htbp]
  \centering
  \rotatebox{90}{
  \includegraphics[width=0.9\textheight]{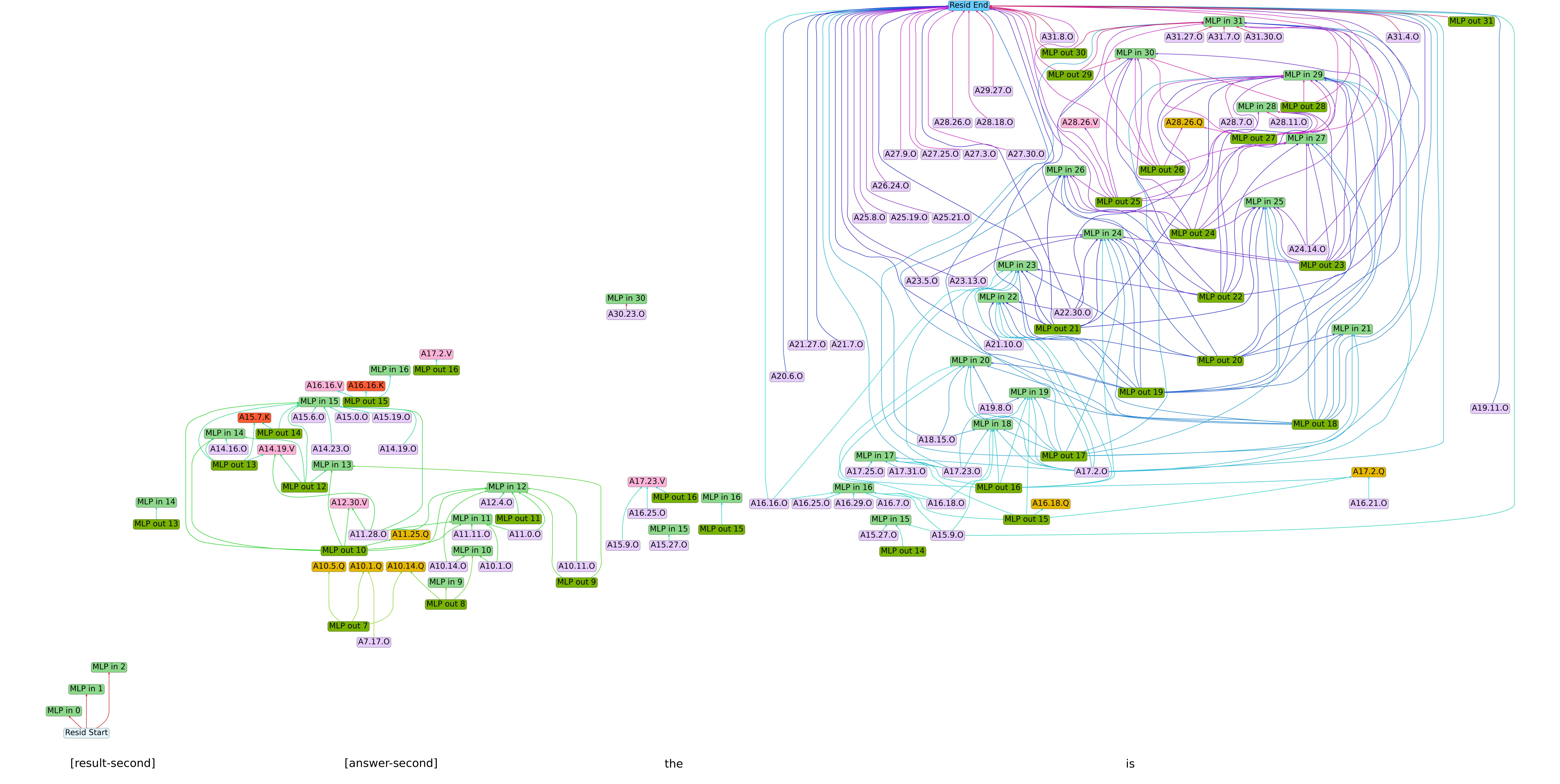}
  }
  \caption{The arithmetic result error identification circuit $\mathcal{C}_{\text{result}}^{(7/8)}$ of Phi-3-Mini-4k-Instruct obtained after taking the soft intersection between all template circuits with a threshold value of $\tau = \frac{7}{8}$.}
  \label{fig:phi_z1_circuit}
\end{figure*}

\begin{figure*}[tbp]
  \centering
  \includegraphics[width=\textwidth]{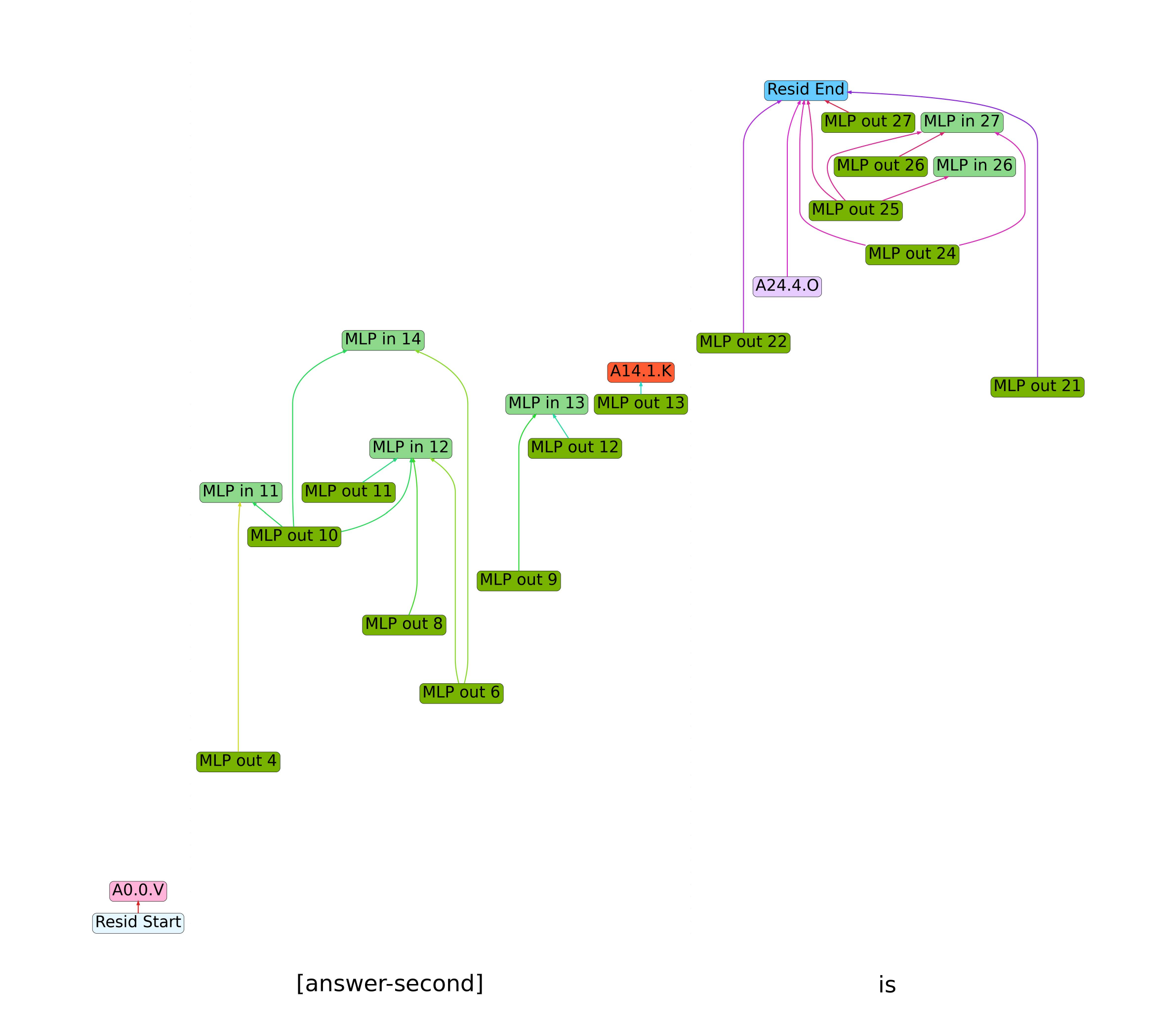}
  \caption{The numeric answer error identification circuit $\mathcal{C}_{\text{answer}}^{(8/8)}$ of Qwen-2.5-Math-1.5B-Instruct obtained after taking the soft intersection between all template circuits with with a threshold value of $\tau = \frac{8}{8}$.}
  \label{fig:qwen_math_z2_circuit}
\end{figure*}

\begin{figure*}[tbp]
  \centering
  \rotatebox{90}{
  \includegraphics[width=0.925\textheight]{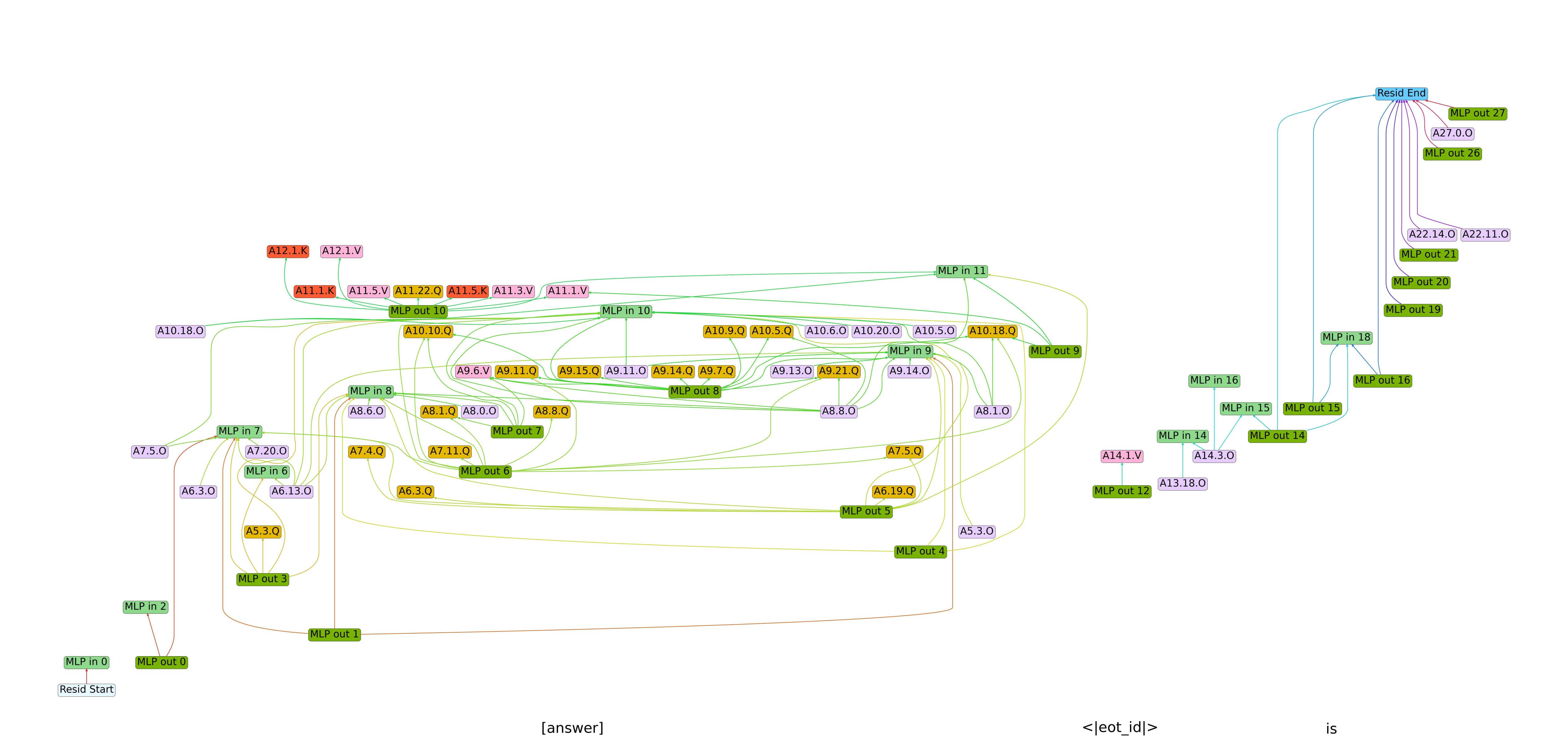}
  }
  \caption{The numeric answer error identification circuit $\mathcal{C}_{\text{answer}}^{(8/8)}$ of Llama-3.2-3B-Instruct obtained after taking the soft intersection between all template circuits with with a threshold value of $\tau = \frac{8}{8}$.}
  \label{fig:llama_z2_circuit}
\end{figure*}

\begin{figure*}[htbp]
  \centering
  \rotatebox{90}{
  \includegraphics[width=0.925\textheight]{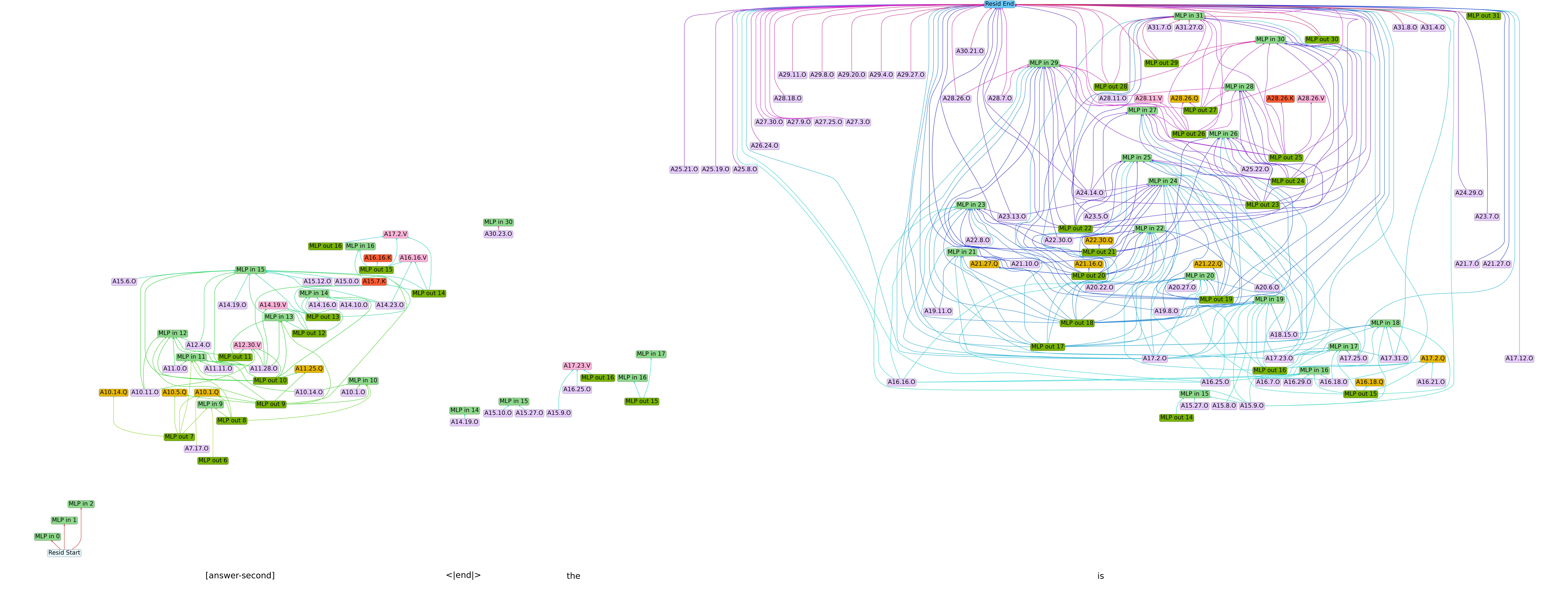}
  }
  \caption{The numeric answer error identification circuit $\mathcal{C}_{\text{answer}}^{(6/8)}$ of Phi-3-Mini-4k-Instruct obtained after taking the soft intersection between all template circuits with with a threshold value of $\tau = \frac{6}{8}$.}
  \label{fig:phi_z2_circuit}
\end{figure*}

Similar results for Qwen-2.5-Math-1.5B-Instruct are shown in Figure~\ref{fig:faithfulnes_qwen-math_full}, where the optimal circuits are $\mathcal{C}_{\text{result}}^{(7/8)}$ and $\mathcal{C}_{\text{answer}}^{(8/8)}$, depicted in Figures~\ref{fig:qwen_math_z1_circuit} and~\ref{fig:qwen_math_z2_circuit}. For Llama-3.2-3B-Instruct, the corresponding results are displayed in Figure~\ref{fig:faithfulnes_llama_full}, with optimal circuits $\mathcal{C}_{\text{result}}^{(7/8)}$ and $\mathcal{C}_{\text{answer}}^{(8/8)}$, visualized in Figures~\ref{fig:llama_z1_circuit} and~\ref{fig:llama_z2_circuit}. Finally, Figure~\ref{fig:faithfulnes_phi_full} provides the results for Phi-3-Mini-4k-Instruct, where the best soft intersection circuits are $\mathcal{C}_{\text{result}}^{(7/8)}$ and $\mathcal{C}_{\text{answer}}^{(6/8)}$, shown in Figures~\ref{fig:phi_z1_circuit} and~\ref{fig:phi_z2_circuit}.

Across all models and error types, a consistent structural pattern emerges. The most relevant edges are concentrated in the middle layers at the position of the final numeric answer. Additionally, a smaller subset of edges appears in the higher layers at the final token position of the prompt, primarily connecting MLP layers with the final residual output. Phi-3-Mini-4k-Instruct exhibits a slight variation, displaying a larger set of edges in the higher layers at the final token position. While many of these edges involve MLP components, others include attention head output matrices. Both types contribute to the residual stream forming the model’s final output. Nonetheless, this model also exhibits a concentration of edges in the middle layers at the numeric answer position, consistent with the overall pattern observed in other models.

\subsection{Edge Overlap of Error Detection Circuits}\label{app:results_edge_overlap}
Table~\ref{tab:iou_iom} shows the the Intersection over Union (IoU) and Intersection over Minimum (IoM) between the error detection circuits $ \mathcal{C}_{\text{result}} $ and  $\mathcal{C}_{\text{answer}} $ for each model. Notably, the IoM between the two circuits remains consistently $\geq 0.75$ across all models. Meanwhile, IoU values exhibit greater variability, ranging from $0.20$ for Qwen-2.5-1.5B-Instruct to $0.80$ for Phi-3-Mini-4k-Instruct. This variability is primarily attributable to the size differences between circuits. Specifically, the circuits responsible for detecting errors at the position of the final numeric answer are generally smaller, with particularly pronounced size reductions for the Qwen family of models (refer to Table~\ref{tab:faith_size}).

\begin{table}[tbp]
\centering
\begin{tabular}{@{}lcc@{}}
    \toprule
    \textbf{Model} & \textbf{IoU} & \textbf{IoM} \\
    \midrule
    Qwen-2.5-1.5B-Instruct & 0.20 & 0.92 \\
    Qwen-2.5-Math-1.5B-Instruct & 0.30 & 0.91 \\
    Llama-3.2-3B-Instruct & 0.59 & 0.75 \\
    Phi-3-Mini-4k-Instruct & 0.80 & 0.91 \\
    \bottomrule
\end{tabular}
\caption{The edge overlap between the two error detection circuits in terms of their Intersection over Union (IoU) and Intersection over Minimum (IoM). The metrics quantify the proportion of edges shared by both circuits while considering the token position in which an edge appears.}
\label{tab:iou_iom}
\end{table}

\begin{table*}[tbp]
    \centering
    \small
    \begin{tabular}{lccccc}
        \toprule
        & Circuit & \textbf{Qwen-2.5-1.5B} & \textbf{Qwen-2.5-Math-1.5B} & \textbf{Llama-3.2-3B} & \textbf{Phi-3-Mini-3.8B} \\
        \midrule
        \multirow{2}{*}{\textit{Result}} & $\cap$ & 78.60 $\pm$ 7.46 & 47.10 $\pm$ 7.64 & 74.26 $\pm$ 5.67 & 73.02 $\pm$ 7.02 \\
        \cmidrule(l){2-6}
        & $\cup$ & 100.3 $\pm$ 0.27 & 99.28 $\pm$ 0.57 & 97.71 $\pm$ 1.50 & 99.03 $\pm$ 0.34 \\
        \cmidrule(r){1-6}
        \multirow{2}{*}{\textit{Answer}} & $\cap$ & 82.59 $\pm$ 5.08 & 46.89 $\pm$ 7.87 & 74.07 $\pm$ 5.84 & 71.96 $\pm$ 6.86 \\
        \cmidrule(l){2-6}
        & $\cup$ & 100.3 $\pm$ 0.27 & 98.80 $\pm$ 0.92 & 97.23 $\pm$ 0.93 & 99.18 $\pm$ 0.44 \\
        \midrule
        \multirow{2}{*}{\# Edges} & $\cap$ & 49 & 20 & 83 & 235 \\
        \cmidrule(l){2-6}
        & $\cup$ & 249 & 66 & 141 & 340 \\
        \bottomrule
    \end{tabular}
    \caption{Faithfulness scores for the intersection ($\cap$) and union ($\cup$) between the final \emph{soft intersection circuits} $\mathcal{C}_{\text{result}}^{(\tau)}$ and $\mathcal{C}_{\text{answer}}^{(\tau)}$ computed for each model.  For Qwen-2.5-1.5B-Instruct, the intersection and union between $\mathcal{C}_{\text{result}}^{(5/8)}$ and $\mathcal{C}_{\text{answer}}^{(8/8)}$ are calculated. For Qwen-2.5-Math-1.5B-Instruct, they are computed for $\mathcal{C}_{\text{result}}^{(7/8)}$ and $\mathcal{C}_{\text{answer}}^{(8/8)}$. For Llama-3.2-3B-Instruct, the intersection and union between $\mathcal{C}_{\text{result}}^{(7/8)}$ and $\mathcal{C}_{\text{answer}}^{(8/8)}$ are shown. For Phi-3-Mini-4k-Instruct, they are shown for $\mathcal{C}_{\text{result}}^{(7/8)}$ and $\mathcal{C}_{\text{answer}}^{(6/8)}$. The last two rows show the number of edges of the resulting circuits.}
    \label{tab:z1-z2-intersection}
\end{table*}

Table~\ref{tab:z1-z2-intersection} reports the faithfulness results obtained from the intersection $\mathcal{C}_{\text{result}} \cap \mathcal{C}_{\text{answer}}$ and union $ \mathcal{C}_{\text{result}} \cup \mathcal{C}_{\text{answer}} $ of the error detection circuits across models. The union circuits exhibit near-perfect faithfulness for both error types across all models, achieving faithfulness scores $\geq 97.00\%$. In contrast, the faithfulness of the intersection circuits is generally lower, although it remains above $70.00\%$ for models such as Qwen-2.5-1.5B-Instruct, Llama-3.2-3B-Instruct, and Phi-3-Mini-4k-Instruct. The lowest faithfulness is observed for the intersection circuit of Qwen-2.5-Math-1.5B-Instruct, likely due to its extreme sparsity, containing only 20 edges in total.

\subsection{Detection of Consistent Errors}
As outlined in Section~\ref{subsec:Decode_Error_Detection}, we expect models to struggle with differentiating between error-free samples and those containing a consistent error in both the arithmetic result and the final numeric answer. We evaluate models on a dataset of 1,000 (\textit{clean}, \textit{corrupt}) prompt pairs for each template $\mathcal{T}_i \in \{\mathcal{T}_1, \ldots, \mathcal{T}_8\}$, where the \textit{clean} prompts contain a consistent error at both specified positions. As mentioned in Section~\ref{sec:Experiments}, a prompt pair is considered correctly classified if the model predicts the \emph{clean} prompt as erroneous ($y_{clean} \in \{\text{invalid}, \text{incorrect}, \text{wrong}\}$) and the \emph{corrupt} prompt as error-free ($y_{corrupt} \in \{\text{valid}, \text{correct}, \text{right}\}$). The respective accuracy of all models is summarized in Table~\ref{tab:both_behavioral_accuracy}. Overall, the results indicate that the models perform poorly on this task. For instance, Qwen-2.5-Math-1.5B-Instruct achieves an average accuracy of only $3.37\% \pm 2.06\%$. Among the evaluated models, Phi-3-Mini-4k-Instruct demonstrates the best performance, with an average accuracy of $40.98\% \pm 18.41\%$.

To study how these findings transfer to larger models, we additionally evaluate the performance of LLaMA-3.1-70B-Instruct and Qwen-2.5-32B-Instruct on different types of errors. As shown in Table~\ref{tab:big_model_behavioral_accuracy}, we observe that, similar to the smaller models, both larger LLMs struggle more with detecting a consistent error at both error positions (result \& answer) than with detecting an error present at only the arithmetic result or the final numeric answer. In particular, the average accuracy of LLaMA-3.1-70B-Instruct for consistent errors is $44.67\% \pm 10.50\%$, a drop of more than half compared to its $100.0\% \pm 0.00\%$ accuracy on detecting errors present only at the position of the arithmetic result or the final numeric answer.

\begin{table*}[tbp]
\small
\centering
\begin{tabular}{llcccc}
\toprule
\textbf{Operation} & \textbf{Error Type} & \textbf{Qwen2.5-1.5B} & \textbf{Llama-3.2-3B} & \textbf{Phi-3-Mini-3.8B} & \textbf{Qwen2.5-Math-1.5B} \\
\midrule
\multirow{4}{*}{Subtraction} & \textit{Arithmetic Result} & 71.51 $\pm$ 11.50 & 99.66 $\pm$ 0.57 & 97.56 $\pm$ 5.74 & 93.54 $\pm$ 11.09 \\
& \textit{Numeric Answer} & 70.05 $\pm$ 11.57 & 99.60 $\pm$ 0.65 & 97.60 $\pm$ 5.68 & 93.61 $\pm$ 10.77\\
& \textit{Result \& Answer} & \cellcolor{red!20}8.68 $\pm$ \cellcolor{red!20}4.98 & \cellcolor{red!20}14.08 $\pm$ 2.01 & 99.64 $\pm$ 0.78 & \cellcolor{red!20}14.33 $\pm$ 1.60\\
\midrule
\multirow{4}{*}{Multiplication} & \textit{Arithmetic Result} & 55.30 $\pm$ 19.46 & 78.53 $\pm$ 31.47 & 99.54 $\pm$ 0.73 & 94.00 $\pm$ 13.68\\
& \textit{Numeric Answer} & 55.22 $\pm$ 18.50 & 78.89 $\pm$ 31.11 & 99.54 $\pm$ 0.62 & 95.03 $\pm$ 13.12\\
& \textit{Result \& Answer} & 62.88 $\pm$ 17.26 & \cellcolor{red!20}42.73 $\pm$ \cellcolor{red!20}30.73 & 99.65 $\pm$ 0.45 & \cellcolor{red!20}80.96 $\pm$ 9.25\\
\midrule
\multirow{4}{*}{Division} & \textit{Arithmetic Result} & 59.86 $\pm$ 19.08 & 86.44 $\pm$ 13.04 & 94.52 $\pm$ 12.35 & 99.30 $\pm$ 0.92\\
& \textit{Numeric Answer} & 60.06 $\pm$ 18.07 & 86.76 $\pm$ 12.97 & 94.18 $\pm$ 12.72 & 99.53 $\pm$ 0.69\\
& \textit{Result \& Answer} & \cellcolor{red!20}50.90 $\pm$ 15.84 & \cellcolor{red!20}11.79 $\pm$ 7.99 & 99.14 $\pm$ 1.50 & \cellcolor{red!20}57.95 $\pm$ 13.89\\
\bottomrule
\end{tabular}
\caption{Accuracy of models in correctly classifying the solutions' validity of (\emph{clean}, \emph{corrupt}) prompt pairs across different arithmetic operations and error types. Values represent the mean accuracy across all templates, reported with their corresponding standard deviation. Cells highlighted in red indicate cases where the accuracy for \emph{Result \& Answer} is notably lower than for \emph{Numeric Answer} and \emph{Arithmetic Result}.}
\label{tab:arithmetic_operations_performance}
\end{table*}

\subsection{Consistency Heads}
\begin{figure*}[htbp]
    \centering
    \begin{subfigure}[b]{\textwidth}
        \includegraphics[width=\linewidth]{figures/attn_patterns/template_0/pattern_qwen_l12h2.png}
        \caption{consistency head L12H2.}
        \label{fig:attention-pattern-qwen-consistency}
    \end{subfigure}
    
    \vspace{\baselineskip}
    
    \begin{subfigure}[b]{\textwidth}
        \includegraphics[width=\linewidth]{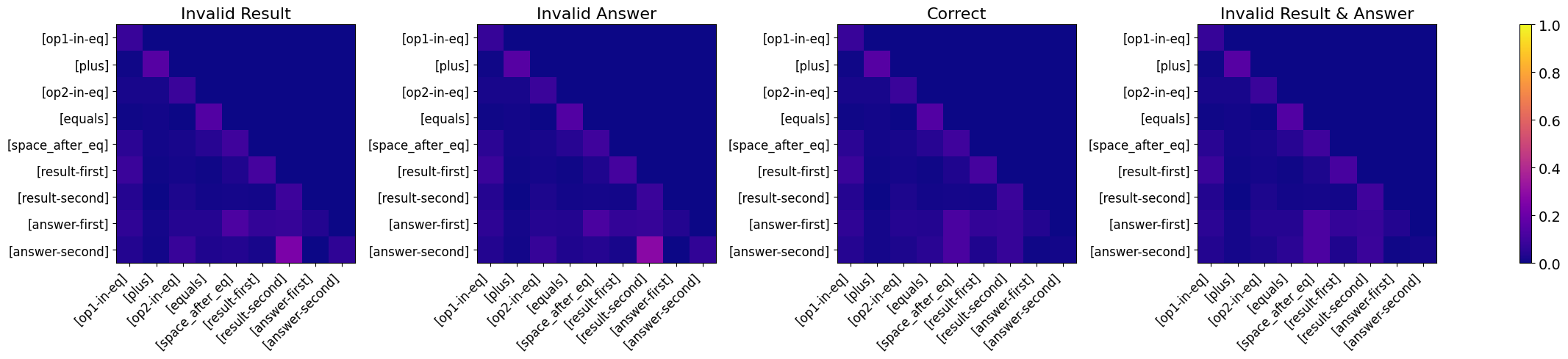}
        \caption{(in)consistency head L13H1.}
        \label{fig:attention-pattern-qwen-inconsistency}
    \end{subfigure}   
    \caption{Attention patterns of two \emph{consistency heads} in Qwen-2.5-1.5B-Instruct. Reported scores are averaged over 5,000 prompts where (\emph{left}) an error is present at the position of the arithmetic result, (\emph{second to left}) an error is present at the position of the final numeric answer, (\emph{second to right}) no error is present, and (\emph{right}) a consistent error is present at both considered positions.}
    \label{fig:attention-pattern-qwen}
\end{figure*}

\begin{figure*}[htbp]
    \centering
    \begin{subfigure}[b]{\textwidth}
        \includegraphics[width=\linewidth]{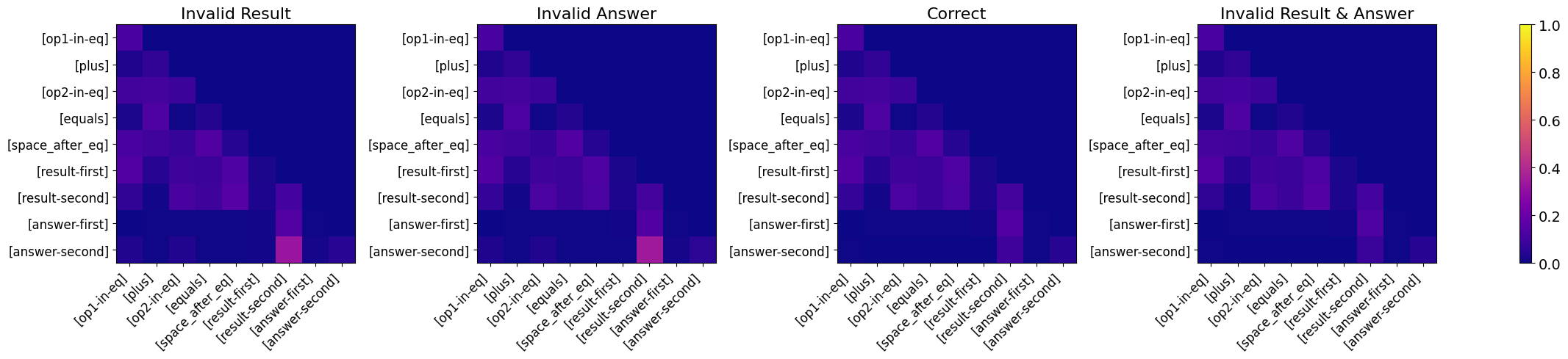}
        \caption{(in)consistency head L13H0.}
        \label{fig:attention-pattern-qwen-math-inconsistency-a}
    \end{subfigure}
    
    \vspace{\baselineskip}

    \begin{subfigure}[b]{\textwidth}
        \includegraphics[width=\linewidth]{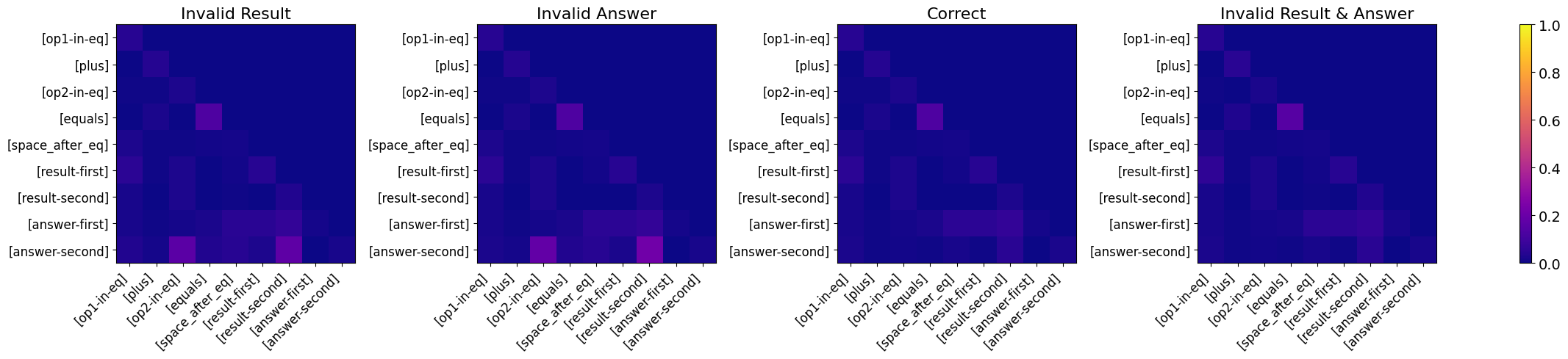}
        \caption{(in)consistency head L13H1.}
        \label{fig:attention-pattern-qwen-math-inconsistency-b}
    \end{subfigure}
    
    \caption{Attention patterns of two \emph{consistency heads} in Qwen-2.5-Math-1.5B-Instruct. Reported scores are averaged over 5,000 prompts where (\emph{left}) an error is present at the position of the arithmetic result, (\emph{second to left}) an error is present at the position of the final numeric answer, (\emph{second to right}) no error is present, and (\emph{right}) a consistent error is present at both considered positions.}
    \label{fig:attention-pattern-qwen-math}
\end{figure*}

\begin{figure*}[htbp]
    \centering
    \begin{subfigure}[b]{\textwidth}
        \includegraphics[width=\linewidth]{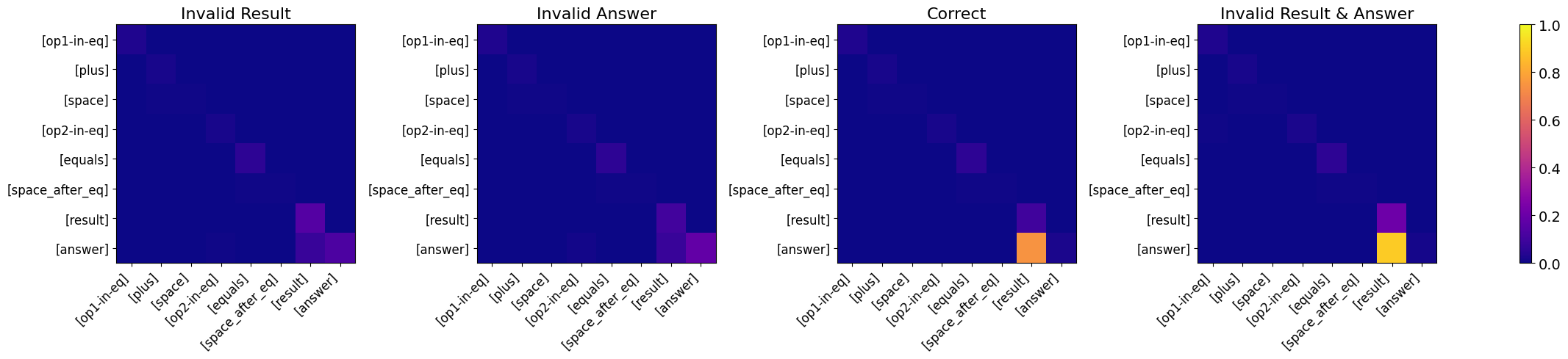}
        \caption{consistency head L4H14.}
        \label{fig:attention-pattern-llama-consistency}
    \end{subfigure}

    \vspace{\baselineskip}
    
    \begin{subfigure}[b]{\textwidth}
        \includegraphics[width=\linewidth]{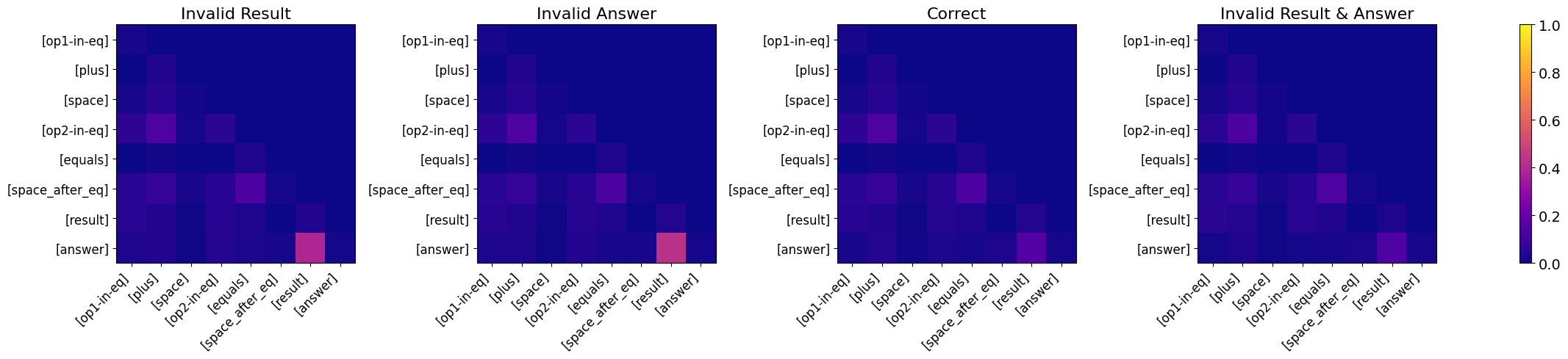}
        \caption{(in)consistency head L8H1.}
        \label{fig:attention-pattern-llama-inconsistency}
    \end{subfigure}
    
    \caption{Attention patterns of two \emph{consistency heads} in Llama-3.2-3B-Instruct. Reported scores are averaged over 5,000 prompts where (\emph{left}) an error is present at the position of the arithmetic result, (\emph{second to left}) an error is present at the position of the final numeric answer, (\emph{second to right}) no error is present, and (\emph{right}) a consistent error is present at both considered positions. Note that Llama-3.2 does not tokenize numbers digit-by-digit.}
    \label{fig:attention-pattern-llama}
\end{figure*}

\begin{figure*}[htbp]
    \centering
    \begin{subfigure}[b]{\textwidth}
        \includegraphics[width=\linewidth]{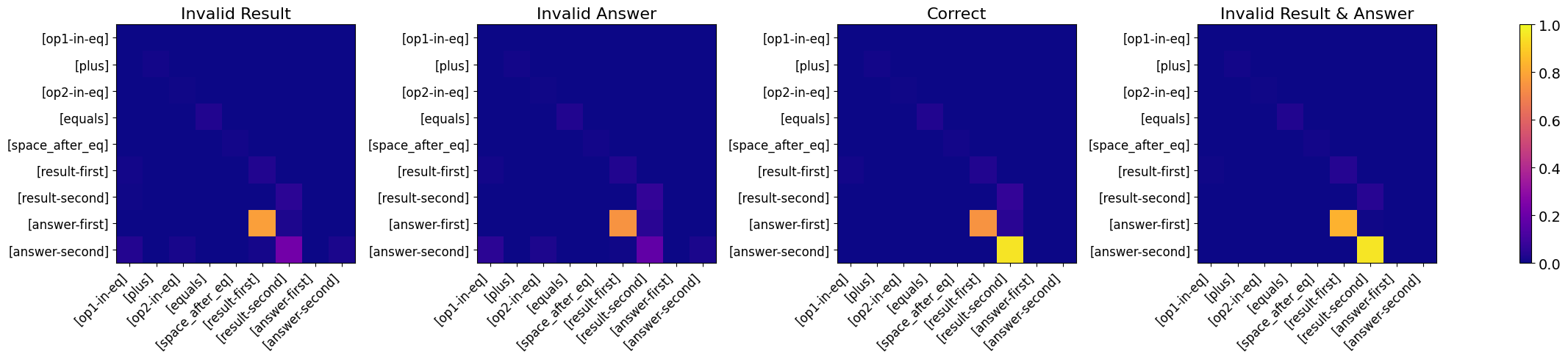}
        \caption{consistency head L10H1.}
        \label{fig:attention-pattern-phi-consistency}
    \end{subfigure}
    
    \vspace{\baselineskip}
    
    \begin{subfigure}[b]{\textwidth}
        \includegraphics[width=\linewidth]{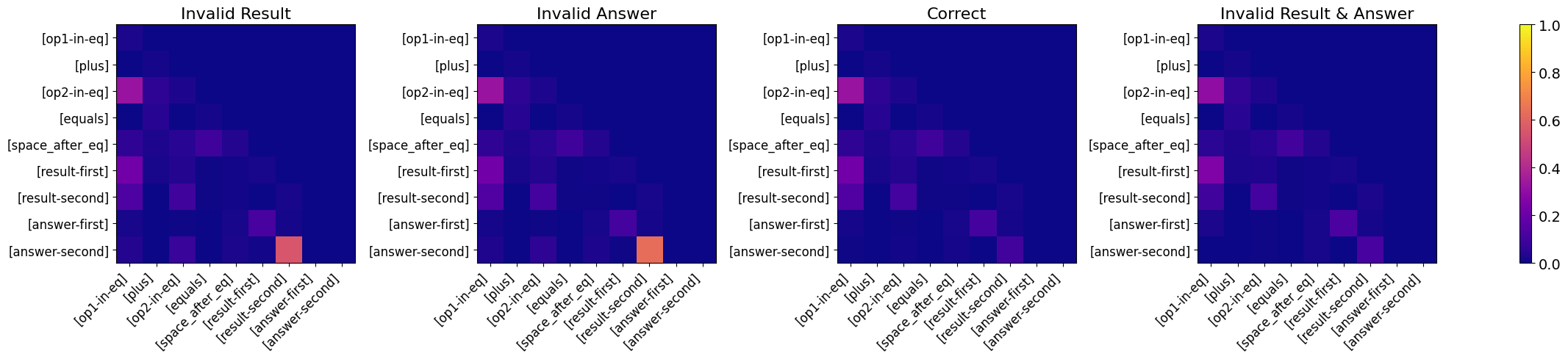}
        \caption{(in)consistency head L10H14.}
        \label{fig:attention-pattern-phi-inconsistency}
    \end{subfigure}
    
    \caption{Attention patterns of two \emph{consistency heads} in Phi-3-Mini-4k-Instruct. Reported scores are averaged over 5,000 prompts where (\emph{left}) an error is present at the position of the arithmetic result, (\emph{second to left}) an error is present at the position of the final numeric answer, (\emph{second to right}) no error is present, and (\emph{right}) a consistent error is present at both considered positions.}
    \label{fig:attention-pattern-phi}
\end{figure*}

As mentioned in Section~\ref{subsec:Decode_Error_Detection}, we find that consistency heads play an important role in the model's arithmetic error detection process. These heads exhibit high average attention scores when the digits of the arithmetic result either align or misalign with the final numeric answer.
Figures~\ref{fig:attention-pattern-qwen-consistency} and~\ref{fig:attention-pattern-qwen-inconsistency} illustrate examples of attention scores for two of these heads in Qwen-2.5-1.5B-Instruct. Similar patterns are presented for Qwen-2.5-Math-1.5B-Instruct in Figures~\ref{fig:attention-pattern-qwen-math-inconsistency-a} and~\ref{fig:attention-pattern-qwen-math-inconsistency-b}, for Llama-3.2-3B-Instruct in Figures~\ref{fig:attention-pattern-llama-consistency} and~\ref{fig:attention-pattern-llama-inconsistency}, and for Phi-3-Mini-4k-Instruct in Figures~\ref{fig:attention-pattern-phi-consistency} and~\ref{fig:attention-pattern-phi-inconsistency}. A comprehensive list of all identified consistency heads for each model is provided in Table~\ref{tab:heads}.

\subsection{Consistency Heads Patching}\label{app:consistency_heads_patching_results}
\begin{figure*}[htbp]
    \centering
    \begin{subfigure}[b]{0.32\textwidth}
        \centering
        \includegraphics[width=\linewidth]{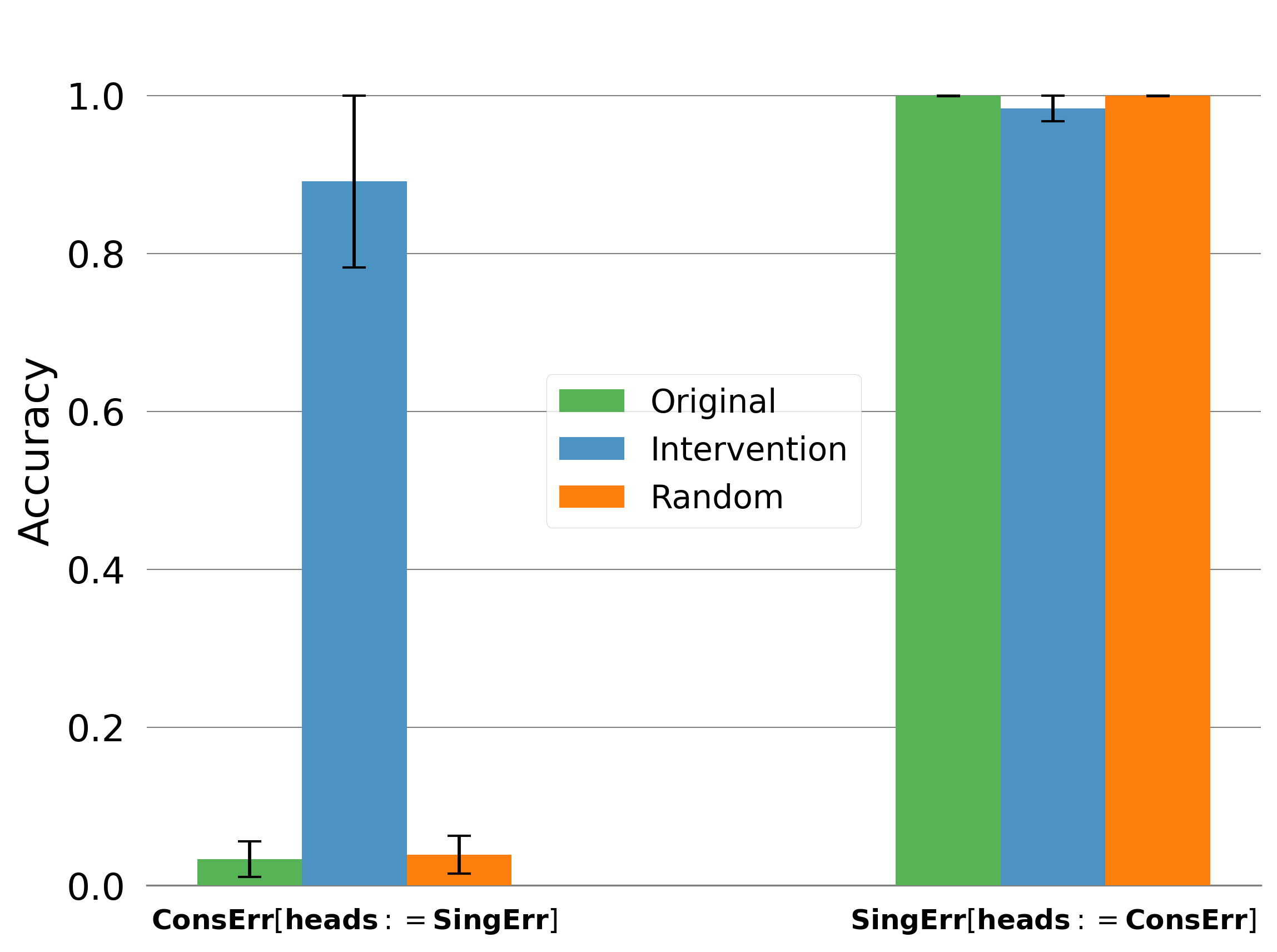}
        \caption{Qwen-2.5-Math-1.5B-Instruct}
    \end{subfigure}
    \begin{subfigure}[b]{0.32\textwidth}
        \centering
        \includegraphics[width=\linewidth]{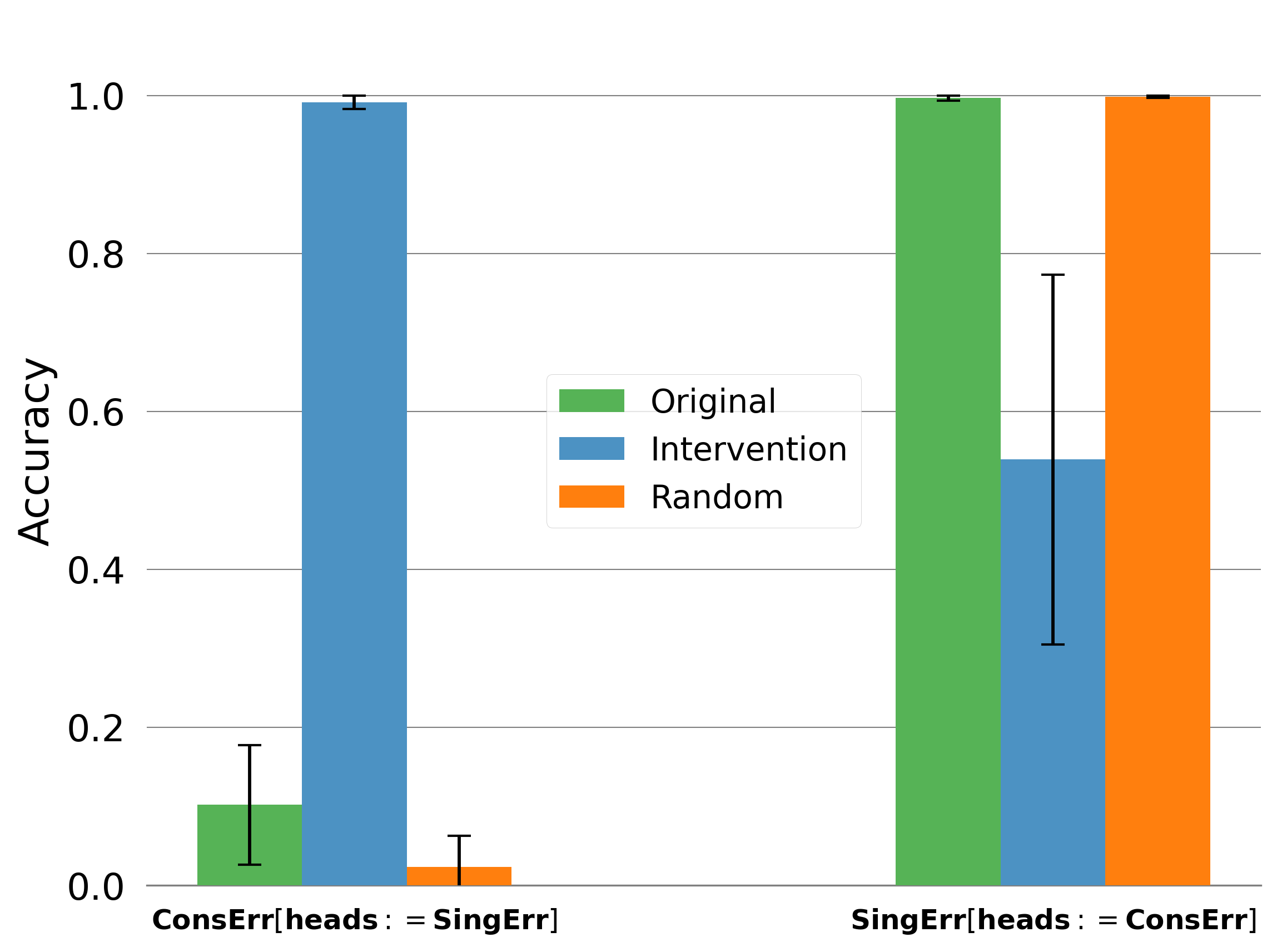}
        \caption{Llama-3.2-3B-Instruct}
    \end{subfigure}
    \begin{subfigure}[b]{0.32\textwidth}
        \centering
        \includegraphics[width=\linewidth]{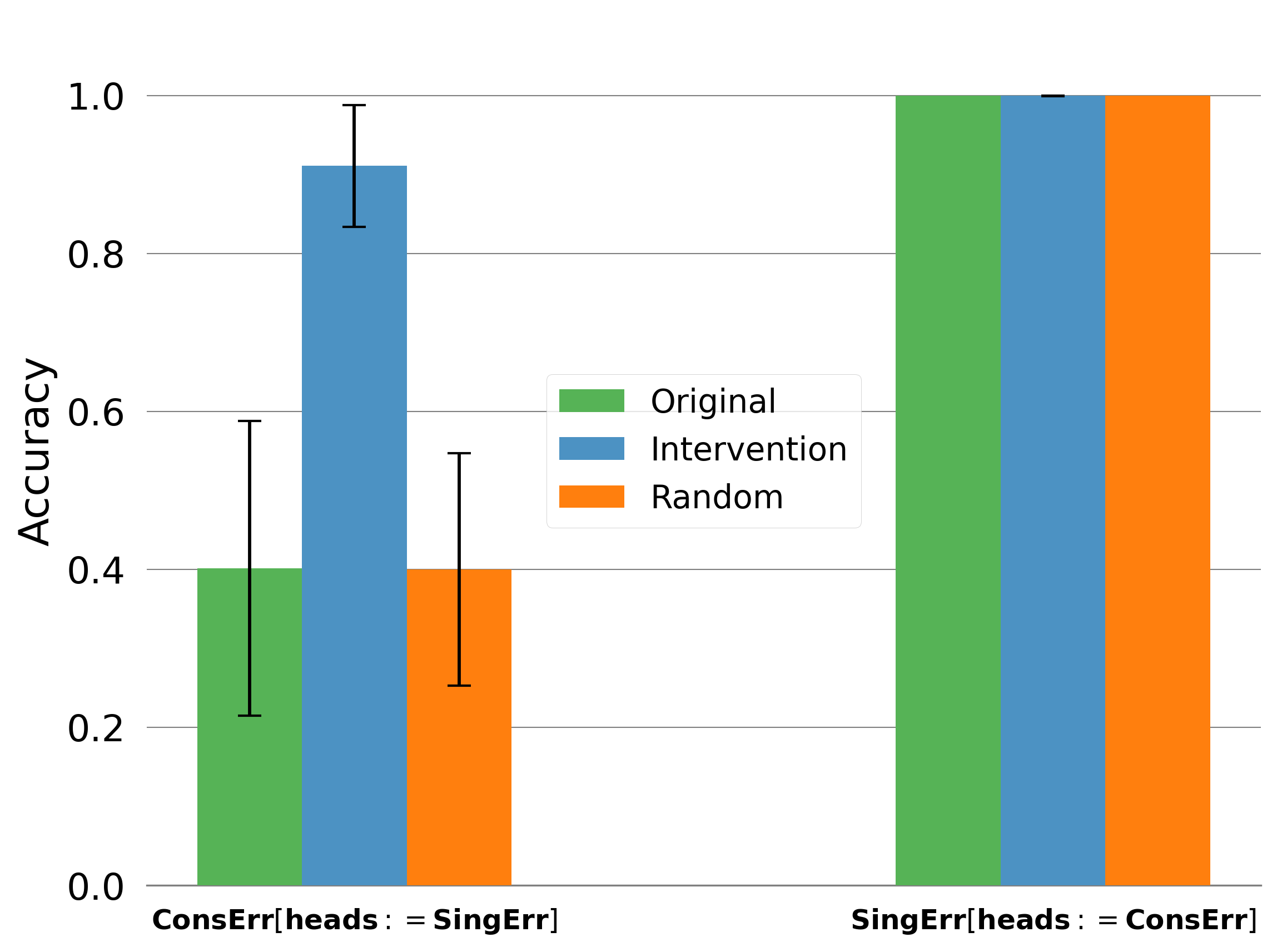}
        \caption{Phi-3-Mini-4k-Instruct}
    \end{subfigure}
    \caption{Accuracy of models on (\emph{clean}, \emph{corrupt}) prompt pairs, where the \emph{clean} prompt contains a consistent error at both error positions. The blue intervention bar represents the result after patching a set of \emph{consistency heads} (for a list of heads, please refer to Table~\ref{tab:heads} in the Appendix). In contrast, the orange bar indicates the accuracy after patching a set of randomly chosen attention heads that are not labeled as consistency heads.}
    \label{fig:attention-patching-full}
\end{figure*}

Figure~\ref{fig:attention-patching-full} shows the models' accuracy in detecting consistent errors at both the position of the arithmetic result and the final numeric answer, before and after patching a small subset of \emph{consistency} heads, as outlined in Section~\ref{subsec:Dissociation}. For the exact list of patched heads, please refer to Table~\ref{tab:heads}. Consistent with the results reported in Section~\ref{subsec:Dissociation} for Qwen-2.5-1.5B-Instruct, we observe a significant improvement in accuracy for Qwen-2.5-Math-1.5B-Instruct, Llama-3.2-3B-Instruct, and Phi-3-Mini-4k-Instruct.

\subsection{Computation Circuits}
\begin{figure*}[htbp]
    \centering
    \begin{subfigure}[b]{0.32\textwidth}
        \centering
        \includegraphics[width=\linewidth]{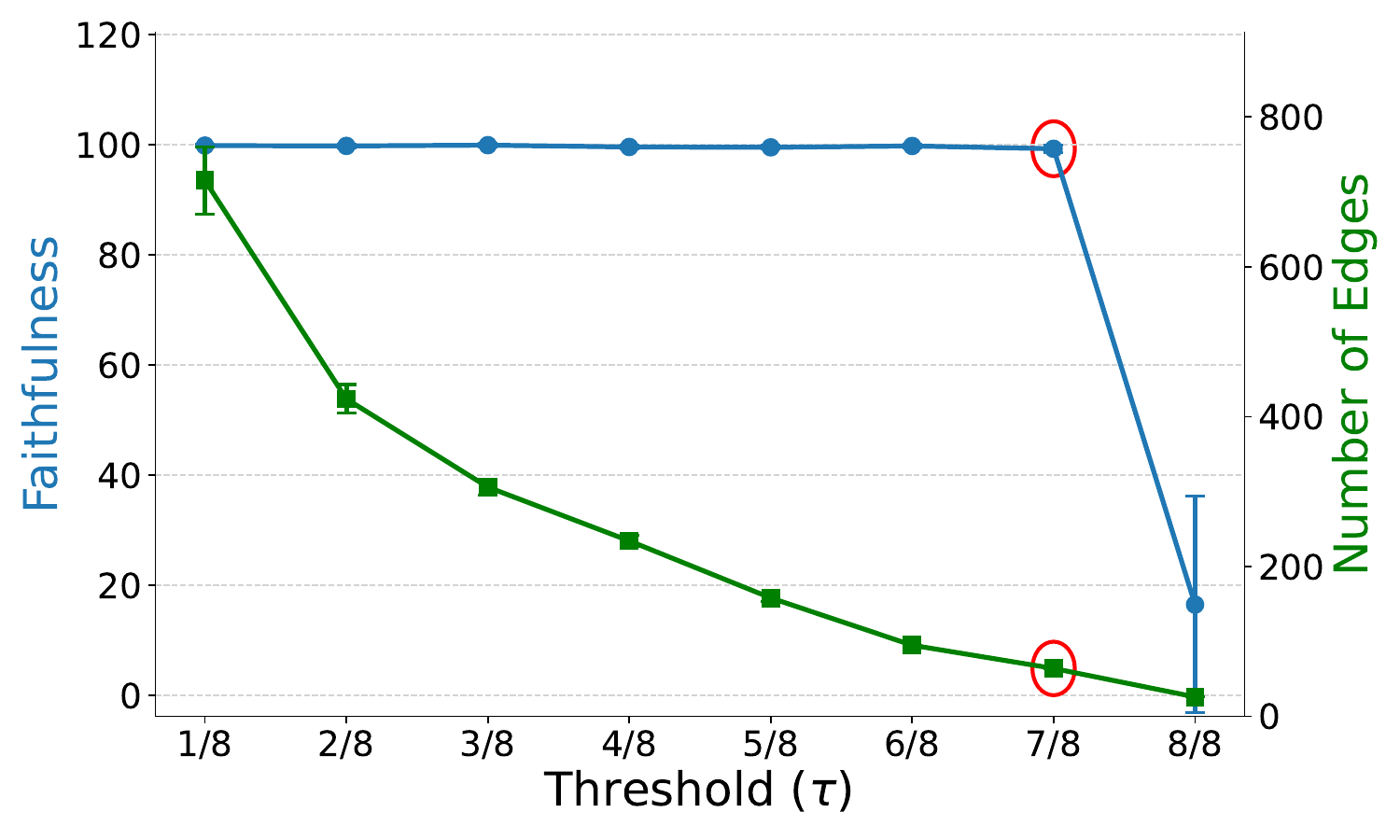}
        \caption{Invalid Result}
        \label{fig:soft_qwen-math_result}
    \end{subfigure}
    \hfill
    \begin{subfigure}[b]{0.32\textwidth}
        \centering
        \includegraphics[width=\linewidth]{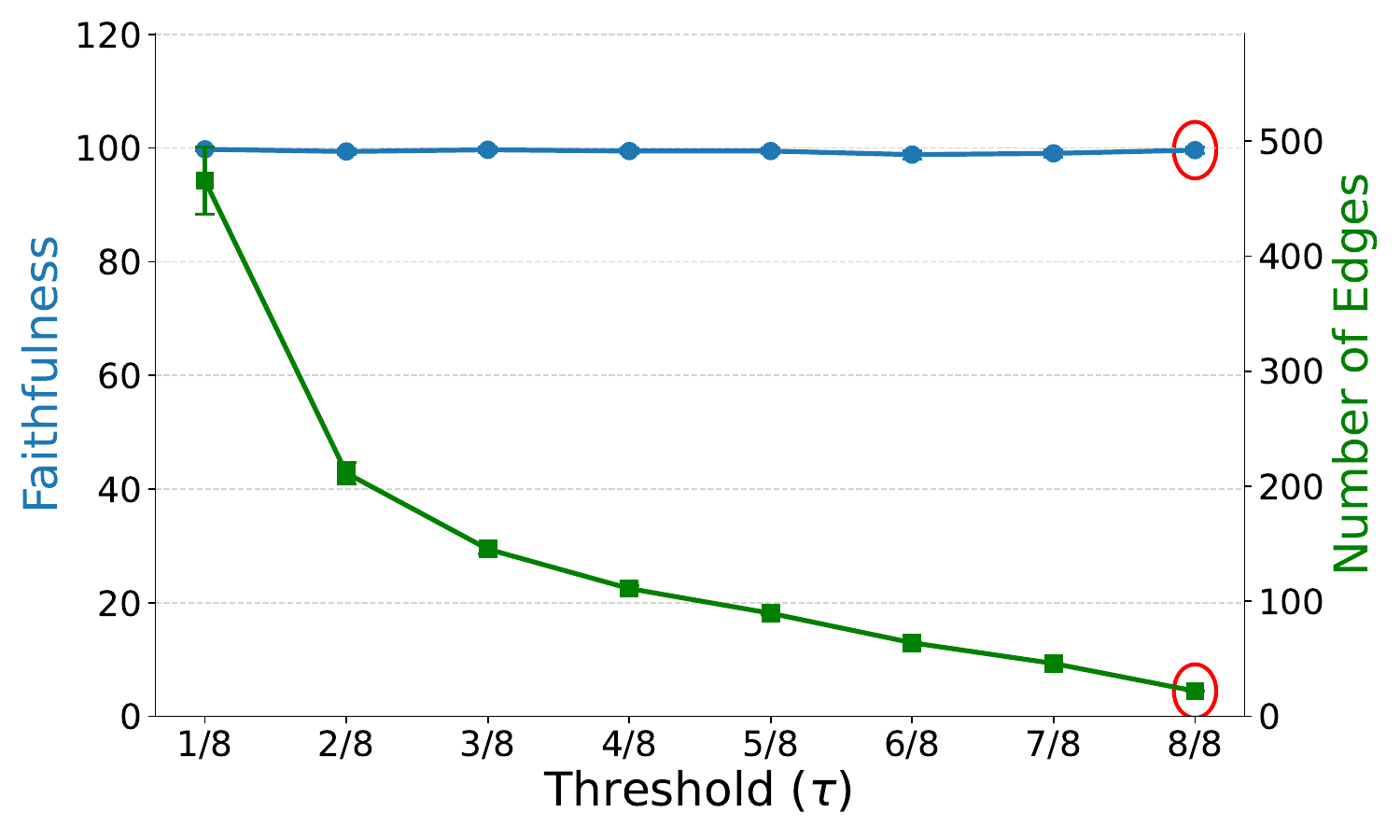}
        \caption{Invalid Answer}
        \label{fig:soft_qwen-math_answer}
    \end{subfigure}
    \hfill
    \begin{subfigure}[b]{0.32\textwidth}
        \centering
        \includegraphics[width=\linewidth]{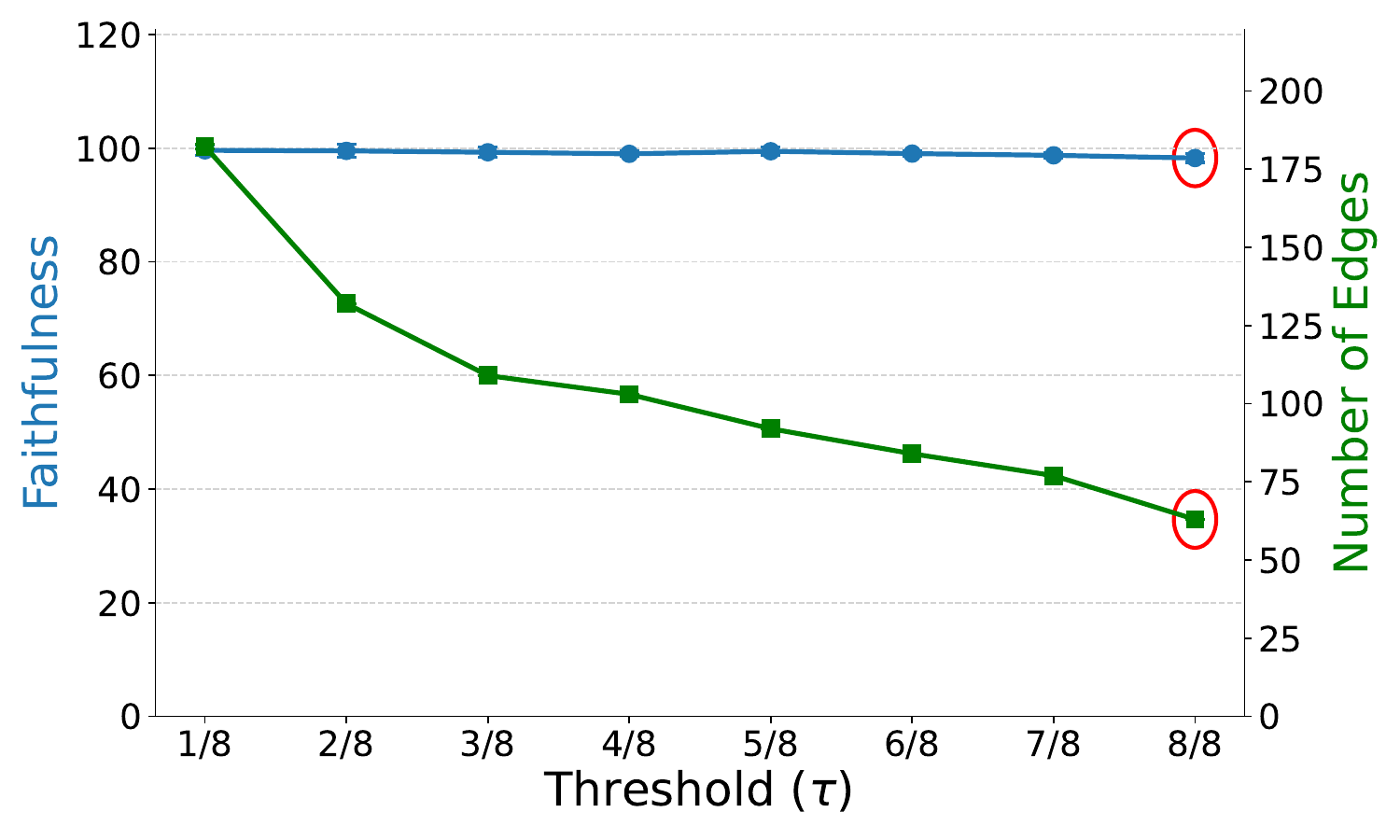}
        \caption{Computation}
        \label{fig:soft_qwen-math_computation}
    \end{subfigure}
    
    \caption{The number of edges and average faithfulness scores of the \emph{soft intersection circuit} for different threshold values, $\tau$. Red circles indicate the soft intersection circuit that best trade offs size with faithfulness. Results are shown for Qwen-2.5-Math-1.5B-Instruct.}
    \label{fig:faithfulnes_qwen-math_full}
\end{figure*}

\begin{figure*}[htbp]
    \centering
    \begin{subfigure}[b]{0.32\textwidth}
        \centering
        \includegraphics[width=\linewidth]{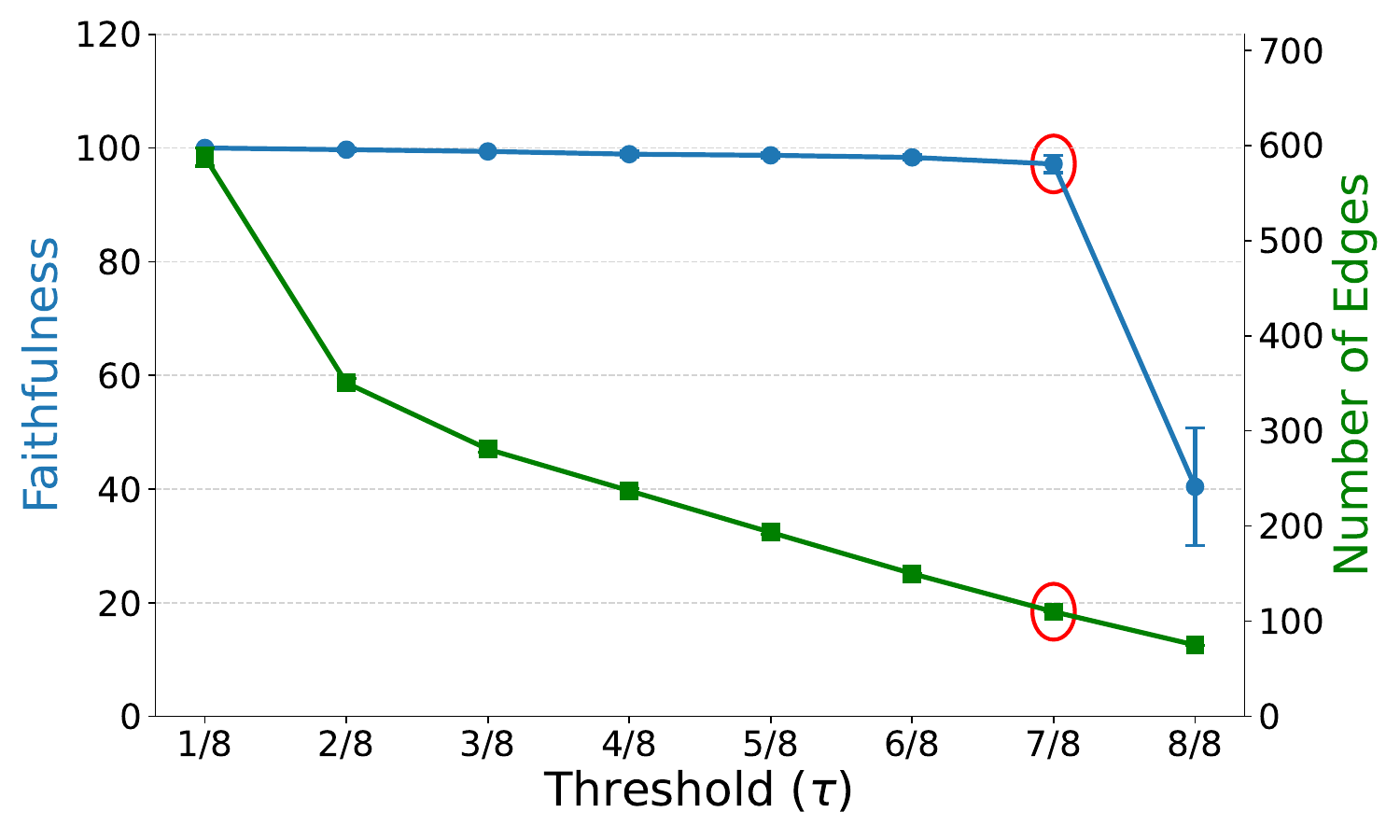}
        \caption{Invalid Result}
        \label{fig:soft_llama_result}
    \end{subfigure}
    \hfill
    \begin{subfigure}[b]{0.32\textwidth}
        \centering
        \includegraphics[width=\linewidth]{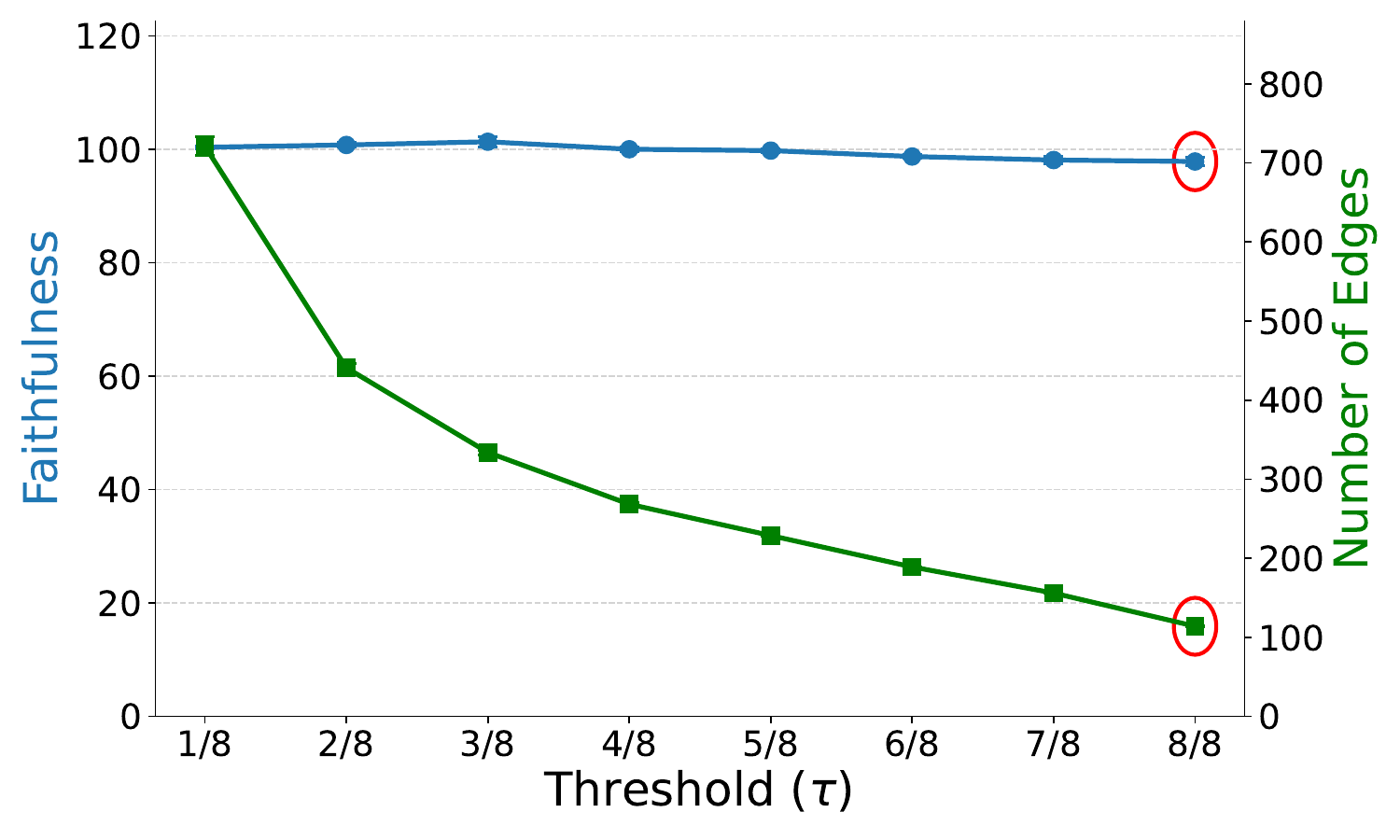}
        \caption{Invalid Answer}
        \label{fig:soft_llama_answer}
    \end{subfigure}
    \hfill
    \begin{subfigure}[b]{0.32\textwidth}
        \centering
        \includegraphics[width=\linewidth]{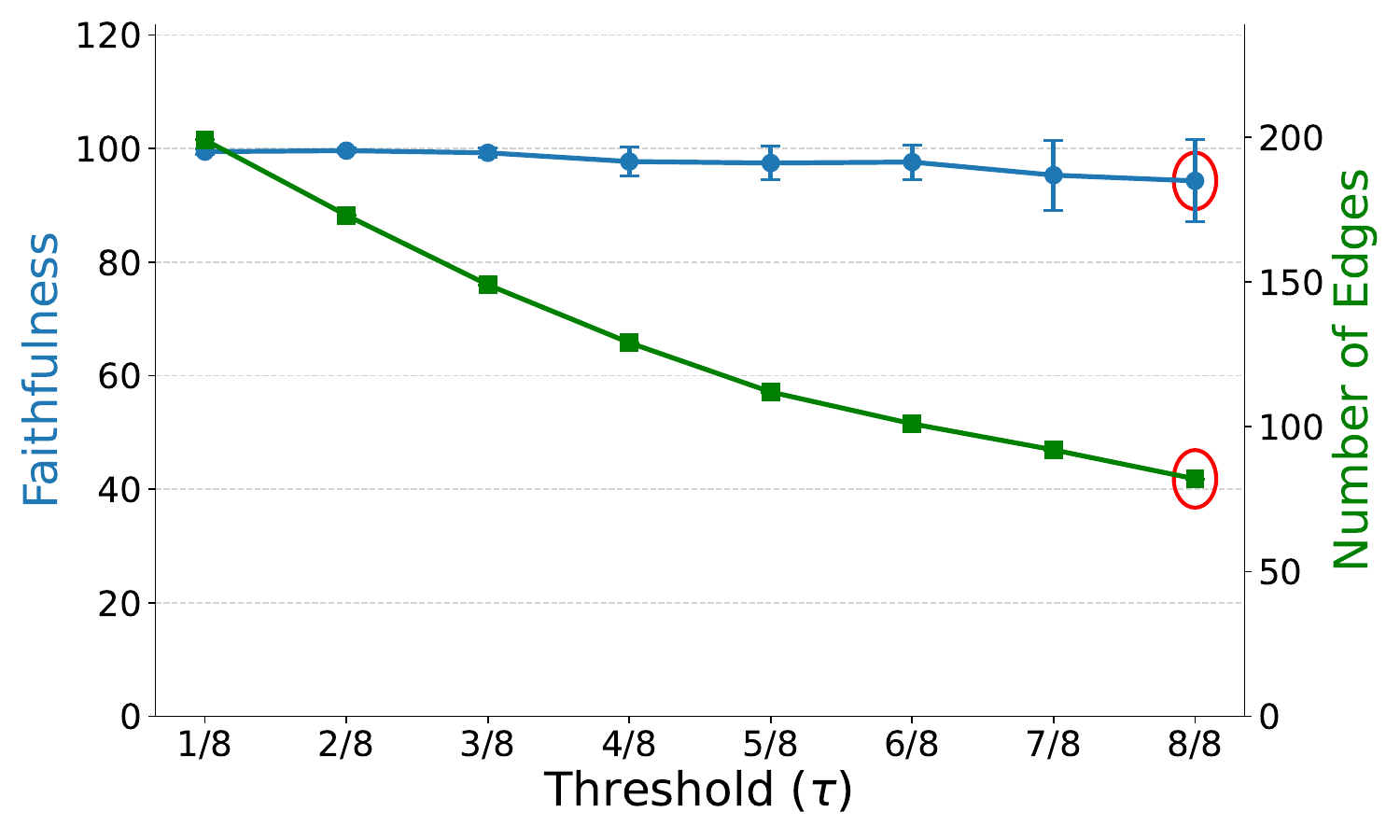}
        \caption{Computation}
        \label{fig:soft_llama_computation}
    \end{subfigure}
    
    \caption{The number of edges and average faithfulness scores of the \emph{soft intersection circuit} for different threshold values, $\tau$. Red circles indicate the soft intersection circuit that best trade offs size with faithfulness. Results are shown for Llama-3.2-3B-Instruct.}
    \label{fig:faithfulnes_llama_full}
\end{figure*}

\begin{figure*}[htbp]
    \centering
    \begin{subfigure}[b]{0.32\textwidth}
        \centering
        \includegraphics[width=\linewidth]{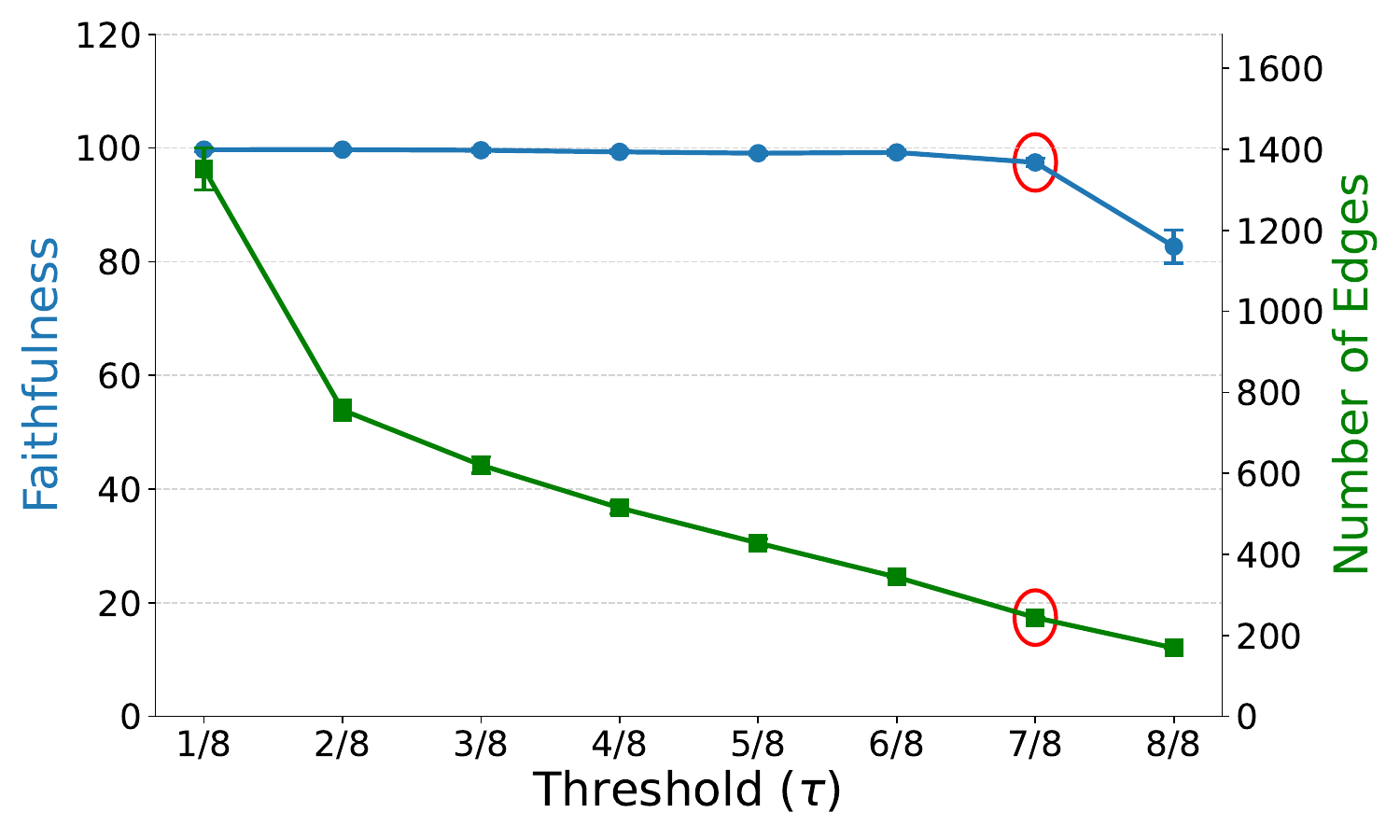}
        \caption{Invalid Result}
        \label{fig:soft_phi_result}
    \end{subfigure}
    \hfill
    \begin{subfigure}[b]{0.32\textwidth}
        \centering
        \includegraphics[width=\linewidth]{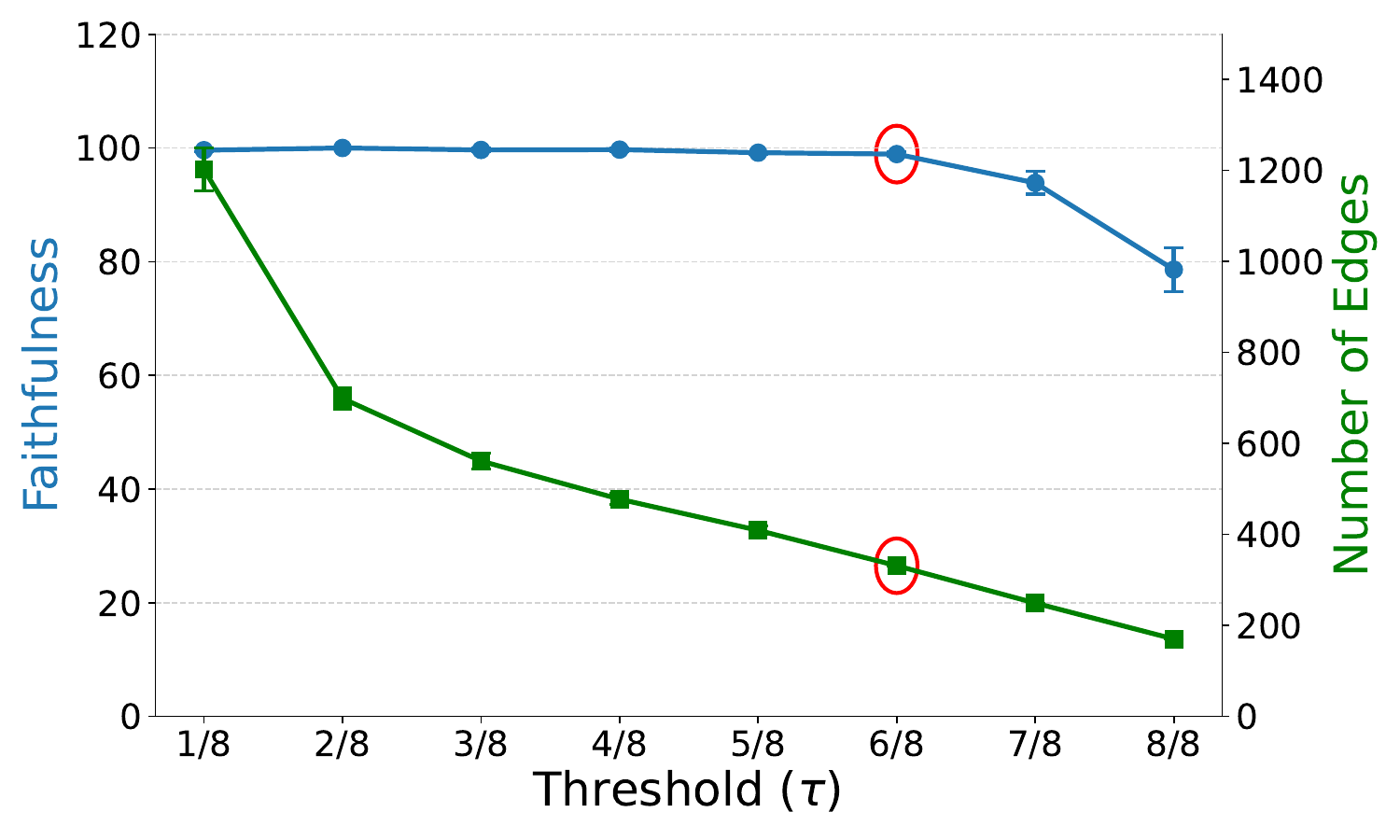}
        \caption{Invalid Answer}
        \label{fig:soft_phi_answer}
    \end{subfigure}
    \hfill
    \begin{subfigure}[b]{0.32\textwidth}
        \centering
        \includegraphics[width=\linewidth]{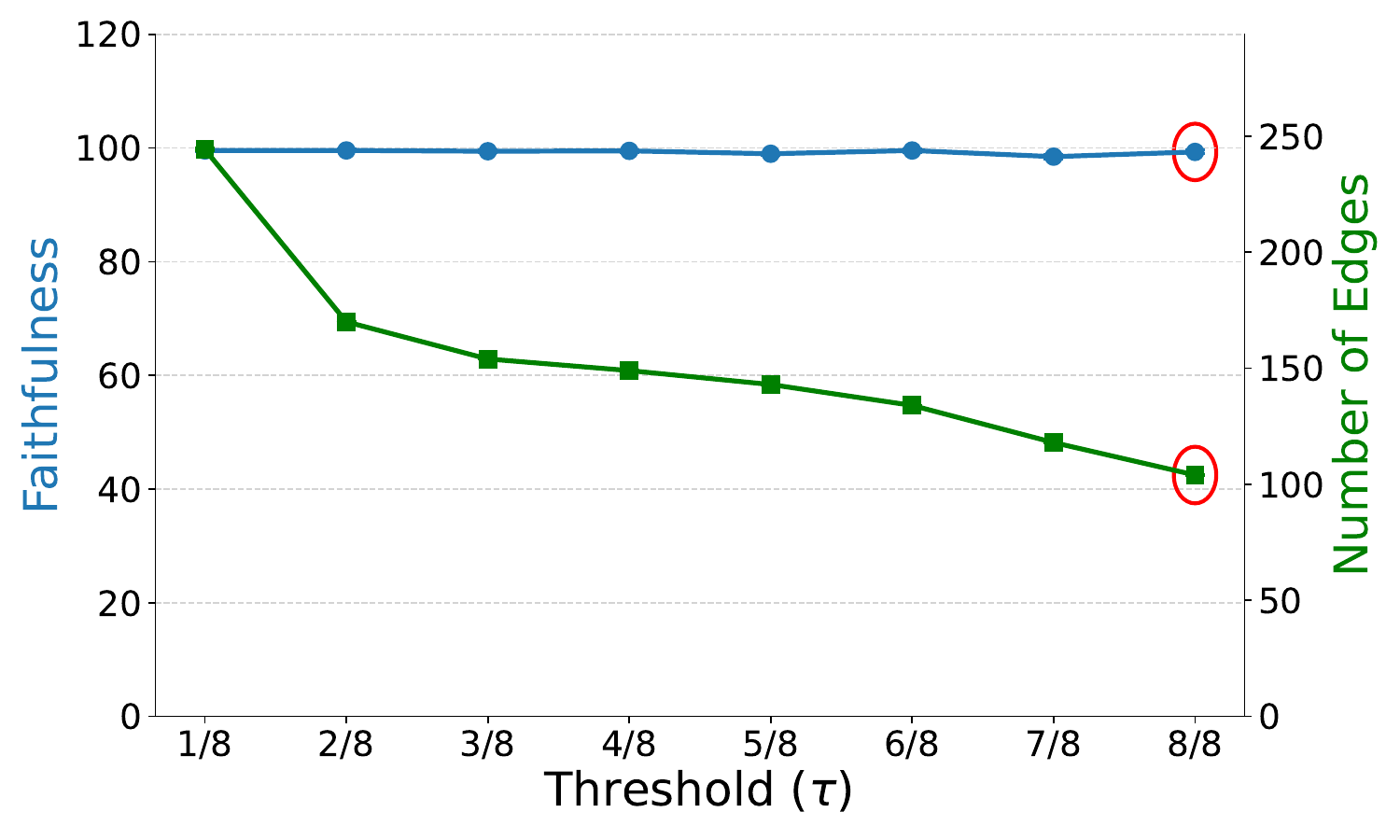}
        \caption{Computation}
        \label{fig:soft_phi_computation}
    \end{subfigure}
    
    \caption{The number of edges and average faithfulness scores of the \emph{soft intersection circuit} for different threshold values, $\tau$. Red circles indicate the soft intersection circuit that best trade offs size with faithfulness. Results are shown for Phi-3-Mini-4k-Instruct.}
    \label{fig:faithfulnes_phi_full}
\end{figure*}

\begin{figure*}[tbp]
  \centering
  \includegraphics[width=\textwidth]{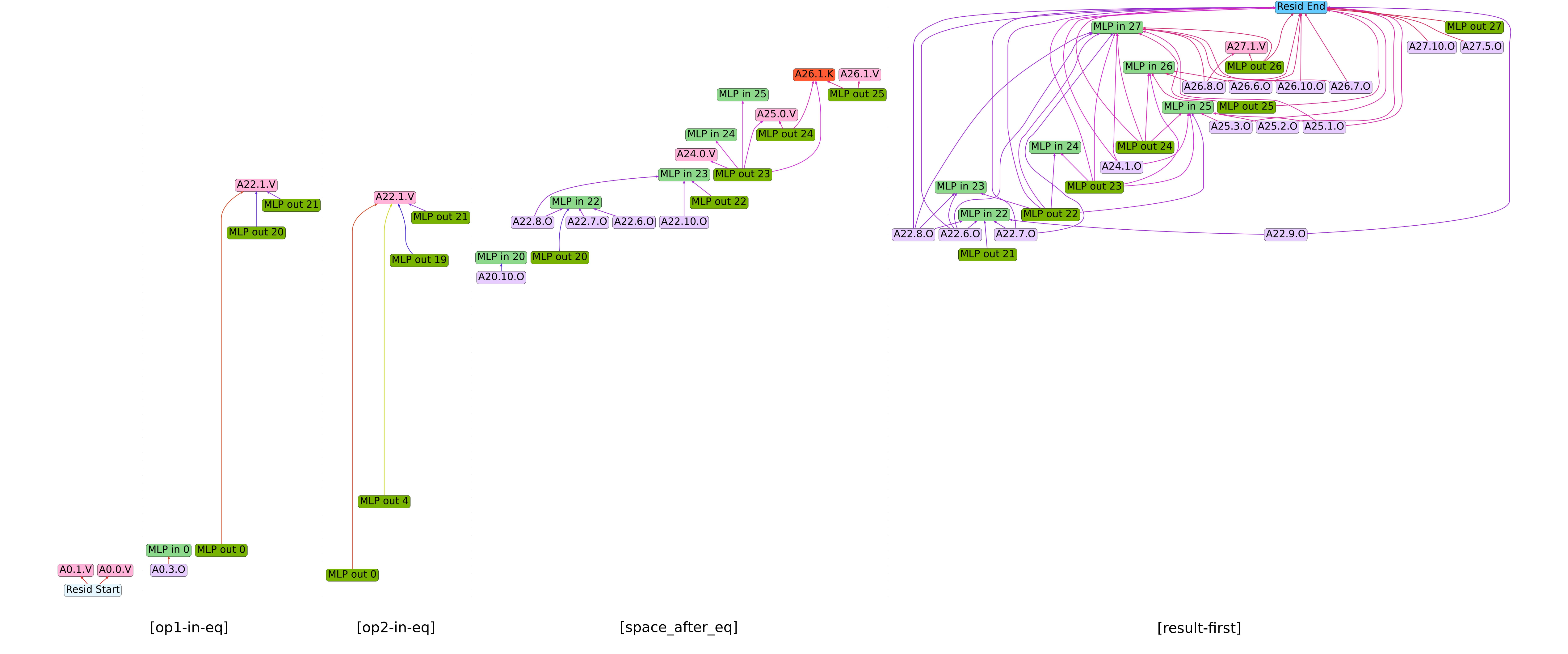}
  \caption{The computation circuit $\mathcal{C}_{\text{computation}}^{(8/8)}$ of Qwen-2.5-1.5B-Instruct obtained after taking the soft intersection between all template circuits with a threshold value of $\tau = \frac{8}{8}$.}
  \label{fig:qwen_computation_circuit}
\end{figure*}

\begin{figure*}[tbp]
  \centering
  \includegraphics[width=\textwidth]{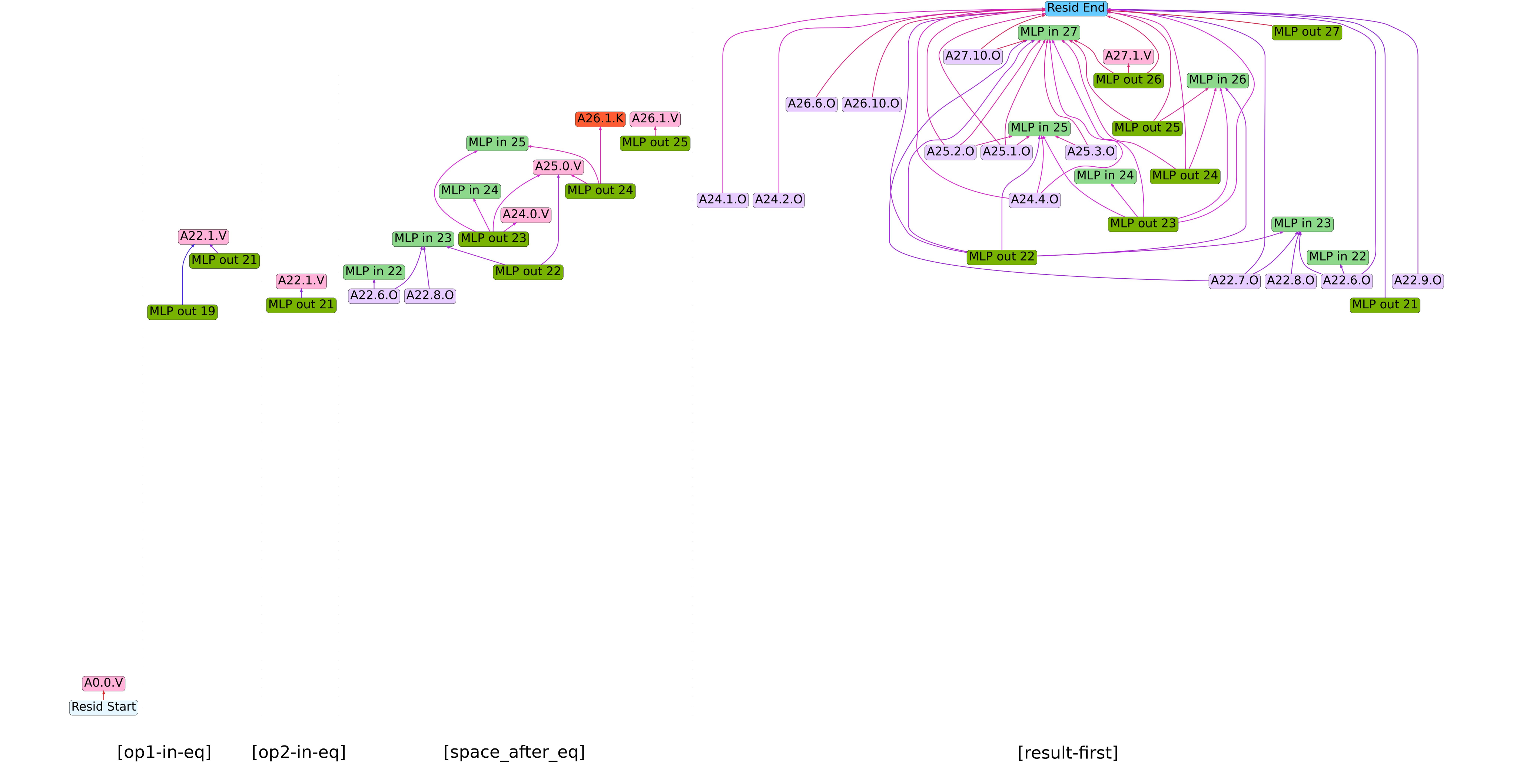}
  \caption{The computation circuit $\mathcal{C}_{\text{computation}}^{(8/8)}$ of Qwen-2.5-Math-1.5B-Instruct obtained after taking the soft intersection between all template circuits with a threshold value of $\tau = \frac{8}{8}$.}
  \label{fig:qwen_math_computation_circuit}
\end{figure*}

\begin{figure*}[tbp]
  \centering
  \includegraphics[width=\textwidth]{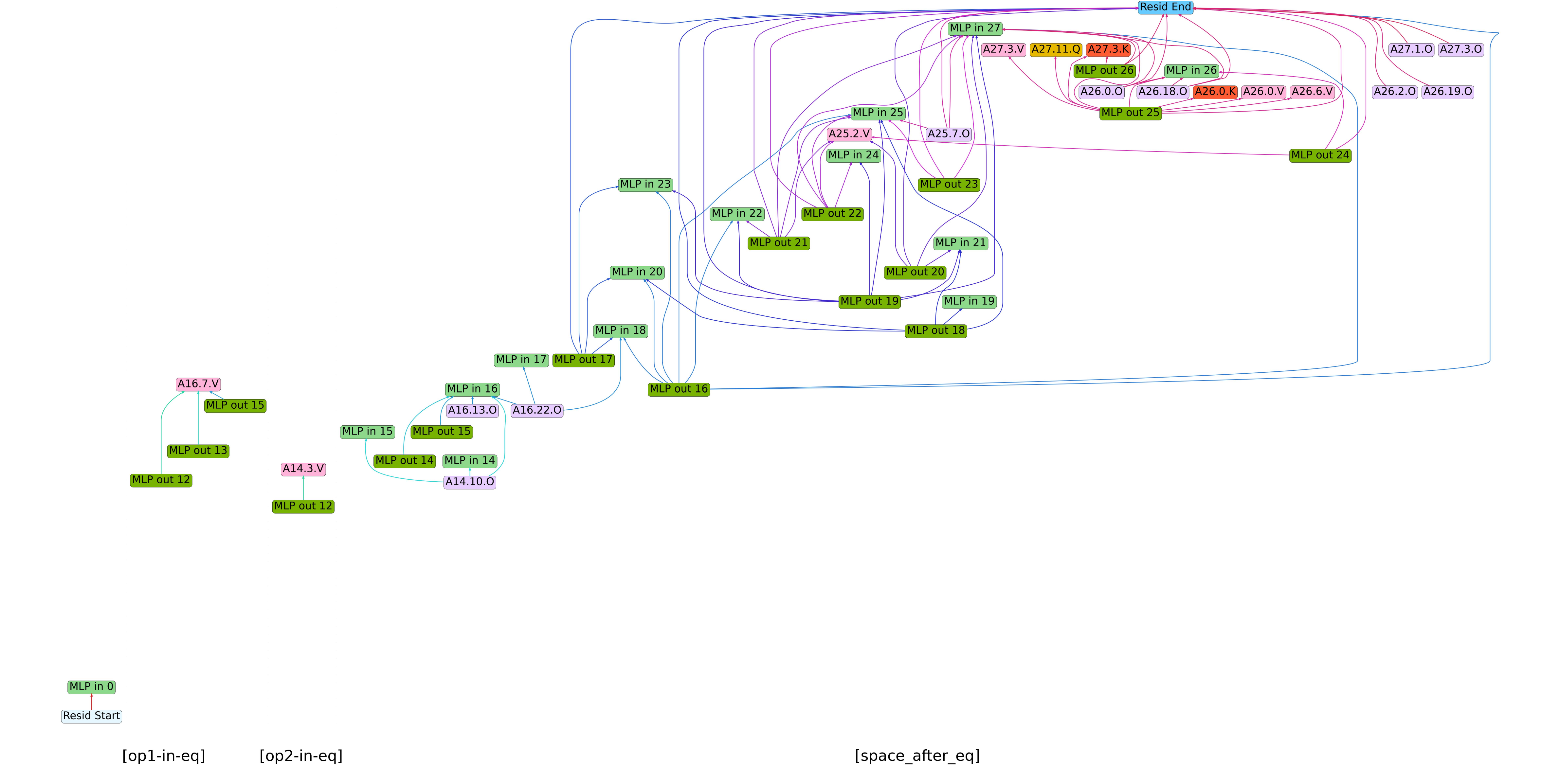}
  \caption{The computation circuit $\mathcal{C}_{\text{computation}}^{(8/8)}$ of Llama-3.2-3B-Instruct obtained after taking the soft intersection between all template circuits with a threshold value of $\tau = \frac{8}{8}$.}
  \label{fig:llama_computation_circuit}
\end{figure*}

\begin{figure*}[tbp]
  \centering
  \includegraphics[width=\textwidth]{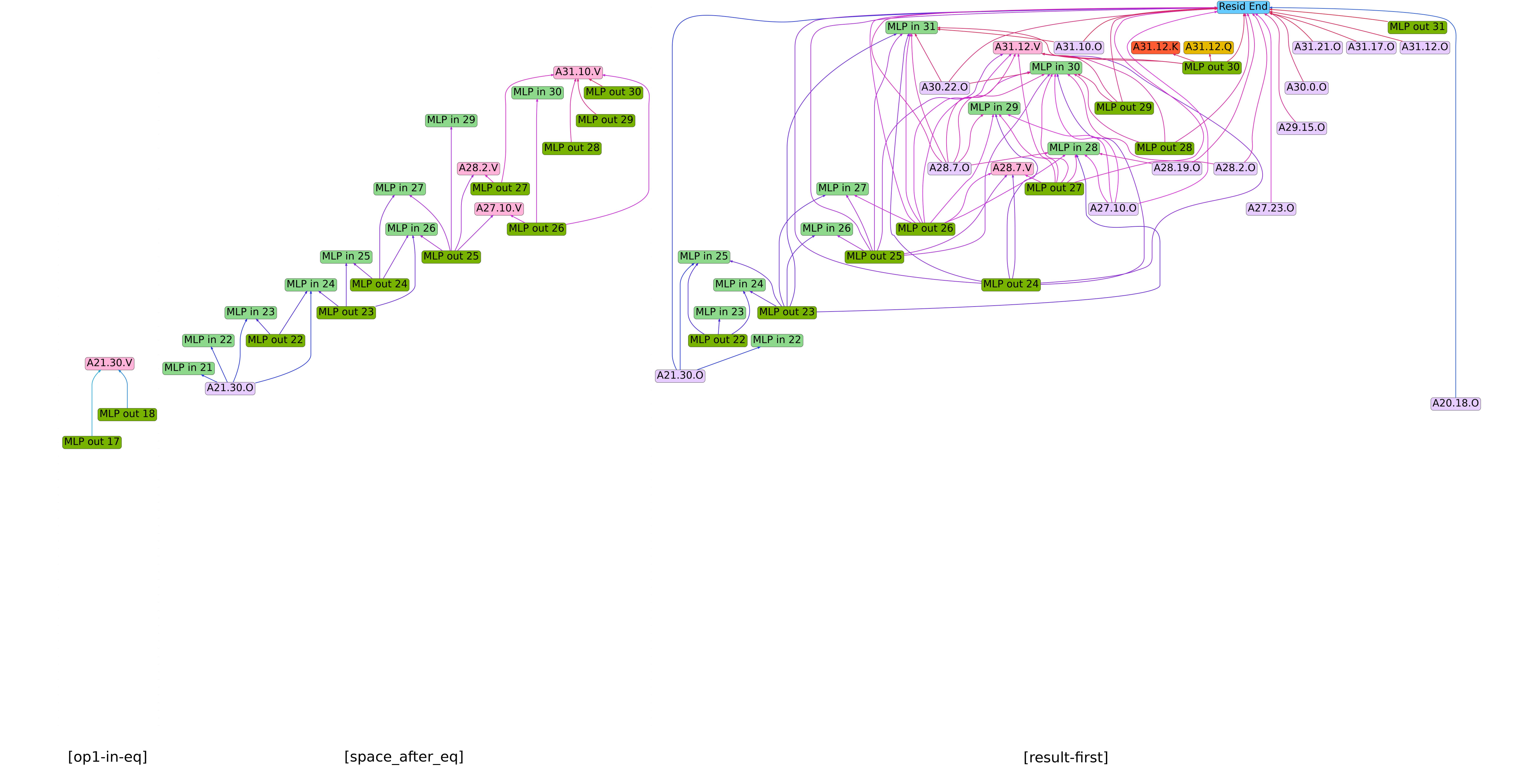}
  \caption{The computation circuit $\mathcal{C}_{\text{computation}}^{(8/8)}$ of Phi-3-Mini-4k-Instruct obtained after taking the soft intersection between all template circuits with a threshold value of $\tau = \frac{8}{8}$.}
  \label{fig:phi_computation_circuit}
\end{figure*}

As outlined in Section~\ref{app:computation_circuit}, we identify the circuits responsible for predicting the correct arithmetic result of the equations in the prompts described in Section~\ref{sec:Circuits_Arithmetic_Error_Detection}. Table~\ref{tab:faith_size} presents the size and faithfulness of the respective circuits for all models and templates. Results for the faithfulness scores and sizes of the \emph{soft intersection circuits} $\mathcal{C}^{(\tau)}$ for different threshold values $\tau$ for all models are shown in Figures~\ref{fig:soft_qwen_computation}, \ref{fig:soft_qwen-math_computation}, \ref{fig:soft_llama_computation}, and \ref{fig:soft_phi_computation}, respectively. The final circuits are visualized in Figures~\ref{fig:qwen_computation_circuit}, \ref{fig:qwen_math_computation_circuit}, \ref{fig:llama_computation_circuit}, and \ref{fig:phi_computation_circuit}.

\begin{table}[b!]
\small
\centering
\begin{tabular}{lcc}
\toprule
Error Type & \textbf{Qwen-2.5-32B} & \textbf{Llama-3.1-70B} \\
\midrule
\textit{Arithmetic Result} & 98.93 $\pm$ 1.57 & 100.00 $\pm$ 0.00\\
\midrule
\textit{Numeric Answer} & 99.51 $\pm$ 0.09 & 100.00 $\pm$ 0.00\\
\midrule
\textit{Result \& Answer} & 80.13 $\pm$ 9.65 & 44.76 $\pm$ 10.50\\
\bottomrule
\end{tabular}
\caption{Accuracy of bigger models in correctly classifying the solutions' validity of (\emph{clean}, \emph{corrupt}) prompt pairs with different error types. Values represent the mean accuracy across all templates, reported with their corresponding standard deviation.}
\label{tab:big_model_behavioral_accuracy}
\end{table}

\subsection{Accuracy of Linear Probes}
The results of the probing experiment, detailed in Section~\ref{subsec:Dissociation} and Appendix~\ref{app:linear_probe}, are presented for Qwen-2.5-Math-1.5B-Instruct, Llama-3.2-3B-Instruct, and Phi-3-Mini-4k-Instruct in Figure~\ref{fig:probing-full}. The findings indicate that for all models, near-perfect accuracy is achieved by probes trained on the hidden representations of the residual stream in the upper layers. Notably, Phi-3-Mini-4k-Instruct is the only model that demonstrates significant probe accuracy in the middle layers.

\subsection{Residual Stream Patching}
\begin{figure*}[htbp]
    \centering
    \begin{subfigure}[b]{0.32\textwidth}
        \centering
        \includegraphics[width=\linewidth]{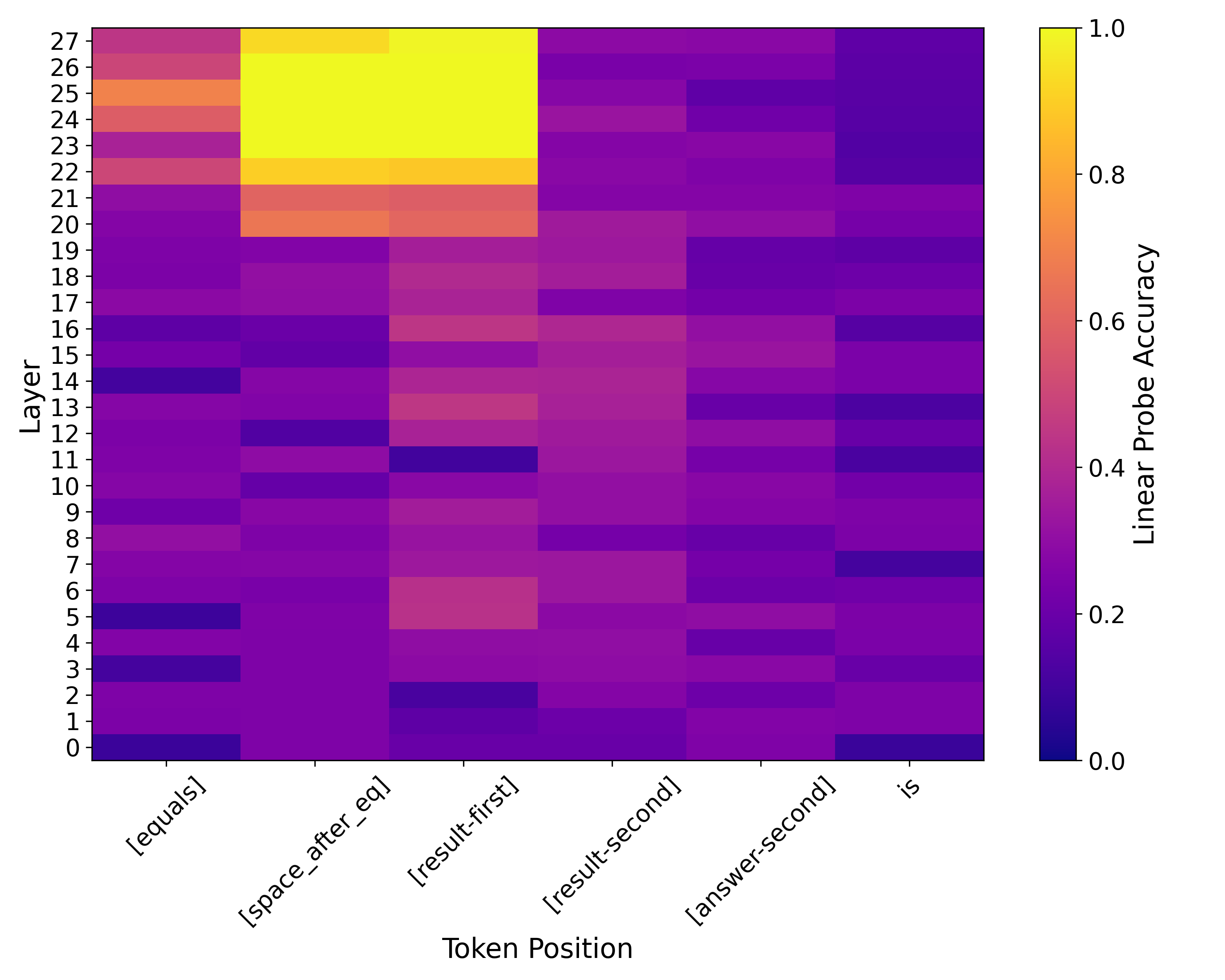}
        \caption{Qwen-2.5-Math-1.5B-Instruct}
        \label{fig:probing-qwen-math}
    \end{subfigure}
    \begin{subfigure}[b]{0.32\textwidth}
        \centering
        \includegraphics[width=\linewidth]{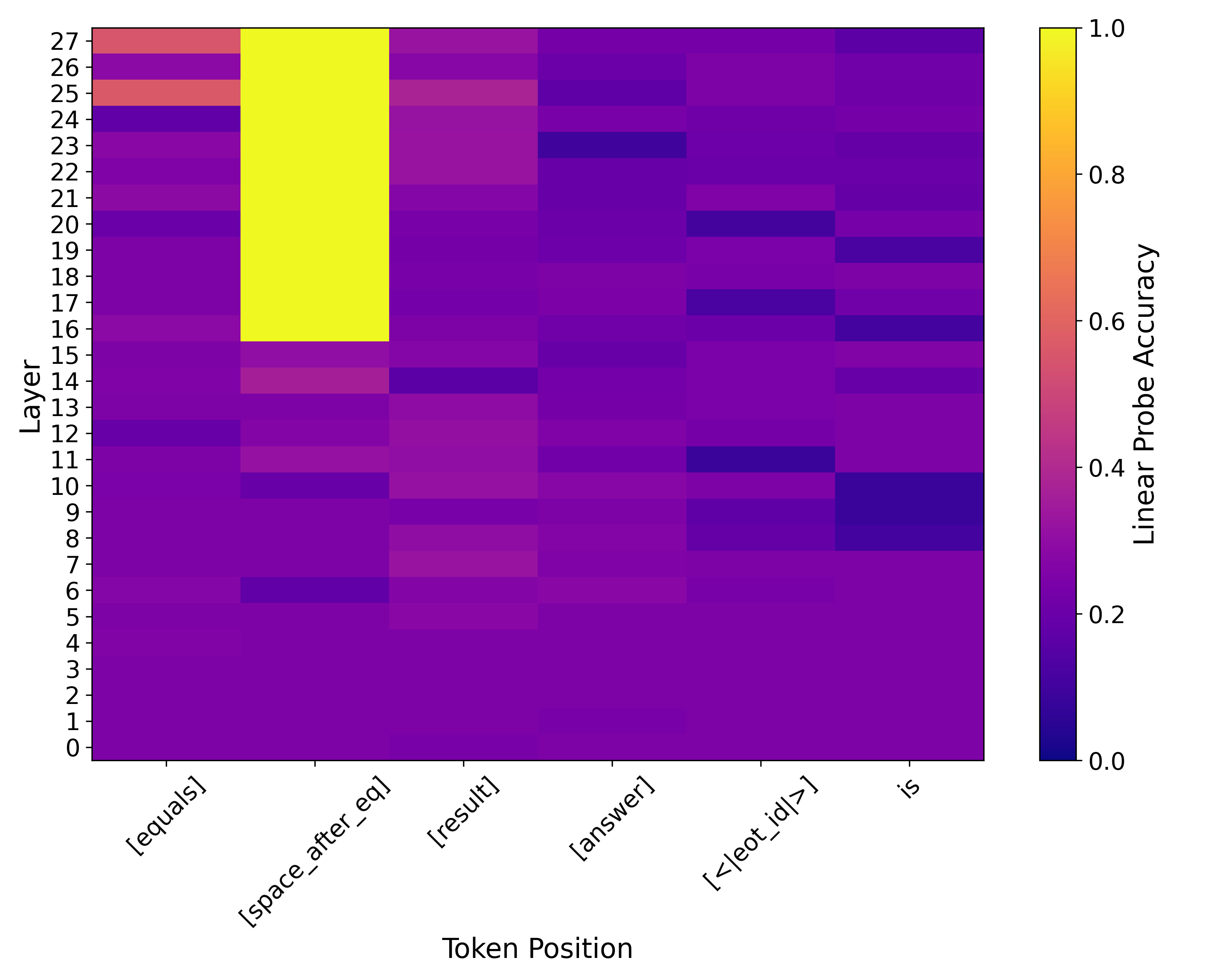}
        \caption{Llama-3.2-3B-Instruct}
        \label{fig:probing-llama}
    \end{subfigure}
    \begin{subfigure}[b]{0.32\textwidth}
        \centering
        \includegraphics[width=\linewidth]{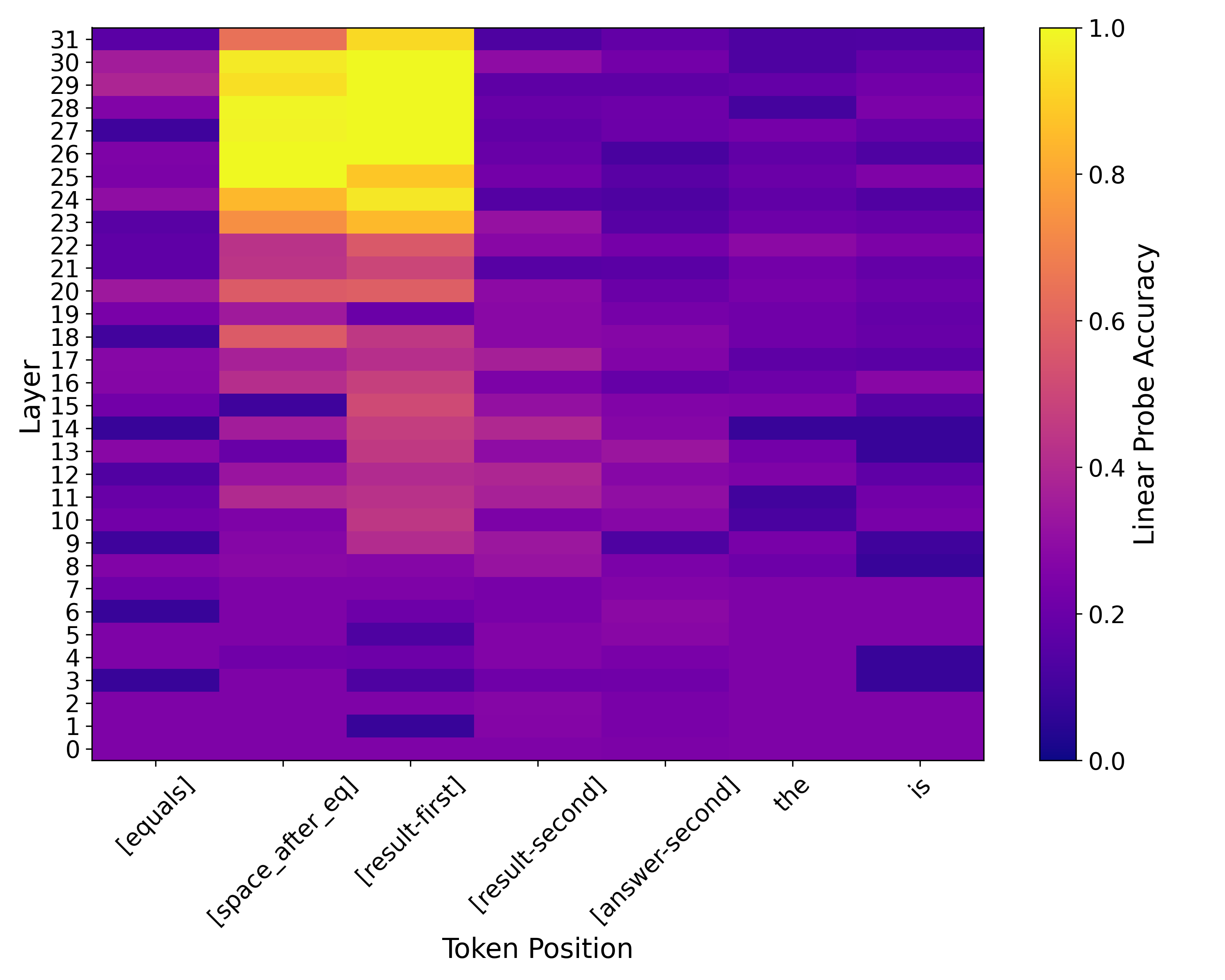}
        \caption{Phi-3-Mini-4k-Instruct}
        \label{fig:probing-phi}
    \end{subfigure}
    
    \caption{The linear probe's accuracy across all layers of Qwen-2.5-Math-1.5B-Instruct (Figure~\ref{fig:probing-qwen-math}), Llama-3.2-3B-Instruct (Figure~\ref{fig:probing-llama}), and Phi-3-Mini-4k-Instruct (Figure~\ref{fig:probing-phi}) at selected token positions. Only in higher layers, the probe is able to achieve predict the correct arithmetic result perfectly. For Qwen-2.5-Math-1.5B-Instruct and Phi-3-Mini-4k-Instruct, we observe moderate accuracies also in middle layers.}
    \label{fig:probing-full}
\end{figure*}

\begin{figure*}[htbp]
    \centering
    \begin{subfigure}[b]{0.32\textwidth}
        \centering
        \includegraphics[width=\linewidth]{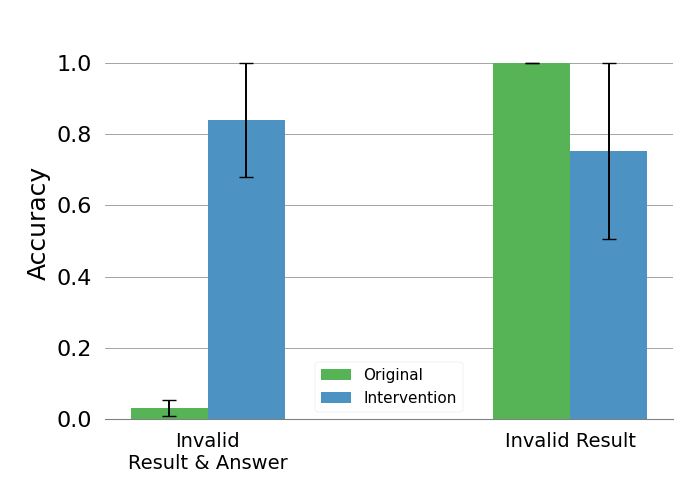}
        \caption{Qwen-2.5-Math-1.5B-Instruct}
        \label{fig:residual-patching-qwen-math}
    \end{subfigure}
    \begin{subfigure}[b]{0.32\textwidth}
        \centering
        \includegraphics[width=\linewidth]{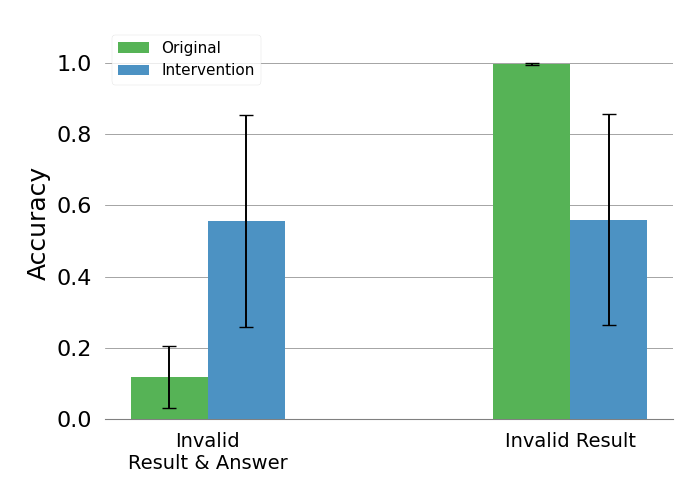}
        \caption{Llama-3.2-3B-Instruct}
        \label{fig:residual-patching-llama}
    \end{subfigure}
    \begin{subfigure}[b]{0.32\textwidth}
        \centering
        \includegraphics[width=\linewidth]{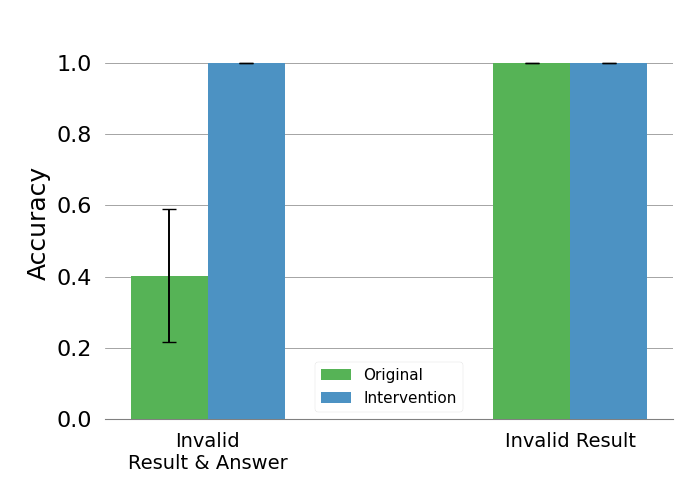}
        \caption{Phi-3-Mini-4k-Instruct}
        \label{fig:residual-patching-phi}
    \end{subfigure}
    \caption{Accuracy of models on (\emph{clean}, \emph{corrupt}) prompt pairs where (\emph{left}) the \emph{clean} prompt contains a consistent error at both error positions (invalid result \& answer), and (\emph{right}) an error is present only at the position of the arithmetic result (invalid result). The blue intervention bar denotes the result after adding the hidden representation of the residual stream in layer higher layers (at [result-first]) to the residual stream in lower layers (at [result-second]). For Qwen-2.5-Math-1.5B-Instruct (Figure~\ref{fig:residual-patching-qwen-math}), the residual steams' hidden representation from layer 22 is added to the one in layer 1. For Llama-3.2-3B-Instruct (Figure~\ref{fig:residual-patching-llama}), we add the representation from layer 16 to layer 2, and for Phi-3-Mini-4k-Instruct (Figure~\ref{fig:residual-patching-phi}), the representation from layer 24 is added to layer 1.}
    \label{fig:residual-patching-full}
\end{figure*}

To bridge the gap between the models' circuits responsible for arithmetic computation and validation, we add the hidden representation from higher layers\textemdash{}where the correct arithmetic result is linearly encoded (see Figure~\ref{fig:probing-full}\textemdash{}to lower layers. Specifically, for Qwen-2.5-Math-1.5B-Instruct, we intervene on layer 1 using the hidden representation from layer 22. For Llama-3.2-3B-Instruct, the intervention is performed on layer 2 using the hidden representation from layer 16, while for Phi-3-Mini-4k-Instruct, layer 1 is modified using the hidden representation from layer 24. The results of these interventions are depicted in Figure~\ref{fig:residual-patching-full}.

Qwen-2.5-1.5B-Instruct, Qwen-2.5-Math-1.5B-Instruct, and Phi-3-Mini-4k-Instruct exhibit significant improvements in accuracy following these interventions. In contrast, Llama-3.2-3B-Instruct demonstrates a more modest performance gain. We attribute this difference to the simplicity of our approach and consider this a promising direction for further investigation.

\section{Implementation Details}\label{app:implementation_details}
For the majority of our circuit identification experiments, we used the \texttt{AutoCircuit} library developed by~\citet{miller2024autocircuit}. For the remaining experiments, we relied on the \texttt{TransformerLens} library by~\citet{nanda2022transformerlens}. All models were loaded with bfloat16 precision. The experiments were conducted on a single A100 GPU with 80GB of memory, consuming approximately 350 GPU hours in total. Additionally, GitHub Copilot was used as an assistant tool for parts of the project’s source code development, and ChatGPT was used to correct minor grammatical errors.

{
\renewcommand{\arraystretch}{1.3}
\begin{table*}[htbp]
    \centering
    \begin{tabular}{|p{0.95\linewidth}|}
    \hline
    \textbf{Templates 1-8} \\
    \hline
    \texttt{[instruction]} Problem: \texttt{[person]} has \texttt{[num1]} \texttt{[object]}. \texttt{[pronoun]} \texttt{[verb]} \texttt{[num2]} more \texttt{[object]}. How many \texttt{[object]} does \texttt{[pronoun]} have now? 
    
    Reasoning: \texttt{[person]} has \texttt{[num1]} + \texttt{[num2]} = \texttt{[num3]} \texttt{[object]}. So, \texttt{[pronoun]} has \texttt{[num3]} \texttt{[object]} in total. 
    
    Answer: The above reasoning is \\
    \hline
    \texttt{[instruction]} Problem: \texttt{[person]} starts with \texttt{[num1]} \texttt{[object]}. After \texttt{[pronoun]} \texttt{[verb]} \texttt{[num2]} more, how many \texttt{[object]} does \texttt{[pronoun]} have in total?
    
    Reasoning: To solve this, we add \texttt{[num1]} and \texttt{[num2]}: \texttt{[num1]} + \texttt{[num2]} = \texttt{[num3]}. Therefore, \texttt{[person]} now has \texttt{[num3]} \texttt{[object]}.
    
    Answer: The above reasoning is \\
    \hline
    \texttt{[instruction]} Problem: Initially, \texttt{[person]} possesses \texttt{[num1]} \texttt{[object]}. \texttt{[pronoun]} then \texttt{[verb]} \texttt{[num2]} additional \texttt{[object]}. What's the new total amount of \texttt{[object]} that \texttt{[pronoun]} has?
    
    Reasoning: We calculate: \texttt{[num1]} (original) + \texttt{[num2]} (added) = \texttt{[num3]} (total). So, \texttt{[person]} now has \texttt{[num3]} \texttt{[object]}.
    
    Answer: The above reasoning is \\
    \hline
    \texttt{[instruction]} Problem: \texttt{[person]}'s collection of \texttt{[object]} grows from \texttt{[num1]} to an unknown amount after \texttt{[pronoun]} \texttt{[verb]} \texttt{[num2]} more.
    
    Reasoning: To find the new total, we add: \texttt{[num1]} + \texttt{[num2]} = \texttt{[num3]} (final amount). Thus, \texttt{[person]} ends up with \texttt{[num3]} \texttt{[object]}.
    
    Answer: The above reasoning is \\
    \hline
    \texttt{[instruction]} Problem: \texttt{[person]} originally owns \texttt{[num1]} \texttt{[object]}. After \texttt{[pronoun]} \texttt{[verb]} \texttt{[num2]} additional \texttt{[object]}, how many does \texttt{[pronoun]} have altogether?
    
    Reasoning: a simple addition gives us \texttt{[num1]} + \texttt{[num2]} = \texttt{[num3]}. Therefore, \texttt{[person]} has \texttt{[num3]} \texttt{[object]} now.
    
    Answer: The above reasoning is \\
    \hline
    \texttt{[instruction]} Problem: \texttt{[person]} possesses \texttt{[num1]} \texttt{[object]} at first. If \texttt{[pronoun]} \texttt{[verb]} \texttt{[num2]} more \texttt{[object]}, what is the total count?
    
    Reasoning: Adding them gives: \texttt{[num1]} + \texttt{[num2]} = \texttt{[num3]}. Consequently, \texttt{[person]} has a total of \texttt{[num3]} \texttt{[object]}.
    
    Answer: The above reasoning is \\
    \hline
    \texttt{[instruction]} Problem: \texttt{[num1]} \texttt{[object]} belong to \texttt{[person]}. \texttt{[pronoun]} \texttt{[verb]} \texttt{[num2]} additional ones. What’s the total?
    
    Reasoning: By addition, we get \texttt{[num1]} + \texttt{[num2]} = \texttt{[num3]}. Thus, \texttt{[person]} has \texttt{[num3]} \texttt{[object]} in total.
    
    Answer: The above reasoning is \\
    \hline
    \texttt{[instruction]} Problem: \texttt{[person]} begins with \texttt{[num1]} \texttt{[object]} and then \texttt{[verb]} \texttt{[num2]} more. How many \texttt{[object]} does \texttt{[pronoun]} have now?
    
    Reasoning: Let's add them up: \texttt{[num1]} + \texttt{[num2]} = \texttt{[num3]}. Therefore, \texttt{[person]} has a total of \texttt{[num3]} \texttt{[object]}.
    
    Answer: The above reasoning is \\
    \hline
    \end{tabular}
    \caption{The 8 problem templates including \texttt{[instruction]}, \texttt{[person]}, \texttt{[object]}, \texttt{[pronoun]}, \texttt{[num1]}, \texttt{[num2]}, \texttt{[num3]} as variable components. While samples within a template contain the same number of tokens, samples across templates vary in length due to differences in their non-variable parts.}
    \label{tab:templates}
\end{table*}
}

{
\renewcommand{\arraystretch}{1.5}
\begin{table*}[htbp]
    \centering
    \scriptsize
    \begin{tabular}{|p{0.31\linewidth}|p{0.31\linewidth}|p{0.31\linewidth}|}
    \hline
    {\small \textbf{Subtraction Templates 1-8}} & {\small \textbf{Multiplication Templates 1-8}} & {\small \textbf{Division Templates 1-8}} \\
    \hline
    
    \texttt{[instruction]} Problem: \texttt{[person]} has \texttt{[num1]} \texttt{[object]}. \texttt{[pronoun]} \texttt{[verb]} \texttt{[num2]} \texttt{[object]}. How many \texttt{[object]} does \texttt{[pronoun]} have now? Reasoning: \texttt{[person]} has \texttt{[num1]} - \texttt{[num2]} = \texttt{[num3]} \texttt{[object]}. So, \texttt{[pronoun]} has \texttt{[num3]} \texttt{[object]} remaining. Answer: The above reasoning is
    
    & 
    
    \texttt{[instruction]} Problem: \texttt{[person]} has \texttt{[num1]} \texttt{[object]} per day. After \texttt{[num2]} days, how many \texttt{[object]} does \texttt{[pronoun]} have in total? Reasoning: \texttt{[person]} has \texttt{[num1]} × \texttt{[num2]} = \texttt{[num3]} \texttt{[object]}. So, \texttt{[pronoun]} has \texttt{[num3]} \texttt{[object]} in total. Answer: The above reasoning is
    
    &
    
    \texttt{[instruction]} Problem: \texttt{[person]} has \texttt{[num1]} \texttt{[object]}. \texttt{[pronoun]} wants to organize them into equal groups of \texttt{[num2]} \texttt{[object]} each. How many groups can \texttt{[pronoun]} make? Reasoning: \texttt{[person]} can make \texttt{[num1]} ÷ \texttt{[num2]} = \texttt{[num3]} groups. So, \texttt{[pronoun]} can make \texttt{[num3]} groups. Answer: The above reasoning is \\
    \hline
    
    \texttt{[instruction]} Problem: \texttt{[person]} starts with \texttt{[num1]} \texttt{[object]}. After \texttt{[pronoun]} \texttt{[verb]} \texttt{[num2]}, how many \texttt{[object]} does \texttt{[pronoun]} have left? Reasoning: To solve this, we subtract \texttt{[num2]} from \texttt{[num1]}: \texttt{[num1]} - \texttt{[num2]} = \texttt{[num3]}. Therefore, \texttt{[person]} now has \texttt{[num3]} \texttt{[object]}. Answer: The above reasoning is
    
    &
    
    \texttt{[instruction]} Problem: \texttt{[person]} buys \texttt{[num1]} \texttt{[object]} each time \texttt{[pronoun]} goes shopping. If \texttt{[pronoun]} goes shopping \texttt{[num2]} times, how many \texttt{[object]} does \texttt{[pronoun]} buy in total? Reasoning: To solve this, we multiply \texttt{[num1]} and \texttt{[num2]}: \texttt{[num1]} × \texttt{[num2]} = \texttt{[num3]}. Therefore, \texttt{[person]} buys \texttt{[num3]} \texttt{[object]} in total. Answer: The above reasoning is
    
    &
    
    \texttt{[instruction]} Problem: \texttt{[person]} starts with \texttt{[num1]} \texttt{[object]}. If \texttt{[pronoun]} puts \texttt{[num2]} \texttt{[object]} in each container, how many containers can \texttt{[pronoun]} fill? Reasoning: To solve this, we divide \texttt{[num1]} by \texttt{[num2]}: \texttt{[num1]} ÷ \texttt{[num2]} = \texttt{[num3]}. Therefore, \texttt{[person]} can fill \texttt{[num3]} containers. Answer: The above reasoning is \\
    \hline
    
    \texttt{[instruction]} Problem: Initially, \texttt{[person]} possesses \texttt{[num1]} \texttt{[object]}. \texttt{[pronoun]} then \texttt{[verb]} \texttt{[num2]} \texttt{[object]}. What's the remaining amount of \texttt{[object]} that \texttt{[pronoun]} has? Reasoning: We calculate: \texttt{[num1]} (original) - \texttt{[num2]} (removed) = \texttt{[num3]} (remaining). So, \texttt{[person]} now has \texttt{[num3]} \texttt{[object]}. Answer: The above reasoning is
    
    &
    
    \texttt{[instruction]} Problem: \texttt{[person]} collects \texttt{[num1]} \texttt{[object]} each week. After \texttt{[num2]} weeks, how many \texttt{[object]} has \texttt{[pronoun]} collected altogether? Reasoning: We calculate: \texttt{[num1]} (per week) × \texttt{[num2]} (weeks) = \texttt{[num3]} (total). So, \texttt{[pronoun]} has collected \texttt{[num3]} \texttt{[object]} altogether. Answer: The above reasoning is
    
    &
    
    \texttt{[instruction]} Problem: \texttt{[person]} has \texttt{[num1]} \texttt{[object]} to share equally. If each person gets \texttt{[num2]} \texttt{[object]}, how many people can receive \texttt{[object]}? Reasoning: We calculate: \texttt{[num1]} (total) ÷ \texttt{[num2]} (per person) = \texttt{[num3]} (people). So, \texttt{[num3]} people can receive \texttt{[object]}. Answer: The above reasoning is \\
    \hline
    
    \texttt{[instruction]} Problem: \texttt{[person]}'s collection of \texttt{[object]} decreases from \texttt{[num1]} to an unknown amount after \texttt{[pronoun]} \texttt{[verb]} \texttt{[num2]} \texttt{[object]}. Reasoning: To find the remaining total, we subtract: \texttt{[num1]} - \texttt{[num2]} = \texttt{[num3]} (final amount). Thus, \texttt{[person]} ends up with \texttt{[num3]} \texttt{[object]}. Answer: The above reasoning is
    
    &
    
    \texttt{[instruction]} Problem: Initially, \texttt{[person]} receives \texttt{[num1]} \texttt{[object]} each month. After \texttt{[num2]} months, what's the total amount of \texttt{[object]} that \texttt{[pronoun]} has received? Reasoning: We calculate: \texttt{[num1]} (per month) × \texttt{[num2]} (months) = \texttt{[num3]} (total). So, \texttt{[person]} has received \texttt{[num3]} \texttt{[object]}. Answer: The above reasoning is
    
    &
    
    \texttt{[instruction]} Problem: Initially, \texttt{[person]} possesses \texttt{[num1]} \texttt{[object]}. \texttt{[pronoun]} wants to arrange them in rows with \texttt{[num2]} \texttt{[object]} per row. What's the total number of rows that can be formed? Reasoning: We calculate: \texttt{[num1]} (total) ÷ \texttt{[num2]} (per row) = \texttt{[num3]} (rows). So, \texttt{[person]} can form \texttt{[num3]} rows. Answer: The above reasoning is \\
    \hline
    
    \texttt{[instruction]} Problem: \texttt{[person]} originally owns \texttt{[num1]} \texttt{[object]}. After \texttt{[pronoun]} \texttt{[verb]} \texttt{[num2]} \texttt{[object]}, how many does \texttt{[pronoun]} have left? Reasoning: A simple subtraction gives us \texttt{[num1]} - \texttt{[num2]} = \texttt{[num3]}. Therefore, \texttt{[person]} has \texttt{[num3]} \texttt{[object]} remaining. Answer: The above reasoning is
    
    &
    
    \texttt{[instruction]} Problem: \texttt{[person]} originally gets \texttt{[num1]} \texttt{[object]} per visit. After \texttt{[num2]} visits, how many \texttt{[object]} has \texttt{[pronoun]} gotten altogether? Reasoning: A simple multiplication gives us \texttt{[num1]} × \texttt{[num2]} = \texttt{[num3]}. Therefore, \texttt{[person]} has gotten \texttt{[num3]} \texttt{[object]} in total. Answer: The above reasoning is
    
    &
    
    \texttt{[instruction]} Problem: \texttt{[person]} originally owns \texttt{[num1]} \texttt{[object]}. If \texttt{[pronoun]} distributes \texttt{[num2]} \texttt{[object]} to each recipient, how many recipients can get \texttt{[object]}? Reasoning: A simple division gives us \texttt{[num1]} ÷ \texttt{[num2]} = \texttt{[num3]}. Therefore, \texttt{[num3]} recipients can get \texttt{[object]}. Answer: The above reasoning is \\
    \hline
    
    \texttt{[instruction]} Problem: \texttt{[person]} possesses \texttt{[num1]} \texttt{[object]} at first. If \texttt{[pronoun]} \texttt{[verb]} \texttt{[num2]} \texttt{[object]}, what is the remaining count? Reasoning: Subtracting them gives: \texttt{[num1]} - \texttt{[num2]} = \texttt{[num3]}. Consequently, \texttt{[person]} has \texttt{[num3]} \texttt{[object]} left. Answer: The above reasoning is
    
    &
    
    \texttt{[instruction]} Problem: \texttt{[person]} earns \texttt{[num1]} \texttt{[object]} per task at first. If \texttt{[pronoun]} completes \texttt{[num2]} tasks, what is the total count of \texttt{[object]}? Reasoning: Multiplying them gives: \texttt{[num1]} × \texttt{[num2]} = \texttt{[num3]}. Consequently, \texttt{[pronoun]} earns \texttt{[num3]} \texttt{[object]} in total. Answer: The above reasoning is
    
    &
    
    \texttt{[instruction]} Problem: \texttt{[person]} possesses \texttt{[num1]} \texttt{[object]} at first. If \texttt{[pronoun]} arranges \texttt{[num2]} \texttt{[object]} in each pile, what is the total number of piles? Reasoning: Dividing them gives: \texttt{[num1]} ÷ \texttt{[num2]} = \texttt{[num3]}. Consequently, \texttt{[person]} can make \texttt{[num3]} piles. Answer: The above reasoning is \\
    \hline
    
    \texttt{[instruction]} Problem: \texttt{[num1]} \texttt{[object]} belong to \texttt{[person]}. \texttt{[pronoun]} \texttt{[verb]} \texttt{[num2]} of them. What's the remainder? Reasoning: By subtraction, we get \texttt{[num1]} - \texttt{[num2]} = \texttt{[num3]}. Thus, \texttt{[person]} has \texttt{[num3]} \texttt{[object]} remaining. Answer: The above reasoning is
    
    &
    
    \texttt{[instruction]} Problem: \texttt{[person]} finds \texttt{[num1]} \texttt{[object]} each time \texttt{[pronoun]} searches. After \texttt{[num2]} searches, what's the total? Reasoning: By multiplication, we get \texttt{[num1]} × \texttt{[num2]} = \texttt{[num3]}. Thus, \texttt{[person]} finds \texttt{[num3]} \texttt{[object]} in total. Answer: The above reasoning is
    
    &
    
    \texttt{[instruction]} Problem: \texttt{[num1]} \texttt{[object]} belong to \texttt{[person]}. \texttt{[pronoun]} places \texttt{[num2]} \texttt{[object]} in each box. What's the total number of boxes needed? Reasoning: By division, we get \texttt{[num1]} ÷ \texttt{[num2]} = \texttt{[num3]}. Thus, \texttt{[person]} needs \texttt{[num3]} boxes. Answer: The above reasoning is \\
    \hline
    
    \texttt{[instruction]} Problem: \texttt{[person]} begins with \texttt{[num1]} \texttt{[object]} and then \texttt{[verb]} \texttt{[num2]} of them. How many \texttt{[object]} does \texttt{[pronoun]} have left? Reasoning: Let's subtract them: \texttt{[num1]} - \texttt{[num2]} = \texttt{[num3]}. Therefore, \texttt{[person]} has \texttt{[num3]} \texttt{[object]} remaining. Answer: The above reasoning is
    
    &
    
    \texttt{[instruction]} Problem: \texttt{[person]} produces \texttt{[num1]} \texttt{[object]} per session and then has \texttt{[num2]} sessions. How many \texttt{[object]} are there in total? Reasoning: Let's multiply them: \texttt{[num1]} × \texttt{[num2]} = \texttt{[num3]}. Therefore, there are \texttt{[num3]} \texttt{[object]} altogether. Answer: The above reasoning is
    
    &
    
    \texttt{[instruction]} Problem: \texttt{[person]} begins with \texttt{[num1]} \texttt{[object]} and then organizes \texttt{[num2]} \texttt{[object]} per shelf. How many shelves does \texttt{[pronoun]} need? Reasoning: Let's divide them: \texttt{[num1]} ÷ \texttt{[num2]} = \texttt{[num3]}. Therefore, \texttt{[person]} needs \texttt{[num3]} shelves. Answer: The above reasoning is \\
    \hline
    \end{tabular}
    \caption{Templates used for other arithmetic operations. We created 8 templates for each operation, including \texttt{[instruction]}, \texttt{[person]}, \texttt{[object]}, \texttt{[pronoun]}, \texttt{[num1]}, \texttt{[num2]}, \texttt{[num3]} as variable components. The subtraction template also has a \texttt{[verb]} variable, for which we use the following verbs: ``lost'', ``sold'', ``gave away'', ``donated'', ``threw away'',}
    \label{tab:operations_templates}
\end{table*}
}

{
\renewcommand{\arraystretch}{1.2}
\begin{table*}[htbp]
\centering
\begin{tabular}{|l|p{0.75\linewidth}|}
\hline
\textbf{Variable} & \textbf{Assignments} \\ \hline
\texttt{[person]} & 
Aaron, Adam, Alan, Alex, Alice, Amy, Anderson, Andre, Andrew, Andy, Anna, Anthony, Arthur, Austin, Blake, Brandon, Brian, Carter, Charles, Charlie, Christian, Christopher, Clark, Cole, Collins, Connor, Crew, Crystal, Daniel, David, Dean, Edward, Elizabeth, Emily, Eric, Eva, Ford, Frank, George, Georgia, Graham, Grant, Henry, Ian, Jack, Jacob, Jake, James, Jamie, Jane, Jason, Jay, Jennifer, Jeremy, Jessica, John, Jonathan, Jordan, Joseph, Joshua, Justin, Kate, Kelly, Kevin, Kyle, Laura, Leon, Lewis, Lisa, Louis, Luke, Madison, Marco, Marcus, Maria, Mark, Martin, Mary, Matthew, Max, Michael, Michelle, Morgan, Patrick, Paul, Peter, Prince, Rachel, Richard, River, Robert, Roman, Rose, Ruby, Russell, Ryan, Sarah, Scott, Sean, Simon, Stephen, Steven, Sullivan, Taylor, Thomas, Tyler, Victoria, Warren, William \\ \hline
\texttt{[object]} & 
apples, bananas, oranges, grapes, pears, mangoes, strawberries, blueberries, cherries, pineapples, lemons, watermelons, kiwis, peaches, plums, books, pens, notebooks, flowers, candies, gifts, toys, bottles, tickets, clothes, shoes, hats, gloves, keys, wallets, phones, laptops, tablets, cameras, headphones, glasses, watches, rings, necklaces, bracelets, purses, backpacks, umbrellas, mugs, plates, bowls, forks, spoons, knives, chairs, tables, lamps, blankets, pillows, towels, socks, scarves, jackets, belts, bookshelves, paintings, mirrors, candles, frames \\ \hline
\texttt{[verb]} & 
won, bought, received, gained, obtained, earned, acquired, collected, accumulated, gathered, got \\ \hline
\texttt{[correct\_pair]} & 
``valid or invalid'', ``correct or incorrect'', ``right or wrong'' \\ \hline
\end{tabular}

\vspace{1em}

\begin{tabular}{|p{0.95\linewidth}|}
\hline
\texttt{[instruction]} \\
\hline
Does the following reasoning chain contain any mistakes? Determine whether it is \texttt{[correct\_pair]}.\\[0.5em]
Does the reasoning chain provided have any errors? Decide whether it is \texttt{[correct\_pair]}.\\[0.5em]
Does the given reasoning chain contain any flaws? Evaluate whether it is \texttt{[correct\_pair]}.\\[0.5em]
Does the reasoning chain shown have any errors? Verify whether it is \texttt{[correct\_pair]}.\\[0.5em]
Does the reasoning chain below have any mistakes? Check if it is \texttt{[correct\_pair]}.\\[0.5em]
Does the following reasoning chain have any errors? Specify whether it is \texttt{[correct\_pair]}.\\[0.5em]
Does the provided reasoning chain contain any flaws? Assess if it is \texttt{[correct\_pair]}.\\[0.5em]
Does the reasoning chain presented have any issues? Judge whether it is \texttt{[correct\_pair]}.\\[0.5em]
Does the reasoning chain contain any mistakes? Examine if it is \texttt{[correct\_pair]}.\\[0.5em]
Does the reasoning chain have any errors? Inspect it and determine if it is \texttt{[correct\_pair]}.\\[0.5em]
Does the reasoning chain have any flaws? Review it and confirm if it is \texttt{[correct\_pair]}.\\[0.5em]
Does the given reasoning chain contain any issues? Analyze it and decide if it is \texttt{[correct\_pair]}.\\ \hline
\end{tabular}

\caption{The full set of variables and possible values that are used to create the data. \texttt{[instruction]} is the only variable that contains \texttt{[correct\_pair]} as another variable part.}
\label{tab:template_variables}
\end{table*}
}

\begin{table*}[htbp]
    \centering
    \begin{minipage}[t]{0.48\textwidth}
        \customsize
        \centering
        \resizebox{\textwidth}{!}{
        \begin{tabular}{cccccc}
            \toprule
             & \textbf{Template} & \textbf{Task} & \textbf{Total Edges} & \textbf{Num Edges} & \textbf{Faithfulness} \\
            \midrule
            \multirow{24}{*}{\rotatebox{90}{Qwen-2.5-1.5B-Instruct}}
                & \multirow{3}{*}{1} & Inv. Result          & 84715  & 281 & 100.00 \\
                &                   & Inv. Answer          & 84715  & 101 & 100.57 \\
                \cmidrule{3-6}
                &                   & Computation         & 84715  & 140 & 99.04 \\
            \cmidrule{2-6}
                & \multirow{3}{*}{2} & Inv. Result          & 84715  & 440 & 100.40 \\
                &                   & Inv. Answer          & 84715  & 115 & 100.00 \\
                \cmidrule{3-6}
                &                   & Computation         & 84715  & 100 & 99.21 \\
            \cmidrule{2-6}
                & \multirow{3}{*}{3} & Inv. Result          & 84715  & 522 & 100.00 \\
                &                   & Inv. Answer          & 84715  & 120 & 100.00 \\
                \cmidrule{3-6}
                &                   & Computation         & 84715  & 300 & 99.13 \\
            \cmidrule{2-6}
                & \multirow{3}{*}{4} & Inv. Result          & 84715  & 261 & 100.49 \\
                &                   & Inv. Answer          & 84715  & 208 & 100.57 \\
                \cmidrule{3-6}
                &                   & Computation         & 84715  & 104 & 99.20 \\
            \cmidrule{2-6}
                & \multirow{3}{*}{5} & Inv. Result          & 84715  & 721 & 100.00 \\
                &                   & Inv. Answer          & 84715  & 120 & 99.29 \\
                \cmidrule{3-6}
                &                   & Computation         & 84715  & 280 & 100.00 \\
            \cmidrule{2-6}
                & \multirow{3}{*}{6} & Inv. Result          & 84715  & 142 & 99.47 \\
                &                   & Inv. Answer          & 84715  & 163 & 100.41 \\
                \cmidrule{3-6}
                &                   & Computation         & 84715  & 120 & 99.15 \\
            \cmidrule{2-6}
                & \multirow{3}{*}{7} & Inv. Result          & 84715  & 200 & 99.42 \\
                &                   & Inv. Answer          & 84715  & 104 & 100.67 \\
                \cmidrule{3-6}
                &                   & Computation         & 84715  & 260 & 99.60 \\
            \cmidrule{2-6}
                & \multirow{3}{*}{8} & Inv. Result          & 84715  & 201 & 99.52 \\
                &                   & Inv. Answer          & 84715  & 112 & 100.60 \\
                \cmidrule{3-6}
                &                   & Computation         & 84715  & 140 & 99.15 \\
            \midrule
            \multirow{24}{*}{\rotatebox{90}{Qwen-2.5-Math-1.5B-Instruct}} 
                & \multirow{3}{*}{1} & Inv. Result          & 84715  & 141 & 99.46 \\
                &                   & Inv. Answer          & 84715  & 200 & 99.45 \\
                \cmidrule{3-6}
                &                   & Computation         & 84715  & 101 & 99.27 \\
            \cmidrule{2-6}
                & \multirow{3}{*}{2} & Inv. Result          & 84715  & 241 & 100.00 \\
                &                   & Inv. Answer          & 84715  & 101 & 100.00 \\
                \cmidrule{3-6}
                &                   & Computation         & 84715  & 100 & 99.27 \\
            \cmidrule{2-6}
                & \multirow{3}{*}{3} & Inv. Result          & 84715  & 340 & 100.00 \\
                &                   & Inv. Answer          & 84715  & 100 & 100.00 \\
                \cmidrule{3-6}
                &                   & Computation         & 84715  & 101 & 99.56 \\
            \cmidrule{2-6}
                & \multirow{3}{*}{4} & Inv. Result          & 84715  & 320 & 99.01 \\
                &                   & Inv. Answer          & 84715  & 100 & 100.00 \\
                \cmidrule{3-6}
                &                   & Computation         & 84715  & 120 & 99.26 \\
            \cmidrule{2-6}
                & \multirow{3}{*}{5} & Inv. Result          & 84715  & 318 & 99.42 \\
                &                   & Inv. Answer          & 84715  & 102 & 99.40 \\
                \cmidrule{3-6}
                &                   & Computation         & 84715  & 122 & 99.61 \\
            \cmidrule{2-6}
                & \multirow{3}{*}{6} & Inv. Result          & 84715  & 102 & 99.47 \\
                &                   & Inv. Answer          & 84715  & 187 & 99.45 \\
                \cmidrule{3-6}
                &                   & Computation         & 84715  & 100 & 99.53 \\
            \cmidrule{2-6}
                & \multirow{3}{*}{7} & Inv. Result          & 84715  & 321 & 99.53 \\
                &                   & Inv. Answer          & 84715  & 323 & 99.53 \\
                \cmidrule{3-6}
                &                   & Computation         & 84715  & 100 & 99.59 \\
            \cmidrule{2-6}
                & \multirow{3}{*}{8} & Inv. Result          & 84715  & 321 & 99.42 \\
                &                   & Inv. Answer          & 84715  & 100 & 100.00 \\
                \cmidrule{3-6}
                &                   & Computation         & 84715  & 100 & 99.55 \\
            \bottomrule
        \end{tabular}
        }
    \end{minipage}
    \hfill
    \begin{minipage}[t]{0.48\textwidth}
        \customsize
        \centering
        \resizebox{\textwidth}{!}{
        \begin{tabular}{cccccc}
            \toprule
             & \textbf{Template} & \textbf{Task} & \textbf{Total Edges} & \textbf{Num Edges} & \textbf{Faithfulness} \\
            \midrule
            \multirow{24}{*}{\rotatebox{90}{Llama-3.2-3B-Instruct}}
                & \multirow{3}{*}{1} & Inv. Result & 389971 & 190 & 99.28 \\
                &                   & Inv. Answer & 389971 & 294 & 100.00 \\
                \cmidrule{3-6}
                &                   & Computation & 389971 & 160 & 99.10 \\
            \cmidrule{2-6}
                & \multirow{3}{*}{2} & Inv. Result & 389971 & 380 & 99.50 \\
                &                   & Inv. Answer & 389971 & 282 & 100.00 \\
                \cmidrule{3-6}
                &                   & Computation & 389971 & 100 & 99.14 \\
            \cmidrule{2-6}
                & \multirow{3}{*}{3} & Inv. Result & 389971 & 187 & 99.10 \\
                &                   & Inv. Answer & 389971 & 280 & 99.52 \\
                \cmidrule{3-6}
                &                   & Computation & 389971  & 100 & 100.00 \\
            \cmidrule{2-6}
                & \multirow{3}{*}{4} & Inv. Result & 389971 & 180 & 99.32 \\
                &                   & Inv. Answer & 389971 & 382 & 99.25 \\
                \cmidrule{3-6}
                &                   & Computation & 389971 & 118 & 99.59 \\
            \cmidrule{2-6}
                & \multirow{3}{*}{5} & Inv. Result & 389971 & 220 & 99.58 \\
                &                   & Inv. Answer & 389971 & 241 & 100.00 \\
                \cmidrule{3-6}
                &                   & Computation & 389971 & 160 & 99.22 \\
            \cmidrule{2-6}
                & \multirow{3}{*}{6} & Inv. Result & 389971 & 443 & 99.28 \\
                &                   & Inv. Answer & 389971 & 240 & 100.00 \\
                \cmidrule{3-6}
                &                   & Computation & 389971 & 100 & 100.00 \\
            \cmidrule{2-6}
                & \multirow{3}{*}{7} & Inv. Result & 389971 & 180 & 99.04 \\
                &                   & Inv. Answer & 389971 & 561 & 100.52 \\
                \cmidrule{3-6}
                &                   & Computation & 389971 & 180 & 99.09 \\
            \cmidrule{2-6}
                & \multirow{3}{*}{8} & Inv. Result & 389971 & 221 & 99.21 \\
                &                   & Inv. Answer & 389971 & 200 & 99.19 \\
                \cmidrule{3-6}
                &                   & Computation & 389971 & 119 & 99.53 \\
            \midrule
            \multirow{24}{*}{\rotatebox{90}{Phi-3-Mini-4k-Instruct}}
                & \multirow{3}{*}{1} & Inv Result & 1592881 & 285 & 99.25 \\
                &                   & Inv. Answer & 1592881 & 581 & 99.24 \\
                \cmidrule{3-6}
                &                   & Computation & 1592881  & 161 & 99.46 \\
            \cmidrule{2-6}
                & \multirow{3}{*}{2} & Inv. Result & 1592881 & 500 & 99.25 \\
                &                   & Inv. Answer & 1592881 & 480 & 99.25 \\
                \cmidrule{3-6}
                &                   & Computation & 1592881 & 122 & 99.08 \\
            \cmidrule{2-6}
                & \multirow{3}{*}{3} & Inv. Result & 1592881 & 855 & 99.24 \\
                &                   & Inv. Answer & 1592881 & 683 & 99.25 \\
                \cmidrule{3-6}
                &                   & Computation & 1592881 & 240 & 99.47 \\
            \cmidrule{2-6}
                & \multirow{3}{*}{4} & Inv. Result & 1592881 & 371 & 99.15 \\
                &                   & Inv. Answer & 1592881 & 504 & 99.13 \\
                \cmidrule{3-6}
                &                   & Computation & 1592881 & 144 & 99.08 \\
            \cmidrule{2-6}
                & \multirow{3}{*}{5} & Inv. Result & 1592881 & 504 & 99.21 \\
                &                   & Inv. Answer & 1592881 & 500 & 99.20 \\
                \cmidrule{3-6}
                &                   & Computation & 1592881 & 145 & 99.49 \\
            \cmidrule{2-6}
                & \multirow{3}{*}{6} & Inv. Result & 1592881 & 569 & 99.22 \\
                &                   & Inv. Answer & 1592881 & 422 & 99.22 \\
                \cmidrule{3-6}
                &                   & Computation & 1592881 & 140 & 99.42 \\
            \cmidrule{2-6}
                & \multirow{3}{*}{7} & Inv. Result & 1592881 & 886 & 99.20 \\
                &                   & Inv. Answer & 1592881 & 605 & 99.21 \\
                \cmidrule{3-6}
                &                   & Computation & 1592881  & 120 & 99.03 \\
            \cmidrule{2-6}
                & \multirow{3}{*}{8} & Inv. Result & 1592881 & 625 & 99.22 \\
                &                   & Inv. Answer & 1592881 & 481 & 99.23 \\
                \cmidrule{3-6}
                &                   & Computation & 1592881 & 146 & 99.50 \\
            \bottomrule
        \end{tabular}
        }
    \end{minipage}
    \caption{The faithfulness score and the number of edges of the circuit identified for each model, template, and task. To better compare circuit sizes, we also present the total number of edges \emph{\textbf{per token position}} for each model. Left: Qwen models (Qwen-2.5 and Qwen-2.5-Math) from templates 1--8; Right: Llama-3.2 and Phi-3-Mini models from templates 1--8.}
    \label{tab:faith_size}
\end{table*}

\begin{figure*}[htbp]
    \centering
    \begin{subfigure}[b]{0.48\textwidth}
        \centering
        \includegraphics[width=\linewidth]{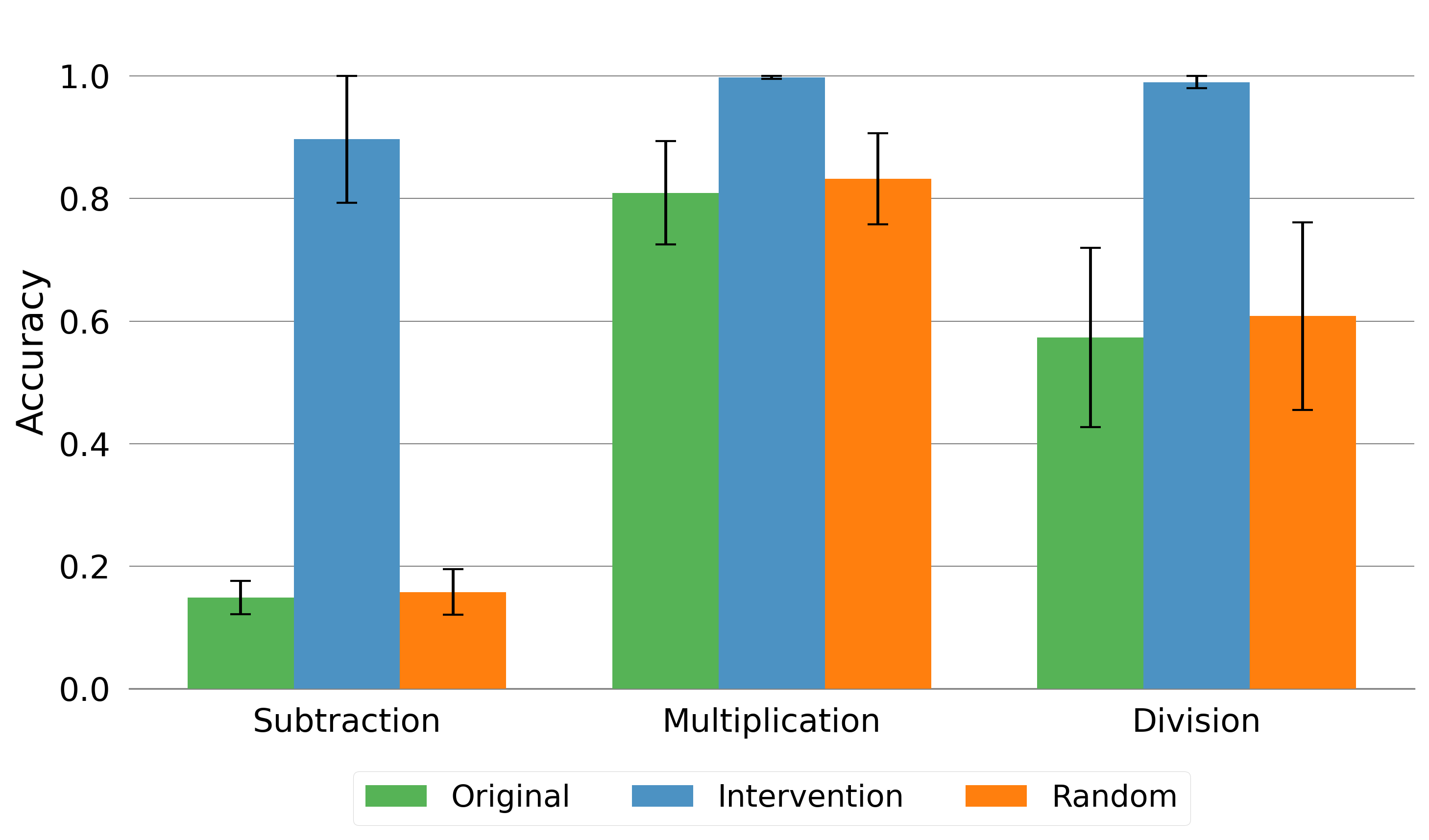}
        \caption{Qwen-2.5-Math-1.5B-Instruct}
    \end{subfigure}
    \begin{subfigure}[b]{0.48\textwidth}
        \centering
        \includegraphics[width=\linewidth]{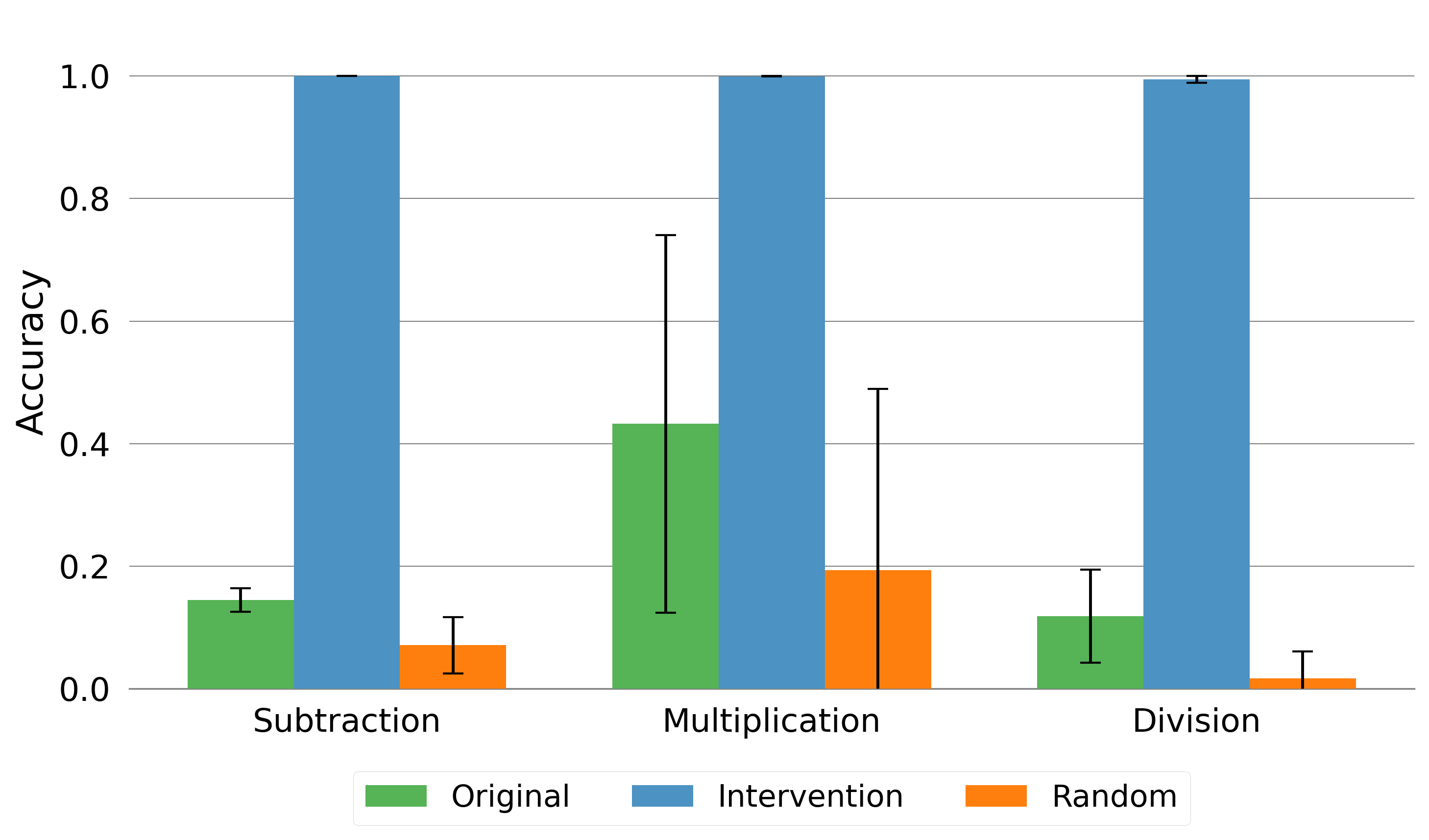}
        \caption{Llama-3.2-3B-Instruct}
    \end{subfigure}
    \caption{Accuracy of models on (\emph{clean}, \emph{corrupt}) prompt pairs for other arithmetic operations, where the \emph{clean} prompt contains a consistent error at both error positions. The blue intervention bar represents the result after patching a set of \emph{consistency heads} (for a list of heads, please refer to Table~\ref{tab:heads} in the Appendix). In contrast, the red bar indicates the accuracy after patching a set of randomly chosen attention heads that are not labeled as consistency heads.}
    \label{fig:attention-patching-operations-full}
\end{figure*}

\end{document}

%% file: math_commands.tex
\usepackage{amsmath,amsfonts,bm}

\def\eqref#1{equation~\ref{#1}}

\def\1{\bm{1}}

\def\vz{{\bm{z}}}

\DeclareMathAlphabet{\mathsfit}{\encodingdefault}{\sfdefault}{m}{sl}
\SetMathAlphabet{\mathsfit}{bold}{\encodingdefault}{\sfdefault}{bx}{n}



%% file: figures/tikz/intro_figure.tex
\begin{tikzpicture}[
  >=stealth,
  thick,
  box/.style={
    draw,
    rectangle,
    rounded corners=2pt,
    minimum width=7mm,
    minimum height=5mm,
    align=center,
    font=\sffamily
  },
  greenBox/.style={
    box,
    draw=cyan!50!gray,
    fill=cyan!50!gray,
    inner sep=0pt,
  },
  blackBox/.style={
    box,
    draw=black!20,
    fill=black!20
  },
  redBox/.style={
  box,
  draw=red!50!gray,    
  fill=red!50!gray,         
  inner sep=0pt,
  },
  hollowBox/.style={
    box,
    draw=white,
  },
  arrow/.style={
    ->,
    shorten >=1pt,
    shorten <=1pt
  }
]


\node[hollowBox] (g1) at (0,0)  {};
\node[hollowBox] (g2) [right=0.3 of g1] {\textbf{+}};
\node[hollowBox] (g3) [right=0.3 of g2] {};
\node[hollowBox] (g4) [right=0.3 of g3] {\textbf{=}};
\node[hollowBox] (g5) [right=0.3 of g4] {};
\node[cyan!50!black] at ($(g1)$) {\textbf{5}};
\node[cyan!50!black] at ($(g3)$) {\textbf{8}};
\node[red!50!black] at ($(g5)$) {\textbf{16}};

\node[blackBox] (g6)  at ($(g1)+(0,0.9)$)    {};
\node[blackBox] (g7)  [right=0.3 of g6]      {};
\node[blackBox] (g8)  [right=0.3 of g7]      {};
\node[blackBox] (g9)  [right=0.3 of g8]      {};
\node[redBox] (g10) [right=0.3 of g9]      {};

\node[hollowBox] (b1) at ($(g6)+(0,0.9)$) {...};
\node[hollowBox] (b2) [right=0.3 of b1]  {...};
\node[hollowBox] (b3) [right=0.3 of b2]  {...};
\node[hollowBox] (b4) [right=0.3 of b3]  {...};
\node[hollowBox] (b5) [right=0.3 of b4] {...};

\node[blackBox] (g11) at ($(b1)+(0,0.9)$)    {};
\node[blackBox] (g12) [right=0.3 of g11]     {};
\node[blackBox] (g13) [right=0.3 of g12]     {};
\node[blackBox] (g14) [right=0.3 of g13]     {};
\node[redBox] (g15) [right=0.3 of g14]     {};

\node[hollowBox] (b6) at ($(g11)+(0,0.9)$) {...};
\node[hollowBox] (b7) [right=0.3 of b6]  {...};
\node[hollowBox] (b8) [right=0.3 of b7]  {...};
\node[hollowBox] (b9) [right=0.3 of b8]  {...};
\node[hollowBox] (b10) [right=0.3 of b9] {...};

\node[greenBox] (g16) at ($(b6)+(0,0.9)$)   {};
\node[greenBox] (g17) [right=0.3 of g16]    {};
\node[greenBox] (g18) [right=0.3 of g17]    {};
\node[greenBox] (g19) [right=0.3 of g18]    {}; 
\node[blackBox] (g20) [right=0.3 of g19]    {}; 

\node[hollowBox] (g21) at ($(g16)+(0,0.9)$)  {};
\node[hollowBox] (g22) [right=0.3 of g21] {};
\node[hollowBox] (g23) [right=0.3 of g22] {};
\node[hollowBox] (g24) [right=0.3 of g23] {};
\node[hollowBox] (g25) [right=0.3 of g24] {};
\node[cyan!50!black] at ($(g24)$) {\textbf{13}};


\node[hollowBox] (d1) [right=0. of g5] {...};
\node[hollowBox] (d2) [right=0. of g10]  {...};
\node[hollowBox] (d3) [right=0. of g15]  {...};
\node[hollowBox] (d4) [right=0. of g20]  {...};
\node[hollowBox] (d5) [right=0. of g25] {};


\node[hollowBox] (r1) [right=0.0 of d1]  {\small Ans:};
\node[blackBox] (r2) [right=0.09 of d2]     {};
\node[hollowBox] (rb2) at ($(r2)+(0,0.9)$) {...};
\node[redBox] (r3) [right=0.09 of d3]     {};
\node[hollowBox] (rb3) at ($(r3)+(0,0.9)$) {...};
\node[blackBox] (r4) [right=0.09 of d4]     {};
\node[hollowBox] (r5) [right=0.09 of d5] {};

\node[hollowBox] (r6) [right=0.21 of r1]  {};
\node[redBox] (r7) [right=0.3 of r2]     {};
\node[hollowBox] (rb7) at ($(r7)+(0,0.9)$) {...};
\node[redBox] (r8) [right=0.3 of r3]     {};
\node[hollowBox] (rb8) at ($(r8)+(0,0.9)$) {...};
\node[redBox] (r9) [right=0.3 of r4]     {};
\node[hollowBox] (r10) [right=0.3 of r5] {};
\node[red!50!black] at ($(r6)$) {\textbf{13}};
\node[red!50!black] at ($(r10)$) {\textbf{invalid}};


\foreach \i/\j in {1/6,3/8, 19/24}{%
  \draw[arrow,cyan!50!black] (g\i.north) -- (g\j.south);
}

\foreach \i/\j in {6/1,8/3, 11/6,13/8}{%
  \draw[arrow,cyan!50!black] (g\i.north) -- (b\j.south);
}

\foreach \i/\j in {1/11,3/13, 6/16,8/18}{%
  \draw[arrow,cyan!50!black] (b\i.north) -- (g\j.south);
}

\foreach \i/\j in {16/17,17/18,18/19}{%
  \draw[arrow,cyan!50!black] (g\i.east) -- (g\j.west);
}

\foreach \i/\j in {2/7,4/9, 16/21,17/22,18/23}{%
  \draw[arrow,black] (g\i.north) -- (g\j.south);
}

\foreach \i/\j in {7/2,9/4, 12/7, 14/9, 15/10}{%
  \draw[arrow,black] (g\i.north) -- (b\j.south);
}

\foreach \i/\j in {2/12,4/14,5/15, 7/17,9/19,10/20}{%
  \draw[arrow,black] (b\i.north) -- (g\j.south);
}

\foreach \i/\j in {6/7,7/8,8/9,9/10, 11/12,12/13,13/14,14/15, 19/20}{%
  \draw[arrow, black] (g\i.east) -- (g\j.west);
}

\foreach \i/\j in {5/10}{%
  \draw[arrow,red!50!black] (g\i.north) -- (g\j.south);
}

\foreach \i/\j in {10/5}{%
  \draw[arrow,red!50!black] (g\i.north) -- (b\j.south);
}

\foreach \i/\j in {5/15}{%
  \draw[arrow,red!50!black] (b\i.north) -- (g\j.south);
}

\foreach \i/\j in {1/2,4/5}{%
  \draw[arrow,black] (r\i.north) -- (r\j.south);
}

\foreach \i/\j in {2/7,4/9}{%
  \draw[arrow, black] (r\i.east) -- (r\j.west);
}

\foreach \i/\j in {2/2,3/3}{%
  \draw[arrow,black] (r\i.north) -- (rb\j.south);
}

\foreach \i/\j in {2/3,3/4}{%
  \draw[arrow,black] (rb\i.north) -- (r\j.south);
}

\foreach \i/\j in {6/7,9/10}{%
  \draw[arrow,red!50!black] (r\i.north) -- (r\j.south);
}

\foreach \i/\j in {7/7,8/8}{%
  \draw[arrow,red!50!black] (r\i.north) -- (rb\j.south);
}

\foreach \i/\j in {7/8,8/9}{%
  \draw[arrow,red!50!black] (rb\i.north) -- (r\j.south);
}

\foreach \i/\j in {3/8}{%
  \draw[arrow, red!50!black] (r\i.east) -- (r\j.west);
}

\end{tikzpicture}

%% file: figures/tikz/data_setup.tex
\begin{tikzpicture}[
    font=\sffamily\scriptsize,
    >=Stealth,
    node distance=4mm,
    thick,
    box/.style={
      draw,
      rounded corners=2mm,
      align=left,
      inner sep=4pt,
      text width=5.5cm
    }
]


\node[box, draw=white, text width=5.5cm] (tempHeading) {Template 1};

\node[box, fill=teal!15, draw=teal!10, text width=5.5cm, below=0 of tempHeading] (tempA)
  {[name] initially has [num1] [object]. After [verb] [num2] more [object], how many [object] does [pronoun] have?};
\node[box, fill=orange!20, text width=5.5cm, below=0.1 of tempA] (tempB)
  {To solve this, we add [num1] + [num2] = [result].\\Thus, [name] has a total of [answer] [object].};
\node[box, fill=pink!20, text width=5.5cm, below=0.1 of tempB] (tempC)
  {The above reasoning is:};

\node[
  rounded corners=3mm,
  draw,
  fill=white,
  fit=(tempHeading)(tempA)(tempB)(tempC),
  inner sep=4pt,
  shift={(-0.3cm, 0.3cm)}
] (templateC) {};

\node[
  rounded corners=3mm,
  draw,
  fill=white,
  fit=(tempHeading)(tempA)(tempB)(tempC),
  inner sep=4pt,
  shift={(-0.15cm, 0.15cm)}
] (templateMainB) {};

\node[
  rounded corners=3mm,
  thick,
  draw,
  fill=white,
  fit=(tempHeading)(tempA)(tempB)(tempC),
  inner sep=4pt
] (templateMainA) {};

\node[box, draw=white, text width=5.5cm] (tempHeading) {\bfseries Template 1};

\node[box, fill=teal!15, draw=teal!15, text width=5.5cm, below=0 of tempHeading] (tempA)
  {[name] initially has [num1] [object].\\After [verb] [num2] more object], how many \\\relax[object] does [pronoun] have?};
\node[box, fill=orange!20, draw=orange!20, text width=5.5cm, below=0.1 of tempA] (tempB)
  {To solve this, we add [num1] + [num2] = \textcolor{cyan!50!black}{\textbf{[result]}}.\\Thus, [name] has a total of \textcolor{cyan!50!black}{\textbf{[answer]}} [object].};
\node[box, fill=blue!5, draw=blue!5, text width=5.5cm, below=0.1 of tempB] (tempC)
  {The above reasoning is:};


\coordinate (split) at ($(templateMainA.east)+(0.5,0.)$);

\draw[-] (templateMainA.east) -- (split);


\node[
  rounded corners=3mm,
  thick,
  draw=none,
  fill=none,
  fit=(tempHeading)(tempA)(tempB)(tempC),
  inner sep=4pt
] (cleanFrameNode) at ($(split)+(4.0,1.8)$) {};

\node[box, draw=white, text width=5.5cm, anchor=north east]
      (cleanHeading) at ($(cleanFrameNode)+(2.5, 1.5)$)
      {\bfseries Clean prompt};

\node[box, fill=teal!15, draw=teal!15, text width=5.5cm, below=0. of cleanHeading] (cleanA)
  {Jane initially has 5 apples. After buying 8 more apples, how many apples does she have?};
\node[box, fill=orange!20, draw=orange!20, text width=5.5cm, below=0.1 of cleanA] (cleanB)
  {To solve this, we add 5 + 8 = \textcolor{red!50!gray}{\textbf{16}}.\\Thus, Jane has a total of \textcolor{green!40!black}{\textbf{13}} apples.};
\node[box, fill=blue!5, draw=blue!5, text width=5.5cm, below=0.1 of cleanB] (cleanC)
  {The above reasoning is:};

\node[
  rounded corners=3mm,
  thick,
  draw,
  fill=none,
  fit=(cleanHeading)(cleanA)(cleanB)(cleanC),
  inner sep=4pt
] (cleanFrame) {};

\draw[->] (split) |- (cleanFrame.west);


\node[
  rounded corners=3mm,
  thick,
  draw=none,
  fill=none,
  fit=(tempHeading)(tempA)(tempB)(tempC),
  inner sep=4pt
] (corrFrameNode) at ($(split)+(4.0,-2.4)$) {};

\node[box, draw=white, text width=5.5cm, anchor=north east]
      (corrHeading) at ($(corrFrameNode)+(2.5, 2.05)$)
      {\bfseries Corrupt prompt};

\node[box, fill=teal!15, draw=teal!15, text width=5.5cm, below=0. of corrHeading] (corrA)
  {Jane initially has 5 apples. After buying 8 more apples, how many apples does she have?};
\node[box, fill=orange!20, draw=orange!20, text width=5.5cm, below=0.1 of corrA] (corrB)
  {To solve this, we add 5 + 8 = \textcolor{green!40!black}{\textbf{13}}.\\Thus, Jane has a total of \textcolor{green!40!black}{\textbf{13}} apples.};
\node[box, fill=blue!5, draw=blue!5, text width=5.5cm, below=0.1 of corrB] (corrC)
  {The above reasoning is:};

\node[
  rounded corners=3mm,
  thick,
  draw,
  fill=none,
  fit=(corrHeading)(corrA)(corrB)(corrC),
  inner sep=4pt
] (corrFrame) {};

\draw[->] (split) |- (corrFrame.west);


\node[box, fill=blue!5, text width=0.7cm,
      right=5mm of cleanFrame.east] (incorrectBox)
      {\centering invalid};
\draw[->,thick] (cleanFrame.east) -- (incorrectBox.west);

\node[box, fill=blue!5, text width=0.5cm,
      right=5mm of corrFrame.east] (correctBox)
      {\centering valid};
\draw[->,thick] (corrFrame.east) -- (correctBox.west);

\end{tikzpicture}